\documentclass[10pt,journal,cspaper,compsoc]{IEEEtran}

\usepackage[left=0.4in,right=0.4in,top=0.5in,bottom=0.5in,%
            footskip=.25in]{geometry}
\ifCLASSOPTIONcompsoc
  \usepackage[nocompress]{cite}
\else
  \usepackage{cite}
\fi

\usepackage{pifont}
\newcommand{\cmark}{\ding{51}}%
\newcommand{\xmark}{\ding{55}}%
\let\appendices\relax

\usepackage{titletoc}
\usepackage{colortbl}
\usepackage{ragged2e}

\usepackage[page,header]{appendix}
\usepackage[dvipsnames]{xcolor}
\definecolor{ballblue}{rgb}{0.13, 0.67, 0.8}
\definecolor{lightseagreen}{rgb}{0.13, 0.7, 0.67}
\definecolor{org}{HTML}{F8A145}
\definecolor{blu}{HTML}{63ACE5}
\definecolor{c1}{HTML}{41B3A3}
\definecolor{c2}{HTML}{3500D3}
\usepackage[colorlinks,linkcolor=blue,citecolor=org,urlcolor=org, linktoc=all]{hyperref}
\usepackage{algorithm,algpseudocode}
\algnewcommand\algorithmicto{\textbf{to}}
\algrenewtext{For}[3]%
{\algorithmicfor\ $#1 \gets #2$ \algorithmicto\ $#3$ \algorithmicdo}

 \usepackage[T1]{fontenc}    
\usepackage{hyperref}       
\usepackage{url}            
\usepackage{booktabs}       
\usepackage{nicefrac}       
\usepackage{amsthm}
\usepackage{amsmath}
\usepackage{multirow}
\usepackage{amssymb}
\usepackage{subfigure}
\usepackage{graphicx}
\usepackage[framemethod=tikz]{mdframed}

\DeclareMathOperator*{\expe}{\mathbb{E}}
\def \newline{{\color{white} dsdsdsa}}

\DeclareMathOperator*{\argmin}{argmin}
\DeclareMathOperator*{\arginf}{arginf}

\newcommand{\PreserveBackslash}[1]{\let\temp=\\#1\let\\=\temp}

\newcommand{\rom}[1]{\uppercase\expandafter{\romannumeral#1}}
\usepackage{amssymb}  
\usepackage{subfiles}
\usepackage{bm}
 
\let\labelindent\relax
\usepackage{enumitem}
\def \aucovo {{\rm AUC}^{\rm {ovo}}}
\def \aucova {{{\rm AUC}^{\rm {ova}}}}
\def \aucij {{{\rm AUC}_{i|j}}}
\def \ezz{ \expe_{\bm{z}_1,\bm{z}_2}}
\def \phil{\phi_{\ell}}
\def \Rx {R_{\mathcal{X}}}
\newtheorem{thm}{Theorem}
\newtheorem*{rthm1}{Restate of Theorem \ref{prop:mo}}
\newtheorem*{rthm2}{Restate of Theorem \ref{prop:bay}}
\newtheorem*{rthm3}{Restate of Theorem \ref{thm:consq}}
\newtheorem*{rthm4}{Restate of Theorem \ref{thm:gen}}
\newtheorem*{rthm5}{Restate of Theorem \ref{thm:gen_2}}
\newtheorem*{rthm6}{Restate of Theorem \ref{thm:chain}}
\newtheorem*{rthm7}{Restate of Theorem \ref{thm:trans}}
\newtheorem*{rthm8}{Restate of Theorem \ref{thm:gen_3}}
\newtheorem*{rthm9}{Restate of Theorem \ref{thm:conv}}

\newtheorem*{rdef3}{Restate of Definition \ref{def:ecover}}

\newtheorem*{rdef4}{Restate of Definition \ref{def:covernum}}

\newtheorem*{rprop1}{Restate of Proposition \ref{prop:ub}}
\def \funt {\tilde{f}}
\def \st{\tilde{\bm{s}}}
\def \s{\bm{s}}
\newtheorem{defi}{Definition}

\newtheorem*{pf}{Proof}
\newtheorem{prop}{Proposition}

\newtheorem{lem}{Lemma}
\newtheorem{rem}{Remark}
\newtheorem{asm}{Assumption}
\newtheorem{col}{Corollary}
\newtheorem{exa}{Example}

\newtheorem{clm}{Claim}

\DeclareRobustCommand{\rchi}{{\mathpalette\irchi\relax}}
\newcommand{\irchi}[2]{\raisebox{\depth}{$#1\chi$}} 

\newcommand{\es}[1]{\expe_{\mathcal{S}}  {#1} }
\def \tfs {T_f(\sigma)}

\def \tfsd{T_\cone{f}(\sigone) - T_{\ctwo{\funt}}(\sigone)}

\def \tfsdi{T_{\cone{f}}(\sigtwo) - T_{\ctwo{\funt}}(\sigtwo)}
\def \sigone {\cone{\sigma}}
\def \sigtwo {\ctwo{\sigma_{\backslash (i,k)}}}
\newcommand{\cone}[1]{{\color{red}#1}}
\newcommand{\ctwo}[1]{{\color{c2}#1}}

\def \conef {{\cone{f}}}
\def \coneP {{\cone{\bm{P}}}}
\def \coneK {{\cone{\bm{K}^{(i)}}}}
\def \coneV {{\cone{\bm{V}^{(i)}}}}
\def \ctwoft {\ctwo{\tilde{f}}}
\def \ctwoP{\ctwo{\tilde{\bm{P}}}}
\def \ctwoK{\ctwo{\tilde{\bm{K}}^{(i)}}}
\def \ctwoV{\ctwo{\tilde{\bm{V}}^{(i)}}}
\def \conehf {{\cone{\hat{f}}}}
\def \ctwohft {{\ctwo{\hat{\funt}}}}
\def \conehfk {{\cone{\hat{f}_{k}}}}
\def \ctwohftk {{\ctwo{\hat{\funt}_{k}}}}
\newcommand{\conehfn}[1]{{\cone{\hat{f}_{#1}}}}
\newcommand{\ctwohftn}[1]{{\ctwo{\hat{\funt}_{#1}}}}

\def \cones {{\cone{\bm{s}}}}
\def \ctwost {\ctwo{\tilde{\bm{s}}}}
\def \conehs {{\cone{\hat{\bm{s}}}}}
\def \ctwohst {{\ctwo{\hat{\tilde{\bm{s}}}}}}
\def \conehsk {{\cone{\hat{\bm{s}}_k}}}
\def \ctwohstk {{\ctwo{\hat{\tilde{\bm{s}}}_k}}}

\newcommand{\conehsn}[1]{{\cone{\hat{\bm{s}}_{#1}}}}
\newcommand{\ctwohstn}[1]{{\ctwo{\hat{\tilde{\bm{s}}}_{#1}}}}

\def \supcover {\sup_{
	\begin{subarray}
	{
	} \hat{\s}_k \in \hat{\mathcal{H}}_{k},\hat{\s}_{k-1} \in \hat{\mathcal{H}}_{k-1}\\
	d_{\infty, \mathcal{S}}(\hat{\s}_k, \hat{\s}_{k-1}) \le 3 \epsilon_{k}
	\end{subarray}
   }
}

\def \pione {\pi^{(i)}(\bm{x}_1,\bm{x}_2)}
\def \pitwo {\pi^{(i)}(\bm{x}_2,\bm{x}_1)}

\def \aucinj {{\rm {AUC}}_{i|\neg i}}

\def \sR {\mathfrak{R}_{\mauc}(\ell\circ\mathcal{H})}
\def \empsR {\hat{\mathfrak{R}}_{\mauc,\mathcal{S}}(\ell\circ\mathcal{H})}
\def \empsRF {\hat{\mathfrak{R}}_{\mauc,\mathcal{S}}(\ell\circ\mathsf{soft}\circ \mathcal{F})}
\def \empsRFb {\hat{\mathfrak{R}}_{\mauc,\mathcal{S}}(\ell\circ\mathsf{soft}\circ \mathcal{F}_{\beta,\nu})}
\def \empsRLin {\hat{\mathfrak{R}}_{\mauc,\mathcal{S}}(\ell\circ \Hlin)}
\def \empsRDnn {\hat{\mathfrak{R}}_{\mauc,\mathcal{S}}(\ell\circ \Hsdnn)}

\def \lfzoi {\ell_{0-1}(f^{(i)},\bm{x}_1,\bm{x}_2,\yif{1},\yif{2}) }

\def \ts{\tilde{S}}
\def \tsp{\tilde{S}'}
\def \Yi {\bm{Y}^{(i)}}
\def \Yit {\bm{Y}^{(i)^\top}}
\def \Rhatf {\hat{R}_{\ell,\mathcal{S}}(f)}
\def \Di {\bm{D}^{(i)}}
\def \Yni {\bm{1} - \Yi}

\newcommand{\aucf}[2]{{\rm AUC}_{#1|#2}}
\newcommand{\yif}[1]{y^{(i)}_{#1}}

\def \Hdnn {\mathcal{H}^{DNN}_{\gamma,R_s, n_h}}
\def \Hsdnn {\mathsf{soft} \circ \mathcal{H}^{DNN,nh}_{\gamma}}
\def \aucji {{\rm AUC}_{j|i}}

\def \lrank {\bm{I}\left [ \Delta(\yi) \cdot \Delta(f^{(i)})  < 0 \right] + \frac{1}{2} \bm{I}\left[ \Delta(f^{(i)}) = 0\right] }
\def \exx {\expe_{\bm{x}_1,\bm{x}_2}}

\newcommand{\fii}[1][\x]{f^{(i)}(#1)}
\newcommand{\fjj}[1][\x]{f^{(j)}(#1)}
\newcommand{\pfij}[1][\x]{\partial_{i}\fjj[#1]}
\newcommand{\hth}[1][\x]{h_{\bm{\theta}}(#1)}
\def \Hlin {\mathcal{H}^{Lin}_{p,\gamma}}

\def \supf { \sup_{f \in \mathcal{H}}
}
\def \nc {\frac{1}{N_C\cdot (N_C-1)}}

\def \fixm {\fii[\x_m]}
\def \fixn {\fii[\x_n]}

\def \fjxm {\fjj[\x_m]}
\def \fjxn {\fjj[\x_n]}

\def \wi{\bm{w}^{(i)}}

\def \dfi {\fixm - \fixn}

\newcommand{\ess}[1]{\expe_{\mathcal{S},\mathcal{S'} } {#1}}
\def \js{\hat{R}_{\mathcal{S}}(f)}
\def \jsp{\hat{R}_{\mathcal{S}'}(f)}
\def \jsi{\hat{R}_{\mathcal{S}_i}(f)}
\def \jst{\hat{R}_{\widetilde{\mathcal{S}}}(f)}
\def \jspt{\hat{R}_{\widetilde{\mathcal{S}}'}(f)}
\newcommand{\jfun}[1]{R_{\widetilde{\mathcal{S}}^{#1}}(f)}
\newcommand{\xpos}[2]{\x^{(#1)\downarrow}_{#2}}
\newcommand{\xneg}[2]{\bm{\tilde{x}}^{(#1)\downarrow}_{#2}}

\newcommand{\jfunp}[1]{R_{\ell,\widetilde{\mathcal{S}}^{#1'}}(f)}

\def \Ni {\mathcal{N}_i}
\def \Nj {\mathcal{N}_j}
\def \ni {{n}_i}
\def \nj {{n}_j}

\newcommand{\lfxy}[2]{\ell(f^{(i)},\bm{#1}_m,\bm{#2}_n)}
\def \supw {\sup_{\bm{W}, ||\bm{W}^{(i)}||_p \le  \gamma }}
\def \suphd {\sup_{~~~f \in \Hdnn }}
\def \eij {\mathcal{E}^{(ij)}}
\def \ei {\mathcal{E}^{(i)}}
\def \yi {y^{(i)}_{1,2}}
\def \supwi {\sup_{
\begin{subarray}{c}
\bm{W}^{(i)}\\
||\bm{W}^{(i)}||_p \le  \gamma
\end{subarray}
}}
\def \x {\bm{x}}
\def \xp {\bm{x}'}
\def \sumi {\sum_{i=1}^{N_C}}
\def \suminj {\sum_{j \neq i}}
\def  \sumxone {\sum_{\bm{x}_m \in \Ni }}
\def  \sumxtwo {\sum_{\bm{x}_n \in \Nj }}
\def \ninj {\frac{1}{n_in_j}}
\def \sigmai {\sigma^{(i)}}
\def \sigmaj {\sigma^{(j)}}

\def \mauc {{\mathsf{MAUC}^\downarrow}}
\def \sumauc {\sumi \suminj \sumxone \sumxtwo \ninj \cdot}

\def \empcomp {\sumi \suminj \sumxone \sumxtwo \frac{\sigmai_m + \sigmaj_n }{2} \cdot   \frac{  \ell(f^{(i)}(\bm{x}_m)-  f^{(i)}(\bm{x}_n))}{n_in_j}}
\def \empcompsim {\sumi \suminj \sumxone \sumxtwo T^{i,j,m,n}}
\def \sigmaterm {\frac{\sigmai_m + \sigmaj_n }{2} \cdot   \frac{  \ell(f^{(i)}(\bm{x}_m)-  f^{(i)}(\bm{x}_n))}{n_in_j}}

\def \epxone {\eta^{(i)}_+(\x_1)}
\def \epxtwo {\eta^{(i)}_+(\x_2)}
\def \epx {\eta^{(i)}_+(\x)}
\def \epxp {\eta^{(i)}_+(\x')}

\def \enxone {\eta^{(i)}_-(\x_1)}
\def \enxtwo {\eta^{(i)}_-(\x_2)}
\def \enx {\eta^{(i)}_-(\x)}
\def \enxp {\eta^{(i)}_-(\x')}

\def \ratone {\frac{\epxone}{\enxone}}
\def \rattwo {\frac{\epxtwo}{\enxtwo}}

\def \Rl {R^{(i)}_\ell}

\newcommand{\fif}[1]{f^{(i)}\left(#1\right)} 

\newcommand{\fifs}[1]{f^{\star(i)}\left(#1\right)}

\newcommand{\ldiff}[2]{\ell'\left(\fifs{#1} - \fifs{#2}\right)}

\begin{document}

\title{Learning with Multiclass AUC: \\ Theory and Algorithms}

%

\author{Zhiyong~Yang, Qianqian~Xu*,~\IEEEmembership{Senior Member,~IEEE,} Shilong Bao,\\
        Xiaochun~Cao,~\IEEEmembership{Senior Member,~IEEE,}
        and~Qingming~Huang*,~\IEEEmembership{Fellow,~IEEE}
\IEEEcompsocitemizethanks{
\IEEEcompsocthanksitem * corresponding authors\protect\\
\IEEEcompsocthanksitem Zhiyong Yang with the School of Computer Science and Technology,
University of Chinese Academy of Sciences, Beijing 101408, China (email: \texttt{yangzhiyong@iie.ac.cn}). \protect\\
\IEEEcompsocthanksitem Qianqian Xu is with the Key Laboratory of
Intelligent Information Processing, Institute of Computing Technology, Chinese
Academy of Sciences, Beijing 100190, China, (email: \texttt{xuqianqian@ict.ac.cn}).\protect\\
\IEEEcompsocthanksitem Shilong Bao is with the State Key Laboratory of Information Security, Institute of Information Engineering, Chinese Academy of Sciences, Beijing 100093, China, and also with the School of Cyber Security, University of Chinese Academy of Sciences, Beijing 100049, China (email: \texttt{baoshilong@iie.ac.cn}).\protect\\
\IEEEcompsocthanksitem Xiaochun Cao is with the State Key Laboratory of Information Security, Institute of Information Engineering, Chinese Academy of Sciences, Beijing 100093, China, also with School of Cyber Science and Technology, Sun Yat-sen University, Shenzhen, 518100, China (email: \texttt{caoxiaochun@iie.ac.cn}).\protect\\
\IEEEcompsocthanksitem Q. Huang is with the School of Computer Science and Technology,
    University of Chinese Academy of Sciences, Beijing 101408, China, also
    with the Key Laboratory of Big Data Mining and Knowledge Management (BDKM),
    University of Chinese Academy of Sciences, Beijing 101408, China,  also
    with the Key Laboratory of Intelligent Information Processing, Institute of
    Computing Technology, Chinese Academy of Sciences, Beijing 100190, China, and also with Peng Cheng Laboratory, Shenzhen 518055, China
    (e-mail: \texttt{qmhuang@ucas.ac.cn}).\protect\\
   }
   }

\markboth{IEEE TRANSACTIONS ON PATTERN ANALYSIS AND MACHINE INTELLIGENCE}%
{Yang \MakeLowercase{\textit{et al.}}: Bare Demo of IEEEtran.cls for Computer Society Journals}



\maketitle
\IEEEpeerreviewmaketitle

\begin{abstract}

The Area under the ROC curve (AUC) is a well-known ranking metric for problems such as imbalanced learning and recommender systems. The vast majority of existing AUC-optimization-based machine learning methods only focus on binary-class cases, while leaving the multiclass cases unconsidered. In this paper, we start an early trial to consider the problem of learning multiclass scoring functions via optimizing multiclass AUC metrics. Our foundation is based on the M metric, which is a well-known multiclass extension of AUC. We first pay a revisit to this metric, showing that it could eliminate the imbalance issue from the  minority class pairs. Motivated by this, we propose an empirical surrogate risk minimization framework to approximately optimize the M metric. Theoretically, we show that: (i) optimizing most of the popular differentiable surrogate losses suffices to reach the Bayes optimal scoring function asymptotically; (ii) the training framework enjoys an imbalance-aware generalization error bound, which pays more attention to the bottleneck samples of minority classes compared with the traditional $O(\sqrt{1/N})$ result. Practically, to deal with the low scalability of the computational operations, we propose acceleration methods for three popular surrogate loss functions, including the exponential loss, squared loss, and  hinge loss, to speed up loss and gradient evaluations. Finally, experimental results on 11 real-world datasets demonstrate the effectiveness of our proposed framework.  
\end{abstract}

\begin{IEEEkeywords}
AUC Optimization, Machine Learning.
\end{IEEEkeywords}


{\section{Introduction}\label{sec:introduction}}
\IEEEPARstart{A}{UC} (Area Under the ROC Curve), which measures the probability that a positive instance has a higher score than a negative instance, is a well-known performance metric for a scoring function's ranking quality. AUC often comes up as a  more appropriate performance metric than accuracy in various applications due to its appealing properties, \textit{e.g.}, insensitivity toward label distributions and costs \cite{rocsur,AUCM}. On one hand, for class-imbalanced tasks such as disease prediction \cite{Diagnosis1,Diagnosis2} and rare event detection \cite{Anomaly1,Anomaly2,transactions1}, the label distribution is often highly skewed in the sense that the proportion of the majority classes instances significantly dominates the others. As for a typical instance, in the credit card fraud detection dataset released in Kaggle \footnote{
\url{https://www.kaggle.com/mlg-ulb/creditcardfraud}}, the fraudulent transactions only account for $0.172\%$ of the total records.
 Accuracy is often not a good choice in this case since it might ignore the performance from the minority classes which are often more crucial than the majority ones. By contrast, the value of AUC does not rest on the label distribution, making it  a natural metric under the class-imbalance scenario. On the other hand, AUC has also been adopted as a standard metric for applications such as ads click-through rate prediction and recommender systems \cite{rec1,rec2,rec3,ctr1} where pursuing a correct ranking between positive and negative instances is much more critical than label prediction. 

 Over the past two decades, the importance of AUC has raised an increasing favor in the machine learning community to explore direct AUC optimization methods, \emph{e.g.}, \cite{logitauc,svmauc1,svmauc2,logitauc2,aucopt1,auconepass,aucopt2}. However, to our knowledge, the vast majority of related studies merely focus on the binary class scenario. Since real-world pattern recognition problems often involve more than two classes, it is natural to pursue its generalization in the world of multiclass problems. To this end, we present a very early study to the problem of AUC guided machine learning framework under the multiclass setting. Specifically, 
 we propose a universal empirical surrogate risk minimization framework with theoretical guarantees. 
 
 First of all, we provide a review of a multiclass generalization of AUC known as the M metric \cite{AUCM}. Specifically, we show that  M metric is an appropriate extension of binary AUC in the sense that it could efficiently avoid the imbalanced issue across class ranking pairs. Motivated by this result, we propose to minimize the 0-1 mis-ranking loss induced by the M metric denoted as $\mauc
 $. However, directly minimizing the objective is intractable.  This intractability has three sources: \textbf{(a)} the 0-1 mis-ranking loss $\mauc$ is a discrete and non-differentiable function, making it impossible to perform efficient optimizations; \textbf{(b)} the data distribution is not a known \textit{priori}, making the calculation of expectation unavailable; \textbf{(c)} the complexity to estimate the loss and gradient function could be as high as $O(N^2N^2_C)$ in the worst case, where $N$ is the number of samples and $N_C$ is the number of classes.

 Targeting at  \textbf{(a)}, in Sec.\ref{sec:b}, we investigate how to construct differentiable surrogate risks as replacements for the 0-1 risk. To do this, we derive the set of Bayes optimal score functions under the $\mauc$ criterion. We then provide a general sufficient condition for the fisher consistency. The condition suggests that a large set of popular surrogate loss functions are fisher consistent under certain assumptions in the sense that optimizing the corresponding surrogate expected risk also leads to the Bayes optimal score functions.

 Based on the consistency result, in Sec.\ref{sec:c}, we construct an empirical surrogate risk minimization framework against \textbf{(b)}, where we propose an unbiased empirical estimation of the surrogate risks over a training dataset as the objective function to avoid using population-level expectation directly.  Moreover, we provide a systematic analysis of its generalization ability. The major challenge here is that the empirical risk function could not be decomposed as a finite sum of independent instance-wise loss terms, making the traditional symmetrization technique not available. To fix this problem, we provide a novel form of Rademacher complexity for $\mauc$. Furthermore, we provide generalization upper bounds for a wide range of model classes, including shallow and deep models. The results consistently enjoy an imbalance-aware property, which pays more attention to the bottleneck samples of minority classes than the traditional result. 
 
  In Sec.\ref{sec:d}, \textbf{(c)} is attacked by novel acceleration algorithms to speed up loss and gradient evaluations which are the fundamental calculations for gradient-based optimization methods. The mini-batch/full-batch loss and gradient evaluations could be done, with the proposed algorithms,  in a complexity comparable with that for the ordinary instance-wise loss functions.
 
 Finally, in Sec.\ref{sec:exp}, we perform extensive experiments on 11 real-world datasets to validate our proposed framework.
 
 To sum up, our contribution is three-fold:
 \begin{itemize}
 \item \textbf{New Framework}: We provide a theoretical framework for empirical risk minimization under the guidance of the  M metric.
 \item \textbf{Theoretical Guarantees}: From a theoretical perspective, our framework is soundly supported by consistency analysis and generalization analysis. 
 \item \textbf{Fast Algorithms}: From a practical perspective, we propose efficient algorithms for three surrogate losses to accelerate loss and gradient evaluations. The experiments show that the \emph{\textbf{acceleration ratio could reach $\bm{1,0000}+$}} confronting medium size datasets.
 \end{itemize}

\section{Related Work}\label{sec:re}
\textbf{Binary Class AUC Optimization.} As a motivating early study, \cite{AUCvErr} points out that maximizing AUC  should not be replaced with minimizing the error rate, which shows the necessity to study direct AUC optimization methods. After that, a series of algorithms have been designed for optimizing AUC. At the early stage, the majority of studies focus on a full-batch off-line setting. \cite{logitauc,logitauc2} optimize AUC based on a logistic surrogate loss function and ordinary gradient descent method. RankBoost \cite{boost} provides an efficient ensemble-based AUC learning method based on a ranking extension of the AdaBoost algorithm. \cite{svmauc2,svmauc3} constructs $SVM^{struct}$-based frameworks that optimize a direct upper bound of the $0-1$ loss version AUC metric instead of its surrogates. Later on, to accommodate big data analysis, researchers start to explore online extensions of AUC optimization methods. \cite{online1} provides an early trial for this direction based on the reservoir sampling technique. \cite{auconepass} provides a completely one-pass AUC optimization method for streaming data based on the squared surrogate loss. \cite{minimaxauc1} reformulates the squared-loss-based stochastic AUC maximization problem as a stochastic saddle point problem. The new saddle point problem's objective function only involves summations of instance-wise loss terms, which significantly reduces the burden from the pairwise formulation. \cite{minimaxauc2,minimaxauc3} further accelerate this framework with tighter convergence rates. On top of the reformulation framework, \cite{aucgenloss} also provides an acceleration framework for general loss functions where the loss functions are approximated by the Bernstein polynomials. \cite{tri} proposes a novel
large-scale nonlinear AUC maximization method based on the triply stochastic gradient descent algorithm. \cite{scalable} proposes a scalable and efficient adaptive doubly stochastic gradient algorithm for generalized regularized pairwise learning problems. Beyond optimization methods,  a substantial amount of researches also provide theoretical support for this learning framework from different dimensions, including generalization analysis \cite{genal,cle1,aucrade,geninde,genpac} and consistency analysis \cite{auccon1,auccons}. Last but not least, there are also some studies focusing on optimizing the  partial area under the ROC curve \cite{pauc1,pauc2,partialspeech}. This paper takes a further step by providing an early study of the theory and algorithms for AUC-guided machine learning under the more complicated multiclass scenario.

\noindent \textbf{Multiclass AUC Metrics}. There exist two general ideas for how to define a multiclass AUC metric.  The first idea is on top of the belief that multiclass counterparts of the ROC curve should be represented as higher-dimensional surfaces. As a result, AUC is generalized naturally to the volume Under some specifically designed ROC Surfaces (VUS) \cite{vus1,vus2}. However, this idea is restricted by its high complexity to calculate the volume of such high-dimensional spaces. According to \cite{vus3}, calculating VUS for $N$ samples and $N_C$ classes requires $O(N\log N + N^{\lfloor N_C/2 \rfloor})$ time complexity and $O(N^{\lfloor N_C/2 \rfloor})$ space complexity. Another idea then comes out to do it in a much simpler manner, which suggests that one can simply take an average of pairwise binary AUCs \cite{AUCM, aucmm,MAUCeval1,MAUCeval2}. This is based on the intuition that if every pair of classes is well-separated from each other in distribution, one can get reasonable performances. Getting rid of calculations on high dimensional spaces renders this formulation a low complexity. Due to its simplicity, the M metric proposed in the representative work \cite{AUCM} has been adopted by a series of popular machine learning software such as {\texttt{scikit-learn}} in \texttt{Python} and \texttt{pROC} in \texttt{R}. Most recently, \cite{maucdyna} also considers online extensions of multi-class AUC metric to deal with the concept drift issue for streaming data. Unlike this line of research, our main focus in this paper is how to learn valuable models from a proper multiclass extension of the AUC metric.\\
\noindent \textbf{AUC Optimization for Multiclass AUC Optimization}. There are few studies that focus on AUC optimization algorithms for multiclass problems, which fall in the following two directions. The first direction of studies focuses on the multipartite ranking problem, which is a natural extension of the bipartite ranking problem where the order is presented with more than two discrete degrees. Hitherto, a substantial amount of efforts have been made to explore the AUC optimization method/theory for the multipartite ranking problem \cite{multipart1,ranking1,ranking2,ranking3, ranking4}. Moreover, a recent work \cite{quad} proposes a novel nonlinear semi-supervised multipartite ranking problem for large-scale datasets.  It is noteworthy that multipartite ranking approaches could solve multiclass problems only if the classes are ordinal values for the same semantic concept. For example, the age estimation task could be regarded as a multiclass problem where the class labels are the ages for a given person; the movie rating prediction could be regarded as a multiclass problem where the class labels are the ratings for the same given movie. The major difference here is that we focus on the generic multiclass problems where different labels present different semantic concepts and do not have clear ordinal relations. Thus the studies along this line are not available to our setting in general. Most recently, \cite{deepmulti} discusses the possibility of exploring AUC optimization for general multiclass settings. The major differences are as follows. First, \cite{deepmulti} only focuses on the square loss function, while our study presents a general framework for multiclass AUC optimization. Second, \cite{deepmulti} focuses more on the optimization properties, where the original problem is reformulated as a minimax problem. By contrast, our study focuses on the learning properties for multiclass AUC optimization, such as its generalization ability and the consistency of different loss functions. Moreover, we also present a series of acceleration methods that do not require any reformulation of the optimization problem.

\section{Learning with Multiclass AUC Metrics: An overview}\label{sec:a}
\subsection{Preliminary}\label{sec:pre}
\textbf{Basic Notations}. In this paper, \textbf{we will constantly adopt two sets of events}: for pair-wise AUC metrics,  $\mathcal{E}^{(ij)}$ denotes the event $y_1 = i, y_2 = j \vee y_1 = j, y_2 = i $, while $\mathcal{E}^{(i)}$ denotes the event $y_1 = i, y_2 \neq i \vee y_1 \neq i, y_2 = i $. Given an event $\mathcal{A}$, ${I}\left[\mathcal{A}\right]$ is the indicator function associated with this event, which equals 1 if $\mathcal{A}$ holds and equals to 0 otherwise. Given a finite dataset $\mathcal{S}$, we denote $N_C$ as the number of classes and $N$ as the total number of sample points in the dataset. 

\noindent{\textbf{Settings}}. For an $N_C$ class problem, we assume that our samples are drawn from a product space $\mathcal{Z} = \mathcal{X} \times  \mathcal{Y}$. $\mathcal{X}$ is the input feature where $\mathcal{X} \subset \mathbb{R}^d$ and $d$ is the input dimensionality. $\mathcal{Y}$ is the label space $[N_C]$. Given a label $y_m= i$, we will use the one-hot vector $\bm{y}_m= [y^{(1)}_m,\cdots,y^{(N_C)}_{m}]$ to represent it, with $y^{(k)}_{m} = 0, k \neq i$, and $y^{(k)}_m = 1$ if $k = i$.  In this paper, we will adopt the one \textit{vs}. all decomposition \cite[\textcolor{org}{Chap.9.4}]{mlfun} of the multiclass  problem. Under this context, an $N_C$-class ($N_C>2$) scoring function refers to  a set of $N_C$  functions $f = (f^{(1)}, \cdots, f^{(N_C)})$, where $f^{(i)}: \mathcal{X} \rightarrow \mathbb{R}$ serves as a \textit{continuous} score function supporting $y=i$.  

\noindent{\textbf{Binary Class AUC}}. Let $\mathcal{D}_{\mathcal{Z}}$ denote the joint data distribution and $f$ denote a score function estimating the possibility that an underlying instance belongs to the positive class. In the context of binary class problems, AUC is known to have a clear statistical meaning: it is equivalent to the Wilcoxon Statistics \cite{ROCmean}, namely the possibility that correct ranking takes place between a random paired samples with distinct labels: $
{\rm AUC}(f) = \mathbb{P}\big[\Delta(y)\Delta(f) >0 | \mathcal{E}^{(0,1)}\big] + \frac{1}{2} \mathbb{P}\big[\Delta(f) = 0 | \mathcal{E}^{(0,1)}\big]= \expe_{\bm{z}_1  \sim \mathcal{D}_{\mathcal{Z} },\bm{z}_2 \sim \mathcal{D}_{\mathcal{Z}}}\big[\bm{I}[\Delta(y)\Delta(f) > 0] + \frac{1}{2}\bm{I}[\Delta(f) = 0] | \mathcal{E}^{(0,1)}\big],
$
   where $ \Delta(y) = y_1 - y_2, ~~ \Delta(f) = f(\x_1)  - f(\x_2)$. Note that we adopt the convention \cite{cle1,auccon1,auccons} to score ties with 0.5. See basic notations in the Sec.\ref{sec:pre} for  $\mathcal{E}^{(0,1)}$. Hereafter, our calculations on AUC follows this definition.
\subsection{Motivation}\label{subsec:moti}
First of all, we start our study by finding a proper multiclass metric from the existing literature. As presented in Sec.\ref{sec:re}, the main idea to construct multiclass AUC is to express it as an average of binary class AUCs. { Just like the way how multiclass classification is transformed into a set of binary classification problems, one can derive multiclass AUC metrics from the following two regimes.} 

\noindent{\textbf{One \textit{vs}. All Regime (ova)}}. Given an ova score function $f = (f^{(1)},\cdots, f^{(N_C)})$, \cite{aucmm} suggests that one can construct a pairwise AUC score for each $f^{(i)}$. Here, the positive instances are drawn from the $i$-th class, and the negative instances are drawn from the $j$-th distribution conditioned on $j \neq i$. The overall AUC score $\aucova$ is then an average of all $N_C$ pairwise scores, which is defined as follows. Note that here we adopt an equally-weighted average to avoid the imbalance issue.
\begin{equation*}
\aucova(f) = \dfrac{1}{N_C}\sum_{i=1}^{N_C}\aucinj(f^{(i)}),
\end{equation*}
where $\aucinj(f^{(i)}) = \expe_{\bm{z}_1,\bm{z}_2}\big[\bm{I}[\Delta(\yi)\Delta(f^{(i)}) > 0| \ei] + \frac{1}{2}\bm{I}[\Delta(f^{(i)}) = 0] | \mathcal{E}^{(i)}\big], ~ \Delta(\yi) = \yif{1} - \yif{2}, ~  \Delta(f^{(i)}) = f^{(i)}(\x_1) - f^{(i)}(\x_2)$. See basic notations in the Sec.\ref{sec:pre} for the definition of $\ei$.

\textbf{One \textit{vs}. One Regime (ovo, M metric)}. Alternatively, according to \cite{AUCM}, we can formulate a multiclass metric as an average of binary AUC scores for every class pair $(i,j)$, which is defined as:
\begin{equation*}
\aucovo(f) = \dfrac{\sum_{i=1}^{N_C}\sum_{j\neq i}\aucij(f^{(i)})}{N_C(N_C-1)},
\end{equation*}
where $\aucij(f^{(i)}) = \ezz\big[\bm{I}[\Delta(\yi)\Delta(f^{(i)}) > 0| \eij] + \frac{1}{2}\bm{I}[\Delta(f^{(i)}) = 0] | \mathcal{E}^{(ij)}\big]$, note that $\aucij \neq \aucji$, since they employ different score functions. See basic notations in Sec.\ref{sec:pre} for the definition of $\eij$.  

The following theorem reveals that $\aucovo$ is more insensitive toward the imbalanced distribution of the class pairs than $\aucova$. 

\begin{thm}[\textbf{Comparison Properties}]\label{prop:mo}
Given the label distribution as $\mathbb{P}[y=i] = p_i>0$ and a multiclass scoring function $f$. The following properties hold:
\begin{enumerate}
\item[(a)] We have that : 
\begin{equation}
	\aucova(f) = \nicefrac{1}{N_C}\sumi\suminj (\nicefrac{p_j}{1-p_i}) \cdot \aucij(f^{(i)}).
\end{equation}
\item[(b)] 
\begin{equation*}
\begin{split}
&\aucova(f)
=\aucovo(f),~ \text{when} ~~  p_i =\nicefrac{1}{N_C}, \\
&i=1,2,\cdots, N_C.
\end{split}
\end{equation*}
\item[(c)] We  have $\aucova(f) = 1$ if and only if $\aucovo(f) = 1.$
\end{enumerate}
\end{thm}
Thm.\ref{prop:mo}-(a) states that  $\aucova$ weights  different 
$\aucij$ with $\nicefrac{p_j}{1-p_i}$, which will overlook the performance of the minority class pairs.  On the contrary, $\aucovo$ assigns equal weights for all pair-wise AUCs,  which naturally avoids this issue. Thm.\ref{prop:mo}-(b) suggests that  $\aucova$ and $\aucovo$
 tend to be equivalent when the label distribution is nearly balanced.  Thm.\ref{prop:mo}-(c) further shows that $\aucova$ and $\aucovo$ agree with each other when $f$ maximizes the performance. 
 Practically, the following example shows that how the imbalance issue of class pairs affects model selection under different criterions.
 \begin{exa}
 Consider a three-class classification dataset with a label distribution $p_1 = 0.5,p_2= 0.45,p_3 =0.05$, we assume that there are two scoring functions $f_a,f_b$, where  $\aucij(f_a)$ are all 1 except that ${\aucf{1}{3}}(f_a) = 0.5$, $\aucij(f_b)$ are all 1 except that ${\aucf{1}{2}}(f_b) = 0.8$. Consequently, we have $\aucova(f_a) = 98.3$ and  $\aucova(f_b) = 93.3$, while $\aucovo(f_a) = 91.6$ and  $\aucovo(f_b) = 96.6$.
 \end{exa}
 Obviously, $f_a$ in the example should be a bad scoring function since  $f^{(1)}_a$ can not tell apart class-1 and class-3, while $f_b$ is a much better choice since it has good performances in terms of all the pairwise AUCs. We see that $\aucovo$ supports choosing $f_b$ against $f_a$, which is consistent with our expectations. However, $\aucovo$ chooses $f_a$ against $f_b$ with a significant margin. This is because that the minority class 3 brings an extremely low weight on $\aucf{1}{3}$ in $\aucova$. This tiny weight makes the fatal disadvantage of $\aucf{1}{3}(f_a)$ totally ignored. 
 
 From the theoretical and practical analysis provided above, we can draw the conclusion that $\aucovo$ is a better choice than $\aucova$.
\subsection{Objective and the Roadmap}
\textbf{Objective}. Our goal in this paper is then to construct learning algorithms that maximize  $\aucovo$. To fit in the standard machine learning paradigm, 
we follow the widely-adopted convention \cite{cle1,auccons,auccon1,pauc1,pauc2} to cast the maximization problem into an expected-risk-based minimization problem \underline{$f \in \argmin_{f} R(f)$}, where: 
\begin{equation*}
\begin{split}
R(f)  &= \mauc = \sumi \suminj \frac{\expe_{\bm{z}_1,\bm{z}_2}[\ell^{i,j,1,2}_{0-1}|\eij]}{N_C(N_C-1)},\\
\end{split}
\end{equation*}
\begin{equation*}
\begin{split}
\ell^{i,j,1,2}_{0-1} &= \lfzoi,\\
&= \lrank
\end{split}
\end{equation*} 
is the $0$-$1$ mis-ranking loss. 

\noindent{\textbf{Roadmap}}. The major challenge in this work is that directly minimizing of $R(f) $ is almost impossible in the sense that: \textbf{(a)} the 0-1 mis-ranking loss $\mauc$ is not differentiable; \textbf{(b)} the calculation of population-level expectation is unavailable; \textbf{(c)} the complexity to estimate the ranking loss is $O(N^2N_C^2)$ in the worst case. In Sec.\ref{sec:b}-Sec.\ref{sec:d}, we will present solutions to address \textbf{(a)}-\textbf{(c)}, respectively.

\section{Bayes Optimal Classifier, and Consistency Analysis for Surrogate Risk Minimization}\label{sec:b}

Since the 0-1 mis-ranking loss is a discrete and non-differentiable function, directly solving the optimization problem is almost intractable.  In this section, we will construct surrogate risks $R_\ell(f)$ with surrogate losses $\ell$ as differentiable proxies for the 0-1 mis-ranking loss. We start with finding the Bayes optimal scoring function that we should approximate. Then we provide the surrogate risk minimization framework and investigate if it could approximate the true minimization problem well.

\noindent \textbf{Bayes Optimal Scoring Functions}. First, let us derive what are the best scoring functions that we need to approximate. With a goal to minimize $\mauc$, $f$ reaches the best performance when it realizes the minimum of expected risk under the 0-1 mis-ranking loss. In this paper, since we are dealing with classification problems, we restrict the choice of each $f^{(i)}$ to measurable functions with range $[0,1]$, \emph{i.e.}, 
 \[ f \in \mathcal{F}^{N_C}_\sigma  =  \underbrace{\mathcal{F}_\sigma \times  \mathcal{F}_\sigma \cdots \times  \mathcal{F}_\sigma}_{N_C}, \]
 where
\begin{equation*}\
  \begin{split}
    \mathcal{F}_\sigma = \{~g: g ~\text{is a measurable function with a range}~ [0, 1]~\}
  \end{split}
\end{equation*} 
Given the restriction, the scoring function should be expressed as:
\[f \in \arginf_{f \in \mathcal{F}^{N_C}_\sigma} R(f),\]

Following the convention of machine learning terminologies, we refer to such functions as Bayes optimal scoring functions. The following theorem gives the solution for Bayes optimal scoring functions when the data distribution is a known priori. 

\label{eq:optbay}
\begin{thm}[\textbf{Bayes Optimal Scoring Functions}]\label{prop:bay}
Given $\eta_i(\cdot) = \mathbb{P}[y = i|x]$, $p_i= \mathbb{P}[y = i]$,  we have the following consequences:
\begin{enumerate}
\item[(a)] $f= \{f^{(i)}\}_{i= 1,2,\cdots, N_C} \in \mathcal{F}^{N_C}_\sigma $ is a Bayes optimal scoring function under the $\mauc$ criterion, if:
\begin{equation}\label{eq:optbay}
\begin{split}
&\Delta(f^{(i)}) \cdot  \Delta(\pi) > 0,\\ 
&\forall  \x_1, \x_2, ~s.t.~\pione \neq \pitwo \\ 
\end{split}
\end{equation}
where:
\begin{equation*}
\begin{split}
&\Delta(f^{(i)}) = f^{(i)}(\bm{x}_1) - f^{(i)}(\bm{x}_2) \\
& \Delta(\pi) = \pione - \pitwo \\
&\pione =  \sum_{j \neq i} \frac{\eta_i(\bm{x}_1)\eta_j(\bm{x}_2)}{2p_ip_j}, \\
& \pitwo =  \sum_{j \neq i} \frac{\eta_j(\bm{x}_1)\eta_i(\bm{x}_2)}{2p_ip_j}.
\end{split}
\end{equation*}

\item[(b)] Define $\sigma(\cdot)$ as the sigmoid function, $s_i(\x)= \eta_i(\x)/p_i$ and $s_{\backslash i}(\x) = \sum_{j \neq i } s_j(\x)$,  then a Bayes optimal scoring function could be given by:
 \begin{equation}
 f^{\star(i)}(\x) =
 \begin{cases}
  \sigma\left(\dfrac{s_i(\x)}{s_{\backslash i}(\x)}\right), & 1>\eta_i(\x)\ge 0,\\ 
  1, & 1=\eta_i(\x),
 \end{cases}
 \end{equation}
\end{enumerate}

\end{thm}
On top of providing a solution for the optimal scoring function, this theorem also sheds light upon what the optimal scoring functions are really after. Thm.\ref{prop:bay} shows that the optimal scoring function provides a consistent ranking with $\left(\dfrac{s_i(\x)}{s_{\backslash i}(\x)}\right)$, which could be regarded as a generalized likelihood ratio of $y=i$ vs. $y\neq i$. Here, the posterior distribution $\mathbb{P}(y=i|\x)$ is weighted by the factor $1/p_i$, which eliminates the dependence on the label distribution since $\mathbb{P}(y=i|\x)/p_i$ is in proportion to the label-distribution-independent class-conditional distribution $\mathbb{P}(\x|y=i)$. This result suggests that the optimal scoring function induced by $\mauc$ is insensitive toward skewed label distribution.

\noindent \textbf{Surrogate Risk Minimization}.  Unfortunately, since the data distribution is not available  and the $0\text{-}1$ loss is not differentiable,  the Bayesian scoring functions are intractable even with the solution shown in Thm.\ref{prop:bay}. Practically,  we need to replace the $0\text{-}1$ loss  with a convex differentiable loss function $\ell$ to find tractable approximations. Once $\ell$ is fixed, we can naturally minimize the much simpler surrogate risk {formulated as $f^\star \in \argmin_{f} R_\ell(f)$}, where
\begin{equation*}
\begin{split}
&R_\ell(f) = \sum_{i} \frac{R^{(i)}_\ell(f^{(i)})}{N_C(N_C-1)}\\
&R^{(i)}_\ell(f^{(i)}) = \sum_{j\neq i} \ezz\left[{\ell(\Delta(\yi)\Delta{f}^{(i)})|\eij}\right].
\end{split}
\end{equation*}
 But how could we find out such surrogate losses at all? A potential candidate must be consistent with the 0-1 loss. In other words, one should make sure that Bayes optimal scoring functions be recovered by minimizing the chosen surrogate risk, at least in an asymptotic way. This leads to the following definition of consistency in a limiting sense.

\begin{defi}[\textbf{$\mauc$ Consistency}]\label{def:cons}\footnote[1]{This condition should be met for all distributions $\mathcal{D}_{\mathcal{X} \times \mathcal{Y}}.$}
$\ell$ is consistent with $\mauc$ if for
every function sequence $\{f_{t}\}_{t=1,2,\cdots}$, we have:
\begin{equation*}
R_\ell(f_{t}) \rightarrow \inf_{f \in \mathcal{F}^{N_C}_\sigma} R_\ell(f)~ \text{implies} ~ R(f_{t}) \rightarrow \inf_{f \in \mathcal{F}^{N_C}_\sigma} R(f).
\end{equation*} 
\end{defi}

\begin{rem}
  Note that since the infimum of $f$ is taken over all possible measurable functions with a proper range $[0,1]$, the result is thus irrelevant with the choice of the hypothesis space (linear model, Neural Networks).  Practically, to reach the theoretical infimum, it is better to consider complicated hypothesis spaces such as deep neural networks. 
\end{rem}

Based on this definition, we provide a sufficient condition for $\mauc$ consistency, which is shown in the following theorem. See Appendix \ref{sec:appb2} for the proof.
\begin{thm}[\textbf{$\mauc$ Consistency}] \label{thm:consq}
  The surrogate loss $\ell$ is consistent with $\mauc$ for all $f \in \mathcal{F}^{N_C}_\sigma$, if it is differentiable, convex, and nonincreasing within $[-1, 1]$ and $\ell'(0) < 0$.
\end{thm}
From the sufficient condition above, we can show that the popular loss functions are all consistent with $\mauc$, which is summarized as the following corollary.
\begin{col}
  The following statements hold according to Thm.\ref{thm:consq}:
  \begin{enumerate}
    \item Logit loss $\ell_{logit}(x) = \log(1+\exp(-x))$ is consistent with $\mauc$.
    \item  Exp loss $\ell_{\exp}(x) = \exp(-x)$ is consistent with $\mauc$.
    \item  Square loss $\ell_{sq}(x) = (1-x)^2$ is consistent with $\mauc$.
    \item The q-norm hinge loss $\ell_{q}(x) = \left(\max(1-x, 0)\right)^q$ is consistent with $\mauc$, if $q>1$.
    \item The generalized hinge loss given by:
    \begin{equation*}
      \ell_{\epsilon, m}(x) = \begin{cases}
        m-t, & t\le 1-\epsilon \\ 
        (t-1-\epsilon)^2/4\epsilon, & 1-\epsilon \le t < 1  \\ 
        0, & \text{otherwise}
      \end{cases}
    \end{equation*}
    is consistent with $\mauc$, if $1/2> \epsilon >0$.
  \item The distance-weighted loss given by:
  \begin{equation*}
    \ell_{d}(x) = \begin{cases}
      1/t, & t > \epsilon \\ 
      1/\epsilon \cdot \left( 2 - t/\epsilon \right), & \text{otherwise}
    \end{cases}
  \end{equation*}
  is consistent with $\mauc$, if $1>\epsilon>0$.
  \end{enumerate}
\end{col}

Note that $\ell_{hinge}(t) = \max(1-t,0) = \lim_{\epsilon \rightarrow 0} \ell_{\epsilon}(t)$. This shows that hinge loss is at least a limit point of the set of all consistent losses.



\section{Empirical Risk Minimization and Imbalance Aware Generalization Analysis}\label{sec:c}
Thus far, we have defined the suitable replacements for the 0-1 loss and found some popular examples for them. However, the arguments are based on the population version of the risk. Calculating the expectation is hardly available since the data distribution is most likely unknown to us. In this section, we take a step further to explore how to find empirical estimations of the risks given a specific training data set. On top of this, we show theoretical guarantees for such estimations.
\subsection{Empirical Surrogate Risk Minimization}
Given a finite training data $\mathcal{S}$ collected from the data distribution,  we start with an  unbiased estimation of the expected surrogate risk  $R_\ell(f)$ based on  $\mathcal{S} = \{\x_i,y_i\}_{i=1}^N$. To represent the label frequencies in $\mathcal{S}$, we denote $\Ni = \{\bm{x}_k: y_k = i, \bm{z}_k \in \mathcal{S} \}$ as the set of samples having a label $i$. We define $n_i = |\Ni|$ as the number of instances  belonging to the $i$-th class in $\mathcal{S}$.
\begin{prop}[\textbf{Unbiased Estimation}]\label{prop:ub} Define $\Rhatf$ as:
\[\Rhatf = \sumi \suminj \sumxone \sumxtwo \ninj \ell^{i,j,m,n} ,\]
where $\ell^{i,j,m,n}$ is a shorthand for $\ell(f^{(i)}(\x_m) - f^{(i)}(\x_n) )$.
 Then $\Rhatf$ is an unbiased estimation of ${R}_{\ell}(f)$, in the sense that:
${R}_{\ell}(f) = \expe\limits_{\mathcal{S}}(\Rhatf)$.
\end{prop}

According to the Empirical Risk Minimization (ERM) paradigm, we can turn to minimize $\Rhatf$ based on a finite training dataset $\mathcal{S}$ facing the unknown data distribution. Practically, the function $f$ here  is  often parameterized by a parameter $\theta$. In this way, we can express $f$ as $f_{\theta}$.  For example, linear models could be defined as $f_\theta(\x) =\theta^\top \x$. Moreover, to prevent the over-fitting issue, we restrict our choice of $f$ on a specific hypothesis class $\mathcal{H}$. Again, $\mathcal{H}$ is essentially a subset of parameterized functions $f_{\theta}$ where $\theta$ is restricted in a set $\Theta$. For example, $\mathcal{H}$ could be defined as all the linear models with $||\theta|| \le \gamma$.  With this hypothesis set, we can construct the following optimization problem to learn $f$ within $\mathcal{H}$: 
\begin{equation}
{  \color{org}(\bm{OP}_1) }~~~~  \min_{f\in\mathcal{H}} \Rhatf.
\end{equation}  
From the parameter perspective, ${  \color{org}(\bm{OP}_1) }$ could be reformulated equivalently by minimizing over $\theta \in \Theta$. Moreover, the constraint that $\theta \in \Theta$ could be reformulated as a regularization term $Reg_{\Theta}(\theta)$. Consequently, ${  \color{org}(\bm{OP}_1) }$ also enjoys an alternative form that could be used for optimization:
\begin{equation}
  {  \color{org}(\bm{OP}_2) }~~~~  \min_{\theta} \hat{R}_{\ell,\mathcal{S}}(f_\theta) + \alpha \cdot Reg_{\Theta}(\theta).
  \end{equation}
This allows the standard machine learning technologies to come into play. In this section, we will adopt the notations for mathematical convenience: $f$ and $\mathcal{H}$. Instead of explicitly defining $\theta$, we provide the parameterization in the definition of different hypothesis classes.

\subsection{$\mauc$ Rademacher Complexity and Its Properties.}
\noindent Our next step is then to investigate how well such approximations will perform. From one point, choosing a consistent surrogate loss $\ell$ ensures that minimizing the surrogate risk suffices to find the Bayes optimal scoring function. From another point, based on a small empirical risk, we also need to guarantee that \textit{the training performance generalizes well to the unseen samples such that the expectation $R_\ell(f)$ is  small}. To ensure the second point, given the assumption that $f$ is chosen from a hypothesis class $\mathcal{H}$, we will provide a Rademacher-Complexity-based worst-case analysis for the generalization ability in the remainder of this section. The key here is to ensure $R_\ell(f) \le \hat{R}_{\ell}(f) +\delta$ with high probability, where $\delta$ goes to zero when $N$ goes to infinity. This way, minimizing $\hat{R}_{\ell}(f)$ suffices to minimize $R_\ell(f)$. 

 Unlike traditional machine learning problems that only involve independent instance-wise losses, AUC's pair-wise formulation makes the generalization analysis difficult to be carried out. Specifically, the terms in MAUC formulation exert certain degrees of interdependency. This way, the standard symmetrization scheme \cite{mlfun,risk} is not available for our problem. For instance, the terms $\ell(f^{(i)}(\bm{x}_1) - f^{(i)}(\bm{x}_2))$ and $\ell(f^{(i)}(\bm{x}'_1)-f^{(i)}(\bm{x}'_2))$ are interdependent as long as $\bm{x}_1 = \bm{x}'_1$ or $\bm{x}_2 = \bm{x}'_2$. To address this issue,  we provide an extended form of Rademacher complexity for $\mauc$ losses, which is defined as follows.

\begin{defi}[\textbf{$\mauc$ Rademacher Complexity}]
The Empirical $\mauc$ Rademacher Complexity over a dataset $\mathcal{S} = \{(\bm{x}_i, y_i)\}_{i=1}^m$, and a hypothesis space $\mathcal{H}$ is defined as:
\begin{equation*}
\empsR = \expe_{\sigma}\left[ \supf  \empcompsim \right],
\end{equation*} 
where
\[T^{i,j,m,n} = \sigmaterm,\]
  for $i=1,2,\cdots, N_C$, $\sigmai_1, \cdots, \sigmai_{\ni}$ are i.i.d Rademacher random variables. The population version of the $\mauc$ Rademacher Complexity is defined as $\sR = \mathbb{E}_{\mathcal{S}}\left[\empsR\right]$.
\end{defi}
The most important property of the $\mauc$ Rademacher complexity is that its magnitude is directly related to the generalization upper bounds, which is shown in the following theorem. The proofs are shown in Appendix.\ref{app:gen_bound} in the supplementary materials.
\begin{thm}[\textbf{Abstract Generalization Bounds}]\label{thm:gen}
Given dataset $\mathcal{S} = \{(\bm{x}_i,y_i)\}_{i=1}^m$, where the instances are sampled independently, for all multiclass scoring functions $f \in \mathcal{H}$, if $Range\{\ell\}\subseteq[0,B]$,  then for any $\delta \in (0,1)$, the following inequality holds with probability at least $1 - \delta$:
\begin{equation*}
\begin{split}
R_\ell(f) \le& \js + C_1 \cdot \frac{\empsR}{N_C(N_C-1)}+ \\
 &C_2 \cdot \frac{B}{N_C}  \cdot \xi(\bm{Y}) \cdot \sqrt{\frac{\log(\nicefrac{2}{\delta})}{N}},
\end{split}
\end{equation*}
where $C_1, C_2$ are universal constants, $\xi(\bm{Y}) = \sqrt{\sum_{i=1}^{N_C}\nicefrac{1}{\rho_i}}, \rho_i =\frac{n_i}{N}.$
\end{thm}
According to Thm.\ref{thm:gen}, we can obtain a generalization bound as soon as we can obtain a proper upper bound on the empirical Rademacher complexity $\empsR$. In the remainder of this subsection, we will develop general techniques to derive the upper bound of $\empsR$ based on the notion of covering number and chaining. With the help of the next few theorems,  we can convert the upper bound of $\empsR$ to upper bounds of Rademacher complexities for much simpler model classes. These techniques are foundations for the practical results developed in the next subsection. Specifically, Thm.\ref{thm:chain}{{\color{blue}-(a)}} is used to prove Thm.\ref{thm:trans};  Thm.\ref{thm:trans} is employed in the second half of the proof of Thm.\ref{thm:gen_3}; and Thm.\ref{thm:chain}{{\color{blue}-(b)}} is employed in the proof of Thm.\ref{thm:conv}.

We are now ready to present the corresponding results. With the sub-Gaussian property proved in Lem.\ref{lem:pro} in Appendix \ref{app:gen_bound}, we can derive chaining upper bounds  for the $\mauc$ Rademacher complexity. The foundation here is the notion of covering number, which is elaborated in Def.\ref{def:ecover} and Def.\ref{def:covernum}.
\begin{defi}[$\epsilon$-covering]\label{def:ecover}\cite{tala} Let $(\mathcal{H}, d)$ be a (pseudo)metric space, and $\Theta \in \mathcal{H}$. $\{h_1,\cdots, h_K\}$ is said to be an  $\epsilon$-covering of $\Theta$ if $\Theta \in \bigcup_{i=1}^K\mathcal{B}(h_i,\epsilon)$, \emph{i.e.}, $\forall \theta \in \Theta$, $\exists i ~s.t.~ d(\theta, h_i) \le \epsilon$.
\end{defi}

\begin{defi}[Covering Number]\label{def:covernum}\cite{tala} Based on the notations in Def.\ref{def:ecover}, the covering number of $\Theta$ with radius $\epsilon$ is defined as:
\begin{equation*}
\mathfrak{C}(\epsilon,\Theta, d) = \min\left\{n: \exists \epsilon-\text{covering over $\Theta$ with size n}\right\}.
\end{equation*}
\end{defi}
Based on the definitions above, we can reach the following results. The proof is shown in Appendix \ref{app:gen_bound} in the supplementary materials.
Note that in the following arguments, the covering number is defined on the  metric $d_{\infty, \mathcal{S}}$. Specifically, given two vector valued functions $\s = (\s^{(1)}, \cdots, \s^{(N_C)})$, $\st = (\st^{(1)}, \cdots, \st^{(N_C)})$, and the training data $\mathcal{S}$, $d_{\infty, \mathcal{S}}(\s, \st)$  is defined as:
\begin{equation}
d_{\infty, \mathcal{S}}(\s, \st) = \max_{\bm{x}_i \in \mathcal{S}, j \in [N_C] } |\s^{(j)}(\bm{z}) - \st^{(j)}({\bm{z}})|.
\end{equation}

\begin{thm}[Chaining Bounds for $\mauc$ Rademacher Complexity]\label{thm:chain} Suppose that the score function $s^{(i)}$ maps $\mathcal{X}$ onto a bounded interval $[-R_s, R_s]$, the following properties hold for $\empsR$:
\begin{enumerate}
\item[(a)] For a decreasing precision sequence $\{\epsilon_k\}_{k=1}^K$, with $\epsilon_{k+1} = \frac{1}{2} \epsilon_{k},~ k=1,2,\cdots, K-1$ and $\epsilon_0 \ge R_s$, we have:
\begin{equation*}
\begin{split}
&\empsR \le N_C \cdot (N_C -1) \cdot \phil\cdot \epsilon_K\\
&~+ 6\cdot \sum_{k=1}^K \epsilon_k \phil(N_C-1) \cdot \xi(\bm{Y})\sqrt{\frac{\log(\mathfrak{C}(\epsilon_k,\mathcal{F}, d_{\infty, \mathcal{S}}))}{N}} 
\end{split}
\end{equation*}

\item[(b)] There exists a universal constant $C$, such that:
\begin{equation*}
\begin{split}
&\empsR \le C\phil \inf_{ R_s \ge \alpha \ge 0} \bigg( N_C(N_C-1)\alpha\\ 
&~~+ (N_C-1)\cdot \xi(\bm{Y}) \cdot \int_{\alpha}^{R_s} \sqrt{\frac{\log(\mathfrak{C}(\epsilon,\mathcal{F}, d_{\infty, \mathcal{S}}))}{N}} d\epsilon  \bigg)
\end{split}
\end{equation*}
\end{enumerate}
\end{thm}
According to Thm.\ref{thm:chain}, we have the following theorem as an extension of a new minorization technique which appears in a recent work \cite{optbound}. 
\begin{thm}[Transformation Upper Bound]\label{thm:trans} Given the Hypothesis class \[ \mathsf{soft} \circ \mathcal{F} = \bigg\{\bm{g}(\x) = \mathsf{soft}(\bm{s}(\x)):~ \s \in \mathcal{F}\bigg\}, \]  where $\mathsf{soft}(\cdot)$ is the softmax function. Suppose that $\s(x) \in [-R_s, R_s]^{N_C}$ and $\ell$ is $\phil$-Lipschitz continuous, the following inequality holds:
\begin{equation*}
\begin{split}
&\frac{\empsRF}{N_C(N_C-1)} \le \phil\bigg( 2^9 \cdot \frac{1}{N_C} \cdot \sqrt{N_C}\cdot\\
 &~~\xi(\bm{Y})\cdot \log^{3/2}\left(e \cdot R_s \cdot N \cdot N_C\right) \cdot \hat{\mathfrak{R}}_{N\cdot N_C}(\Pi \circ \mathcal{F})\\
&~~+ \sqrt{\frac{1}{N}}\bigg), 
\end{split}
\end{equation*}
where the Rademacher complexity  $\hat{\mathfrak{R}}_{N\cdot N_C}(\Pi \circ \mathcal{F})$ is defined as:
\begin{equation*}
\begin{split}
&\hat{\mathfrak{R}}_{N\cdot N_C}(\Pi \circ \mathcal{F})  \\
&~~= \expe_{\sigma}\left[\sup_{f = (f^{(1)}, \cdots, f^{(N_C)})\in \mathcal{F} } \frac{1}{N\cdot N_C} \sum_{j=1}^{N_C} \sum_{i=1}^N \sigma^{(i)}_j \cdot f^{(j)}(\x_i)\right]
\end{split}
\end{equation*}
where $\{\sigma^{(i)}_j\}_{(i,j)}$ is a sequence of independent Rademacher random variables.
\end{thm}
The result in Thm.\ref{thm:trans} directly relates the complicated pairwise Rademacher complexity $\empsRF$ to the instance-wise Rademacher complexity $\hat{\mathfrak{R}}_{N\cdot N_C}$. This makes the derivation of the generalization upper bound much easier.  Once a bound on the ordinary Rademacher complexity over the functional class $\mathcal{F}$ is available, we can directly plug it into this theorem and find a resulting bound over the $\mauc$ complexity.

\subsection{Generalization Bounds for Deep and Shallow Model Families}

Next, we derive the generalization bounds for three hypothesis classes:  (1) the $\ell_p$ penalized linear models, (2) deep neural networks with fully-connected layers, and (3) the deep convolutional neural networks. 

\begin{asm}[Common Assumptions]\label{asm:comm} In this subection, we require some common assumptions listed as follows:

\begin{itemize}
   \item The sample points in the training dataset $\mathcal{S} = \{(\bm{x}_i,y_i)\}_{i=1}^m$ are sampled independently
   \item  $Range\{\ell\} \subseteq [0, B]$
    \item The loss function $\ell$ is $\phil$-Lipschitz continuous
    \item The input features are sampled from $\mathcal{X} \subset \mathbb{R}^{d}$, and for all $\bm{x} \in \mathcal{X}$, we have $||\bm{x}||^2_2 \le R_{\mathcal{X}}$ .
\end{itemize}

  \end{asm}

\subsubsection{Generalization Bound for $\ell_p$ Penalized Linear Models}
We begin with  the generalization bound for $\ell_p$ norm penalized linear models. The proof is shown in Appendix \ref{sec:appch} in the supplementary materials.

\begin{thm}[\textbf{Practical Generalization Bounds for Linear Models}]\label{thm:gen_2}
 Define the $\ell_p$ norm penalized linear model as:
\begin{equation*}
\begin{split}
\mathcal{H}^{Lin}_{p,\gamma} = \big\{&f=(f^{(1)},\cdots,f^{(n_C)}): f^{(i)}(\x) = \bm{W}^{(i)}\bm{x},\\
 &||\bm{W}^{(i)}||_p \le \gamma \big\},
\end{split}
\end{equation*}
with $0<p<\infty$,  $\frac{1}{p} + \frac{1}{\bar{p}} =1$. Based on assumption \ref{asm:comm}, for all $f\in \Hlin$, we have the following inequality holds with probability at least $1-\delta$:
\begin{equation*}
{R}_{\ell}(f) \le \hat{R}_{\ell, \mathcal{S}}(f)  +  \mathcal{I}_{{Lin}}\bigg(\chi(\bm{Y}),\xi({\bm{Y}}),\delta\bigg) \cdot \sqrt{\frac{1}{N}},
\end{equation*}
where 
\begin{equation*}
\begin{split}
& \mathcal{I}_{{Lin}}\bigg(\chi(\bm{Y}),\xi({\bm{Y}}),\delta\bigg) =  \frac{4 R_{\mathcal{X}}\phil\gamma}{N_C-1} \cdot  \sqrt{\frac{2(\bar{p} -1)}{N_C}} \cdot \rchi(\bm{Y})  \\
&+  \dfrac{5B }{N_C } \cdot \sqrt{2\log\left(\dfrac{2}{\delta}\right)} \cdot \xi(\bm{Y}).
\end{split}
\end{equation*}
\end{thm}

\subsubsection{Generalization Bound for Deep Fully-Connected Neural Networks}
Next, we take a further step to explore the generalization ability of a specific type of deep neural networks where only fully-connected layers and activation functions exist. The detailed here is as follows.

\noindent \textbf{Settings}. We denote an $L$-layer deep fully-connected neural network with $N_C$-way output as :
\[ f(\x) =  \bm{W} f_{\bm{\omega},L}(\bm{x}) =\bm{W} s(\bm{\omega}_{L-2}\cdots s(\bm{\omega}_1\bm{x})),\]
\noindent The notations are as follows: $s(\cdot)$ is the activation function; $n_{h_j}$ is the number of hidden neurons for the $j$-th layer;  $\bm{\omega}_{j} \in \mathbb{R}^{n_{h_{j+1}} \times n_{h_{j}}}, j = 1,2,\cdots, L-2$ are the weights for the first $L-1$ layers; $\bm{W} \in \mathbb{R}^{n_{h_{L-1}}\times N_C }$ is the weight for the output layer. Moreover, the output from the $i$-th layer of the network is defined as :
  \[ f^{(i)}(\x) =  \bm{W}^{(i)^\top} f_{\bm{\omega},L}(\bm{x}),\]
  \noindent where $\bm{W}^{(i)^\top}$ is the $i$-th row of $\bm{W}$.  In the next theorem, we focus on a specific hypothesis class for such networks where the product of weight norms 
  \[\Pi_{\bm{W},\bm{\omega}} =  ||\bm{W}||_F \cdot\prod_{j=1}^{L-2}||\bm{\omega}_j||_F \] 
\noindent are no more than $\gamma$. We denote such a hypothesis class as $\Hdnn$:
  \begin{equation*}
    \begin{split}
    \Hdnn =\bigg\{&  f: f^{(i)}(\bm{x}) = \bm{W}^{(i)^\top} f_{\bm{\omega},L}(\bm{x}), ||f^{(i)}||_\infty \le R_s,\\ 
  &i =1,\cdots,N_C, ~ \Pi_{\bm{W},\bm{\omega}} \le \gamma\bigg\},
    \end{split}
    \end{equation*}
    To obtain the final output, we perform a softmax operation over $f(\x)$. Thus, the valid model under this setting could be chosen  from the following hypothesis class:
    \begin{equation*}
      \begin{split}
      \mathsf{soft} \circ \Hdnn =\bigg\{  g:& g^{(i)} = \dfrac{\exp(f^{(i)}(\x))}{\sum_{j=1}^{N_C} \exp(f^{(j)}(\x))}, \\
      &f \in \Hdnn  \bigg \}.
      \end{split}
      \end{equation*}

      We have the following bound for this type of models. The result here is a merge of two independent results we proposed in Appendix \ref{app:gen_bound} and Appendix \ref{sec:appch} in the supplementary material, which is based on the Talagrand contraction properties and the  chaining technology  shown in Thm.\ref{thm:chain} and Thm.\ref{thm:trans}.
\begin{thm}[\textbf{Practical Generalization Bounds for Deep Models}]\label{thm:gen_3}
On top  of assumption \ref{asm:comm}, if we further assume that $s(\cdot)$ is a 1-Lipshiptz and positive homogeneous activation function,  then for all $ f \in \mathsf{soft} \circ \Hdnn$, we have the following inequality holds with probability at least $1-\delta$:
\begin{equation*}
R_\ell(f) \le \js  + \min\bigg(\mathcal{I}_{{DNN,1}},~  \mathcal{I}_{{DNN,2}}\bigg) \cdot \sqrt{\frac{1}{N}}
\end{equation*}
where $\rchi(\bm{Y}) =\sqrt{\sumi \suminj \nicefrac{1}{\rho_i\rho_j}},~~ \xi(\bm{Y}) = \sqrt{\sumi \nicefrac{1}{\rho_i}}
, ~~\rho_i = \dfrac{\ni}{N},$
\begin{equation*}
\begin{split}
\mathcal{I}_{{DNN,1}} =&~  C_1\frac{\sqrt{2}}{2} \phil \cdot \frac{\rchi(\bm{Y})}{N_C-1} + \bigg( \frac{\sqrt{2}C_1\Rx\phil\gamma}{2}  \cdot C_3 \\ 
&~+  \dfrac{C_2B }{{N_C}} \cdot \sqrt{2\log\left(\dfrac{2}{\delta}\right)}\bigg) \cdot \xi(\bm{Y}),
\end{split}
\end{equation*}
\begin{equation*}
\begin{split}
\mathcal{I}_{{DNN,2}}  =&~ C_1\phil\bigg( \frac{2^9}{N_C} \cdot \xi(\bm{Y})\cdot \log^{3/2}\left(K\cdot N \cdot N_C\right)\\  
&~ \gamma \cdot \Rx \cdot (\sqrt{2\log(2)L}+1)+ 1\bigg) \\ 
&~+ C_2 \frac{ B\cdot\sqrt{{\log(\nicefrac{2}{\delta})}}\cdot \xi(\bm{Y})}{N_C},
\end{split}
\end{equation*}
$C_1$, $C_2$ are universal constants as thm.\ref{thm:gen}, $K = e\cdot R_s$, $C_3 =\frac{\sqrt{L\log2} + \sqrt{N_C}}{\sqrt{N_C-1}}$.
\end{thm}
\subsubsection{Generalization Bound for Deep Convolutional Neural Networks}
Now we use the result in Thm.\ref{thm:chain} to derive a generalization bound for a  class of deep neural networks where fully-connected layers and convolutional layers coexist. In a nutshell, the result is essentially an application of Thm.\ref{thm:chain} to a recent idea appeared in \cite{convbound}. \\
\textbf{Settings.} Now we are ready to introduce the setting of the deep neural networks employed in the forthcoming theoretical analysis, which is adopted from \cite{convbound}. We focus on the deep neural networks with $N_{conn}$ fully-connected layers and $N_{conv}$ convolutional layers. The $i$-th convolutional layer has a kernel $\bm{K}^{(i)} \in \mathbb{R}^{k_i \times k_i \times c_{i-1} \times c_i}$. Recall that convolution is a linear operator. For a given kernel $\bm{K}$, we denote its associated matrix as $op(\bm{K})$, such that $\bm{K}(\bm{x}) = op(\bm{K}) \bm{x}$. Moreover, we assume that, each time, the convolution layer is followed by a componentwise non-linear activation function and an optional pooling operation. We assume that the activation functions and the pooling operations are all $1$-Lipschitz. For the $i$-th fully-connected layer,  we denote its weight as $\bm{V}^{(i)}$. Above all, the complete parameter set of a given deep neural network could be represented as $\bm{P}= \{\bm{K}^{(1)},\cdots, \bm{K}^{(N_{conv})}, \bm{V}^{(1)}, \cdots, \bm{V}^{(N_{conn})}\}$. Again, we also assume that the loss function is $\phil$-Lipschitz and $Range\{\ell\} \subseteq [0,B]$. Finally, given two deep neural networks with parameters $\bm{P}$ and $\bm{\tilde{P}}$, we adopt a metric $d_{NN}(\cdot,\cdot)$ to measure their distance:
\begin{equation*}
\begin{split}
&d_{NN}(\bm{P},\bm{\tilde{P}}) = \sum_{i=1}^{N_{conv}} ||op(\bm{K}^{(i)}) -op(\bm{\tilde{K}}^{(i)})||_2 \\
&+ \sum_{i=1}^{N_{conn}} ||\bm{V}^{(i)} -\bm{\tilde{V}}^{(i)}||_2.
\end{split}
\end{equation*} \\
\textbf{Constraints Over the Parameters}. First, we define  $\mathcal{P}^{(0)}_{\nu}$ as the class for \underline{initialization of the parameters}:
\begin{equation*}
\begin{split}
\mathcal{P}^{(0)}_{\nu} &= \bigg\{\bm{P}: \left(\max_{i\in \{1,\cdots, N_{conv}\}}||op(\bm{K}^{(i)})||_2 \right)\\
&\le 1+\nu, ~ \left(\max_{j \in \{1,\cdots, N_{conn}\}} ||\bm{V}^{(j)}||_2\right) \le 1+\nu \bigg\}.
\end{split}
\end{equation*}
Now we further assume that the learned parameters should be chosen from a class denoted by $\mathcal{P}_{\beta, \nu}$, where the distance between the learned parameter and the fixed initialization residing in $\mathcal{P}^{(0)}_{\nu}$ is no bigger than $\beta$:
\begin{equation*}
\mathcal{P}_{\beta, \nu} = \bigg\{\bm{P}: d_{NN}(\bm{P},\bm{\tilde{P}}_0) \le \beta, ~\bm{\tilde{P}}_0 \in \mathcal{P}^{(0)}_\nu \bigg\}.
\end{equation*}
We have the following result for this class of deep neural networks, which is an application of Thm.\ref{thm:chain}. The proof is shown in Appendix \ref{sec:appch} in the supplementary materials.
\begin{thm}\label{thm:conv} Denote the hypothesis class,
\[ \mathsf{soft} \circ \mathcal{F}_{\beta,\nu} = \bigg\{\bm{g}(\x) = \mathsf{soft}(\bm{s}_{\bm{P}}(\x)):~ \bm{s}_{\bm{P}} \in\mathcal{F}_{\beta,\nu}\bigg\}, \]
\begin{equation*}
\begin{split}
\mathcal{F}_{\beta,\nu} = \{&s_{\bm{P}}:\mathbb{R}^{N_{N_L-1}} \rightarrow \mathbb{R}^{N_C}| ~ \bm{P} \in \mathcal{P}_{\beta, \nu},\\
&Range(\bm{s}_P) \subseteq [-R_s, R_s]^{N_C} \}.
\end{split}
\end{equation*}
On top of Assumption \ref{asm:comm}, if we further assume that  
\begin{equation*}
\begin{split}
R_s >&1/ \min\bigg\{\sqrt{{N}} ,\\
&\frac{\xi(\bm{Y})}{N_C} \cdot \sqrt{N_{par}\left(\nu N_L + \beta + \log (3\Rx\cdot \beta \cdot {N}) \right)}\bigg\},
\end{split}
\end{equation*}
then for all multiclass scoring functions $f \in \mathsf{soft} \circ \mathcal{F}_{\beta,\nu}$,  $\forall \delta \in (0,1)$, the following inequalities hold with probability at least $1 - \delta$:
\begin{equation*}
\begin{split}
R_\ell(f) \le& \js + C_1  \alpha_1  \alpha_2 \\
&+ C_2 \cdot \frac{B\xi(\bm{Y})}{N_C}   \cdot \sqrt{\frac{\log(\nicefrac{2}{\delta})}{{N}}},
\end{split}
\end{equation*}
where
\begin{align*}
  \alpha_1 &=  \tilde{C} \cdot\frac{\phil  \cdot R_s \cdot \xi(\bm{Y})}{N_C},\\ 
  \alpha_2 &= \sqrt{\frac{N_{par}\left(\nu N_{L} + \beta +\log \left(3\beta{\Rx{N}}\right)\right)}{{N}}},\\
\end{align*}
$\tilde{C}$, $C_1, C_2$ are universal constants, $N_L = N_{conv} + N_{conn}$, $N_{par}$ is total number of parameters in the neural network.
\end{thm}
\subsection{Summary}
Note that there are two imbalance-aware factors that appear in the  upper bounds above. The first factor  $\xi(\bm{Y})$ captures the imbalance of class label distribution in $\mathcal{S}$. With an analogous spirit,  $\rchi(\bm{Y})$ captures  the degree of imbalance of the class pair distribution in $\mathcal{S}$.  
To improve generalization, the training data $\mathcal{S}$ must simultaneously have a large sample size $N$ and a small degree of imbalance captured by $(\xi({\bm{Y}}), \rchi(\bm{Y}))$. In other words, our bounds suggest that blindly increasing the sample size of the training dataset will not improve generalization. To really improve generalization, one should increase data points of the minority classes which are the sources for performance bottleneck. Consequently, compared with the ordinary $O(\sqrt{1/N})$ result, our bound is much more aware of the imbalance issue hidden behind the training data.

\section{Efficient Computations in Optimization}\label{sec:d}
\begin{table*}[htbp]
  \centering \small
  \caption{Acceleration for three losses \label{tab:acc}, where  $\bar{N} = \sumi \ni \log \ni + (N-\ni) \log (N-\ni)$   }
    \begin{tabular}{llll}
      Algorithms    & \multicolumn{1}{l}{loss} & \multicolumn{1}{l}{gradient} & \multicolumn{1}{l}{requirement}\\
    \toprule
    exp + acceleration   &   $O(N_C \cdot N \cdot T_\ell)$      & $O(N_C \cdot N \cdot T_{grad})$ & $\min_i \ni \gg 2$  \\
    squared + acceleration & $O(N_C \cdot N \cdot T_\ell)$     & $O(N_C \cdot N \cdot T_{grad})$ & $e^{\frac{1}{2} (N-\ni)} \gg  \ni \gg N - e^{\frac{1}{2}\ni}$ \\
    hinge + acceleration &   $O(N_C\cdot \bar{N} \cdot  T_\ell)$      & $O(N_C \cdot \bar{N} \cdot T_{grad})$ & $\min_i \ni \gg 2$\\
    \midrule
    w/o  acceleration &    $O(\sumi \suminj n_in_j\cdot T_\ell)$   & $O(\sumi \suminj n_in_j\cdot T_{grad})$ & $\backslash$ \\
    \bottomrule
    \end{tabular}%
\end{table*}%
Till now, we have conducted a series of theoretical analyses for our framework. In this section, we turn our focus to the practical issues we must face during optimization. As shown in the previous sections, the complicated formulations of $\mauc$ surrogate losses bring a great burden to the fundamental operations of the downstream optimization algorithm. Specifically, given $T_\ell$ as the time complexity for loss evaluation of a single sample pair and given $T_{grad}$ as that for the gradient evaluation, we can find that even a single full(mini)-batch loss and gradient evaluation takes $O(\sumi \suminj n_in_jT_\ell)$ and  $O(\sumi \suminj n_in_jT_{grad})$, which scales almost quadratically to the sample (batch) size.  In general, such complexities are unaffordable facing medium and large-scale datasets. However, we find that for some of the popular surrogate losses, the pairwise computation could be largely simplified. Consequently, we propose acceleration algorithms for loss and gradient evaluations for three well-known losses: exponential loss, squared loss, and hinge loss. The time complexity with/without acceleration is shown in Tab.\ref{tab:acc}, which shows the effectiveness of our proposed algorithms.

Generally, the empirical surrogate risk functions $\hat{R}_{\ell} $ have the following abstract form:
\begin{equation*}
\hat{R}_{\ell} = \sumi \suminj \sumxone \sumxtwo \ninj \cdot   \ell^{i,j},
\end{equation*}
where:
\[ \ell^{i,j} = \ell\left(f^{(i)}(\x_m) - f^{(i)}(\x_n)\right) ,\]
\[\fii = g_i(
  \bm{W}\hth),\]
\[\bm{W} = [\bm{w}^{(1)},\cdots, \bm{w}^{(N_C)}]^\top.\]

We adopt this general form since it covers a lot of popular models. For example, when $g_i(\cdot)$ is defined as the activation function of the last layer of a neural network (say a softmax function), $\bm{w}^{(i)}$ are the weights of the last layer, and $\hth[\cdot]$ is the neural net where the last layer is excluded, $\fii$ becomes a deep neural network architecture with the last layer designed as a fully-connected layer. As an another instance, if both $g_i(\x)= \x$ and $\hth =\x$, then we reach a simple linear multiclass scoring function. Note that the scalability with respect to sample size only depends on the choice of $\ell$. The choice of $g_i(\cdot)$ and  $\hth[\cdot]$ only affects the instance-wise chain rule. This allows us to provide a general acceleration framework once the surrogate loss is fixed.

\textit{Due to the limited space of this paper, we only provide loss evaluation acceleration methods in the main paper. The readers are referred to  Appendix \ref{sec:appe} for more details, where we also present a discussion to deal with the acceleration for general loss functions.}

\subsection{Exponential Loss}
For the exponential loss, we can simplify the computations by a factorization scheme:
\begin{equation*}
\begin{split}
\hat{R}_{exp} &= \sumi\sumxone  \suminj \sumxtwo \ninj \cdot  \exp\bigg(\alpha \cdot \left(f^{i,j,m,n} \right)\bigg),\\
 = &\sumi\underbrace{\left(\sumxone \exp(\alpha\cdot \fixm)\right)}_{(a_i)} \cdot\\
 &\underbrace{\left(\suminj \sumxtwo \ninj \cdot  \exp(-\alpha\cdot\fixn)\right)}_{(b_i)},
\end{split}
\end{equation*}
where $f^{i,j,m,n} = \dfi.$
From the derivation above, the loss evaluation could be done by first calculating $(a_i),(b_i)$ separately and then performing the multiplication, which only takes $O(NN_CT_\ell)$.  This is a significant improvement compared with the original $O(\sumi\suminj n_in_j N_C T_\ell)$ result. 
\subsection{Hinge loss}
First we put down the hinge surrogate loss as:
\begin{equation*}
\hat{R}_{hinge} = \sumi\sumxone  \suminj \sumxtwo \ninj \cdot  \left(\alpha - \left(f^{i,j,m,n}\right)\right)_+. 
\end{equation*}
The key of our acceleration is to notice that the terms are non-zero only if $\left(\fixm - \fixn\right) \le \alpha$. Moreover, for these non-zero terms $\max(x,0) =x$. This means that the hinge loss degenerates to an identity function for the activated nonzero terms, which enjoys efficient computation. So the key step in our algorithm is to find out the non-zero terms in an efficient manner.

Given a fixed class $i$ and an instance $\x_m \in \Ni$, we denote: 
\[\mathcal{A}^{(i)}(\x_m) =\{\x_n \notin \Ni, \alpha > \fixm -\fixn \}.\]
With $\mathcal{A}^{(i)}(\x_m)$, one can reformulate $\hat{R}_{hinge}$ as:
 \begin{equation*}
 \small
 \begin{split}
 \hat{R}_{hinge} &= \sumi\sumxone \bigg[\left(\suminj \sum_{\x_n \in \Nj \cap \mathcal{A}^{(i)}(\x_m) } \ninj\right) \cdot\\
 &  \left(\alpha - \fixm \right) + \suminj\sum_{\x_n \in \Nj \cap \mathcal{A}^{(i)}(\x_m) } \ninj \cdot \fixn  \bigg],\\
 & =  \sumi\sumxone \left[\delta^{(i)}(\x_m) \cdot (\alpha - \fii[\x_m])+ \Delta^{(i)}(\x_m)\right],
 \end{split}
\end{equation*}
where 
\begin{equation}
\begin{split}
\delta^{(i)}(\x_m) &= \suminj \sum_{\x_n \in \Nj \cap \mathcal{A}^{(i)}(\x_m) } \ninj,\\
 \Delta^{(i)}(\x_m)&= \suminj \sum_{\x_n \in \Nj \cap \mathcal{A}^{(i)}(\x_m) }  \ninj \fixn.
\end{split}
\end{equation}

According to this reformulation, once $\delta^{(i)}(\x)$ and $\Delta^{(i)}(\x)$ are all fixed, we can calculate the loss function within $O(N)$. So the efficiency bottleneck comes from the calculations of $\delta^{(i)}(\x)$ and $\Delta^{(i)}(\x)$.

Next we propose an efficient way to calculate  $\delta^{(i)}(\x_m), \Delta^{(i)}(\x_m), \mathcal{A}^{(i)}(\x_m)$.  To do so, given a specific class $i$, we first sort the relevant (resp. irrelevant) instances descendingly according to their scores $\fii$:
\begin{equation*}
\begin{split}
&\fii[\bm{x}^\downarrow_0] \ge  \fii[\bm{x}^\downarrow_1] \cdots, \ge \fii[\bm{x}^\downarrow_{n_i-1}],~~~~  \bm{x}^\downarrow_i \in \mathcal{N}_i,  \\
&\fii[\bm{\tilde{x}}^\downarrow_0] \ge  \fii[\bm{\tilde{x}}^\downarrow_1] \cdots, \ge \fii[\bm{\tilde{x}}^\downarrow_{N-n_i-1}],~~~~ \bm{\tilde{x}}^\downarrow_i \notin \mathcal{N}_i.
\end{split}
\end{equation*}
\noindent It immediately follows that:
\begin{equation*}
\mathcal{A}^{(i)}(\bm{x}^\downarrow_0) \subseteq \mathcal{A}^{(i)}(\bm{x}^\downarrow_1) \subseteq \cdots \subseteq \mathcal{A}^{(i)}(\bm{x}^\downarrow_{\ni}). 
\end{equation*}
This allows us to find out all $\mathcal{A}^{(i)}(\bm{x}^\downarrow_k), k=1,2,\cdots,n_i,$ within a single pass of the whole dataset with an efficient dynamic programming.

Based on the construction of $\mathcal{A}^{(i)}_k = \mathcal{A}^{(i)}(\bm{x}^\downarrow_k)$, we provide the following recursive rules to obtain $\delta^{(i)}(\x)$ and $\Delta^{(i)}(\x)$:
\begin{equation*}
\begin{split}
&\delta^{(i)}(\x^\downarrow_{k+1}) = \delta^{(i)}(\x^\downarrow_{k}) +  \suminj \sum_{ \Nj \cap  \mathcal{A}^{(i)}_{k+1} \backslash  \mathcal{A}^{(i)}_k } \ninj, \\
&\Delta^{(i)}(\x^\downarrow_{k+1}) = \Delta^{(i)}(\x^\downarrow_{k}) +\\
&~~~~~~~~~~~~~~~~  \suminj \sum_{\x_n \in \Nj \cap \mathcal{A}^{(i)}_{k+1} \backslash  \mathcal{A}^{(i)}_k  } \ninj \fixn.
\end{split}
\end{equation*}
These recursive rules induce an efficient dynamic programming which only requires a single pass of the whole dataset. Putting all together, we then come to a $O(N_C\cdot \bar{N} \cdot T_{l})$ time complexity speed-up for hinge loss evaluation. \textit{An implementation of this idea is detailed in the Appendix \ref{sec:appe}.} 
\subsection{Squared Loss}

For each fixed class $i$, we construct an affinity matrix $\boldsymbol{Aff}^{(i)}$ such that 
\begin{equation*}
\boldsymbol{Aff}^{(i)}_{m,n} = 
\begin{cases}
\dfrac{1}{n_in_{y_m}}, & n \in \Ni, m \notin \Ni, \\
\dfrac{1}{n_in_{y_n}}, & m \in \Ni, n \notin \Ni, \\
0, & \text{Otherwise},
\end{cases}
\end{equation*}
which could be written in a matrix form,
    \begin{equation*}
     \boldsymbol{Aff}^{(i)} = \bm{D}^{(i)}({{\mathbf{1}}}-{{\mathbf{Y}}^{(i)}}){{\mathbf{Y}}^{(i)}}^{\top 
     }+{{\mathbf{Y}}^{(i)}}{{({{\mathbf{1}}}-{{\mathbf{Y}}^{(i)}})}^{\top }}\bm{D}^{(i)},
    \end{equation*}
 Then the empirical risk function under the squared surrogate loss could be reformulated as: 
   \begin{equation*}\label{wlseq}
   \hat{R}_{sq}   = \sumi   \bm{\Delta}^{(i)^\top}_{sq}
      \mathcal{L}^{(i)}\bm{\Delta}^{(i)}_{sq},
   \end{equation*} 
   where
   \begin{equation}
   \begin{split}
     \bm{\Delta}^{(i)}_{sq} &= \bm{Y}^{(i)} - \fii[\bm{X}],\\
      \fii[\bm{X}] &= [\fii[\x_1],\cdots,\fii[\x_N] ]^\top,
   \end{split}
   \end{equation}
,$\mathcal{L}^{(i)} = diag(\bm{Aff}^{(i)}\bm{1}) - \bm{Aff}^{(i)}$ is the graph Laplacian matrix associated with $\bm{Aff}^{(i)}$. Based on this reformulation, one can find that the structure of $\mathcal{L}^{(i)}$ provides an efficient factorization scheme to speed-up loss evaluation. To see this,  we have:
   \begin{equation*}\label{wlseq}
   \hat{R}_{sq}   = \sumi   \frac{1}{2}\bm{\Delta}^{(i)\top}_{sq} (\bm{\kappa}^{(i)}\odot\bm{\Delta}^{(i)}_{sq}) -{\Delta}^{(i)}_1\cdot {\Delta}^{(i)}_2,
   \end{equation*} 
where
\begin{equation*}
\begin{split}
&{\Delta}^{(i)}_2 = \Yit \bm{\Delta}^{(i)}_{sq} \\
&{\Delta}^{(i)}_1 = \bm{\Delta}^{(i)^\top}_{sq}\Di (\Yni)\\
&\bm{\kappa}^{(i)} = \ni \Di (\Yni) + \dfrac{N_C-1}{\ni} \Yi.
\end{split}
\end{equation*}

\indent Taking all together, we see that the accelerated evaluation requires only $O(N \cdot N_C \cdot T_l)$ time.

\section{Experiments}\label{sec:exp}
\begin{figure*}[h]  

  \subfigure[Balance]{
  
    \includegraphics[width=0.315\textwidth]{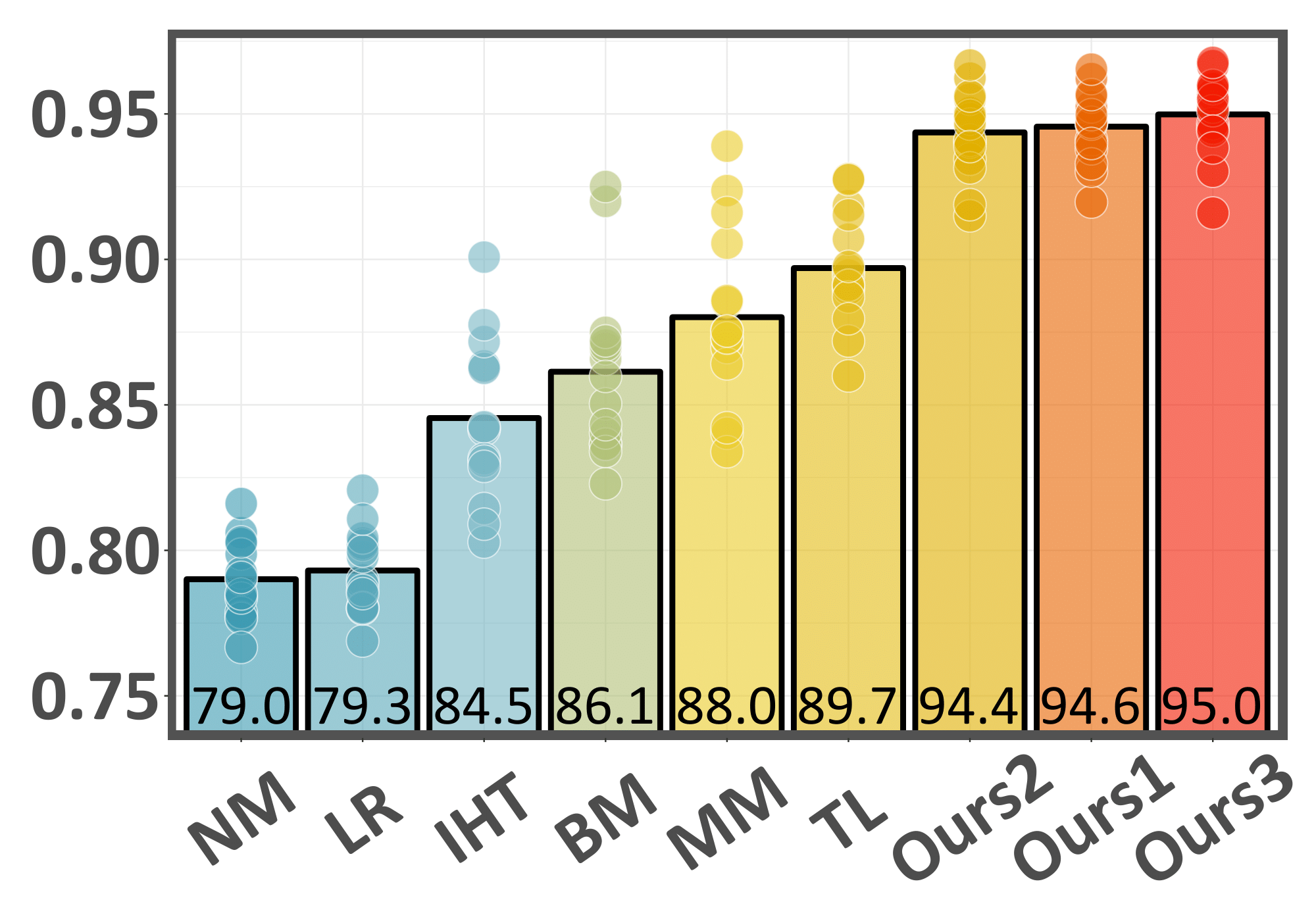} 
  }
  \subfigure[Dermatology]{
    \includegraphics[width=0.315\textwidth]{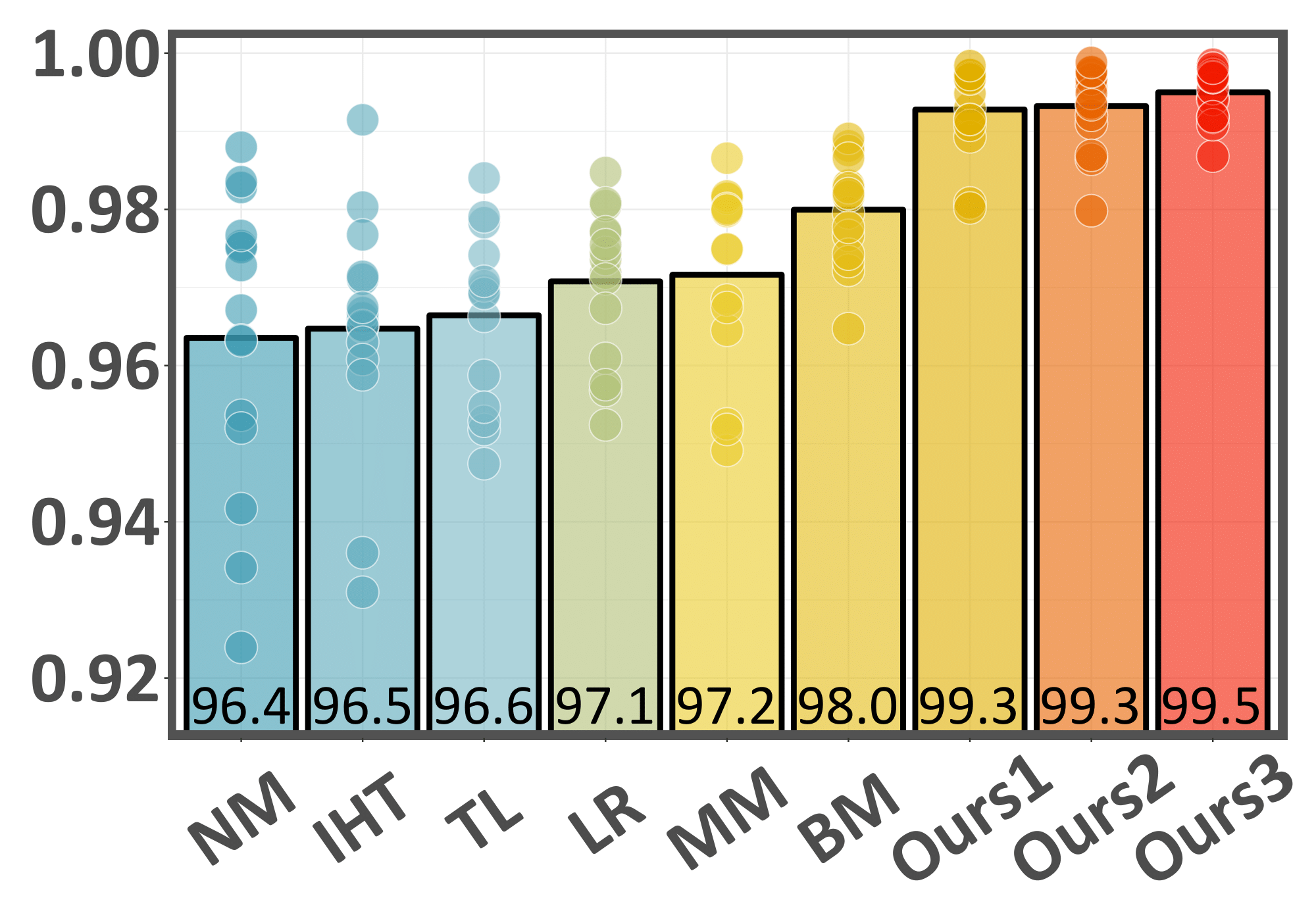} 
  }
  \subfigure[Ecoli]{
    \includegraphics[width=0.315\textwidth]{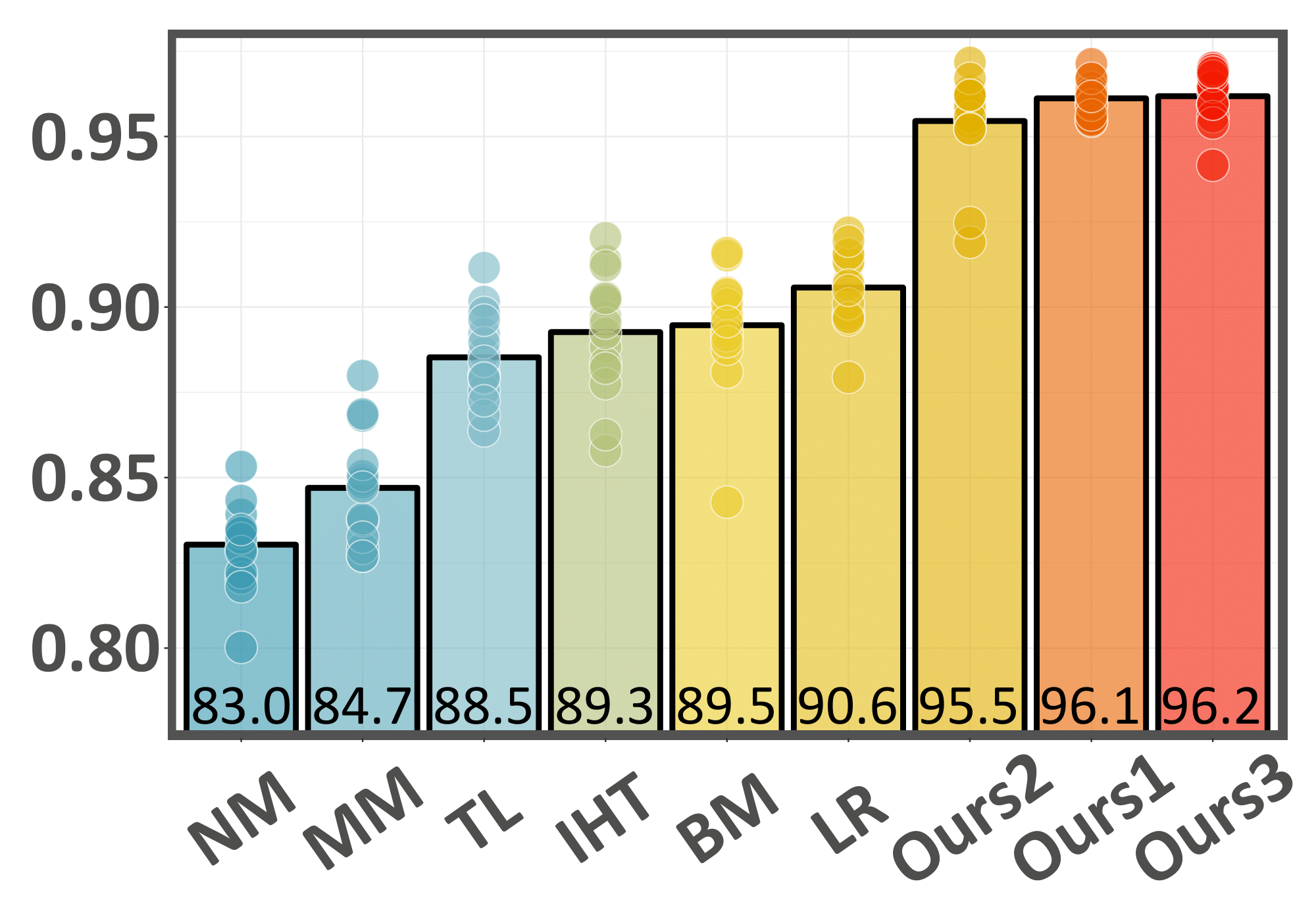} 
  
  }
  
  \subfigure[New Thyroid]{
    \includegraphics[width=0.315\textwidth]{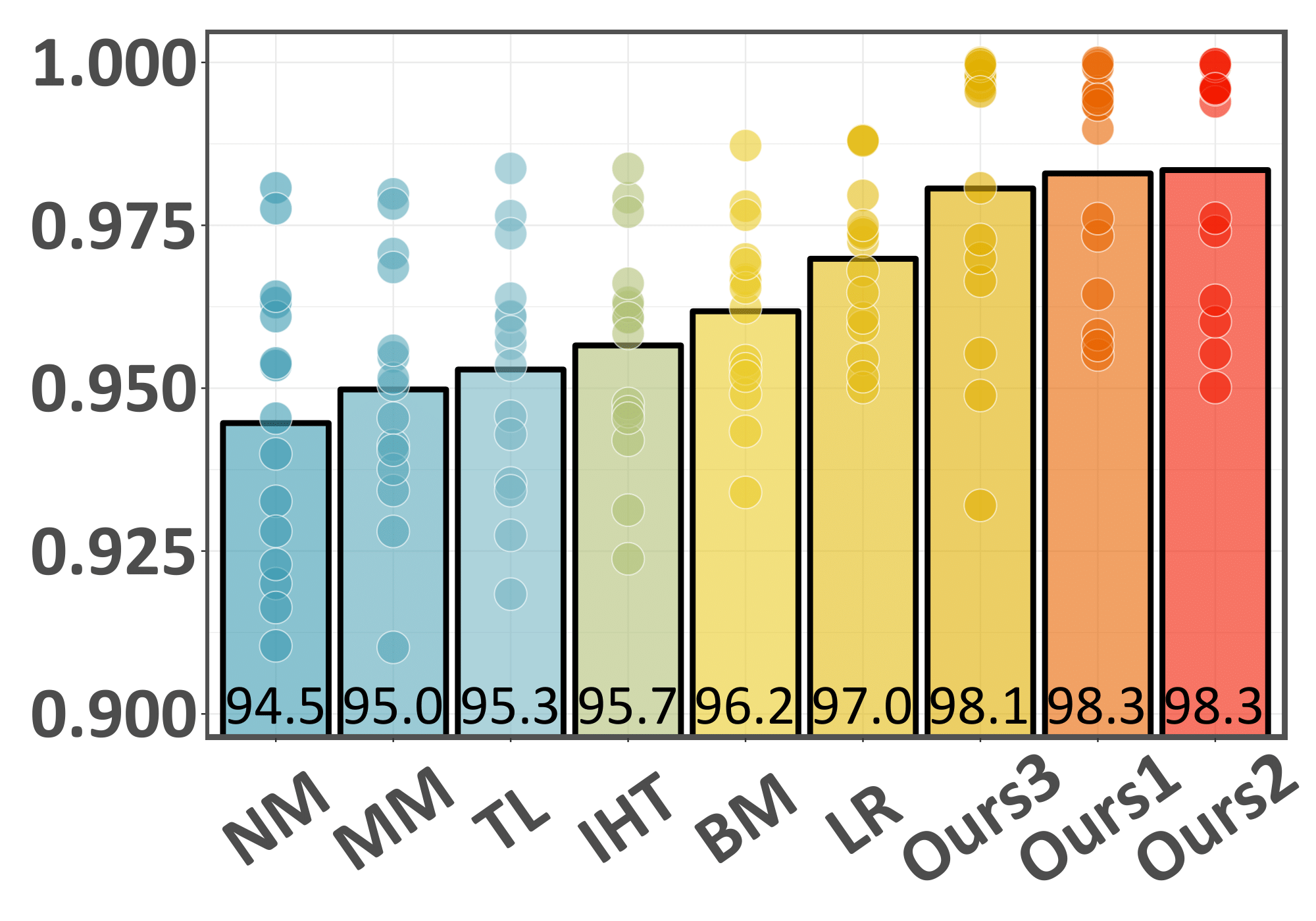} 
  }
  \subfigure[Page Blocks]{
    \includegraphics[width=0.315\textwidth]{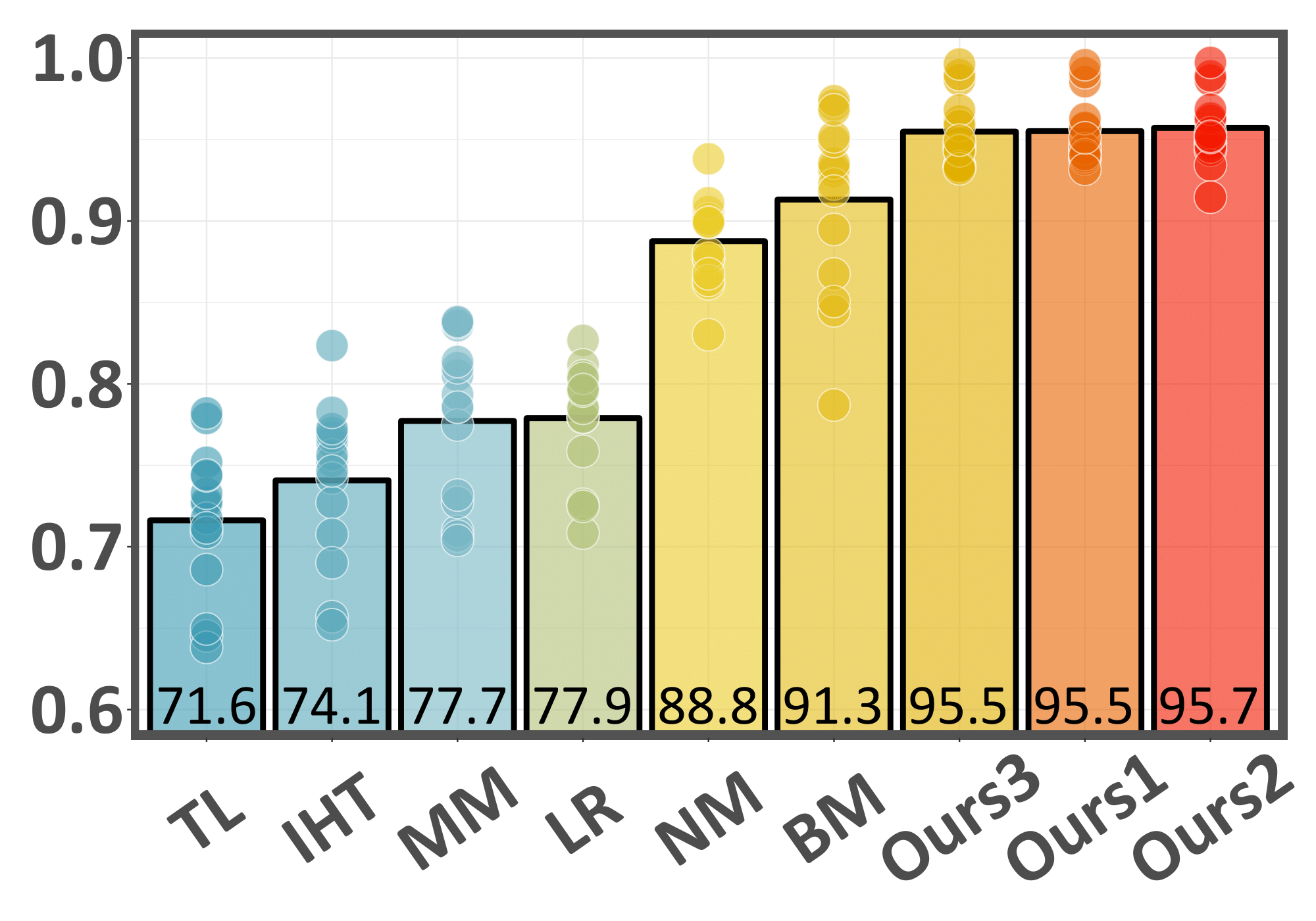} 
  }
  \subfigure[SegmentImb]{
    \includegraphics[width=0.315\textwidth]{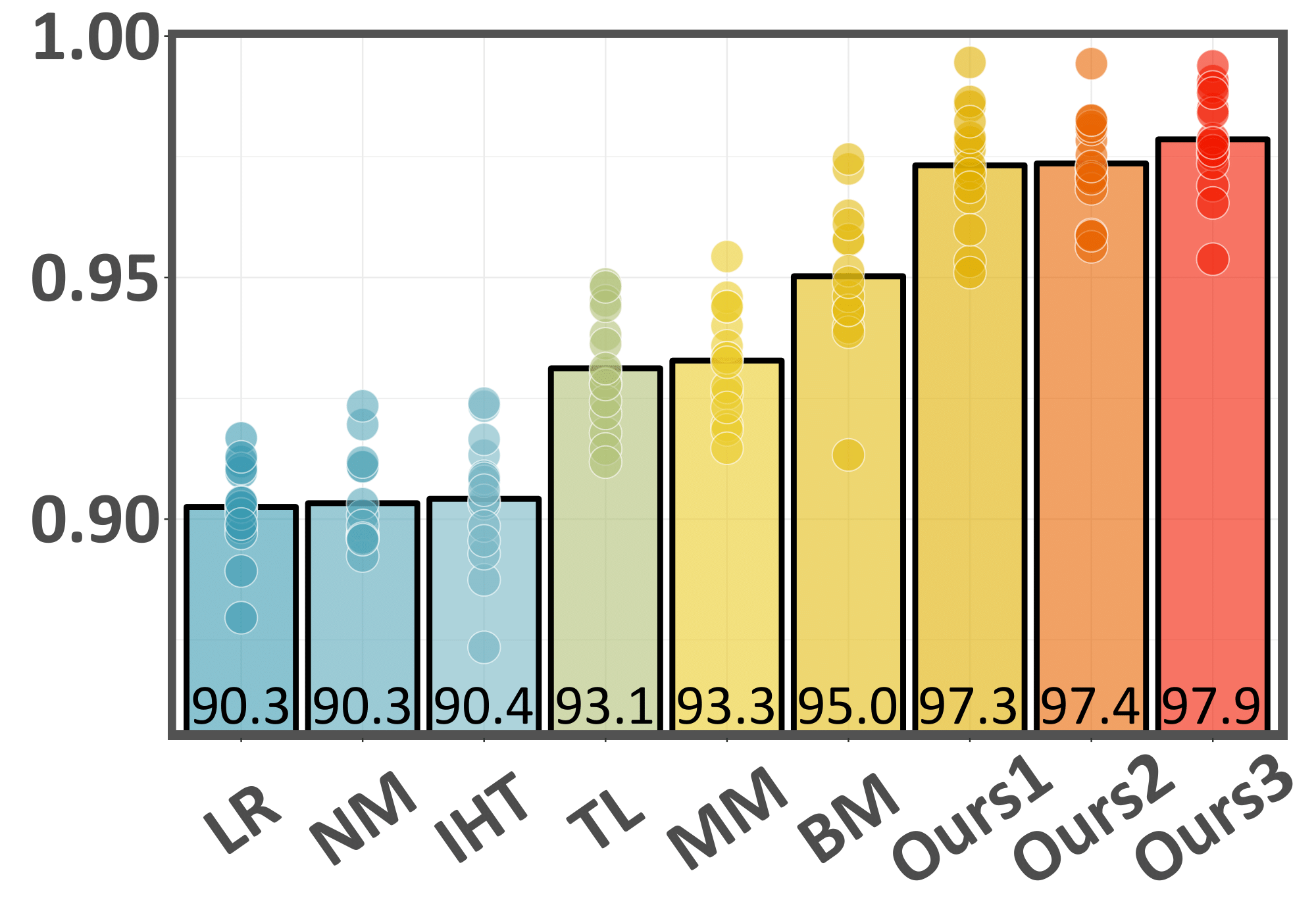} 
  }
  
  \subfigure[Shuttle]{
    \includegraphics[width=0.315\textwidth]{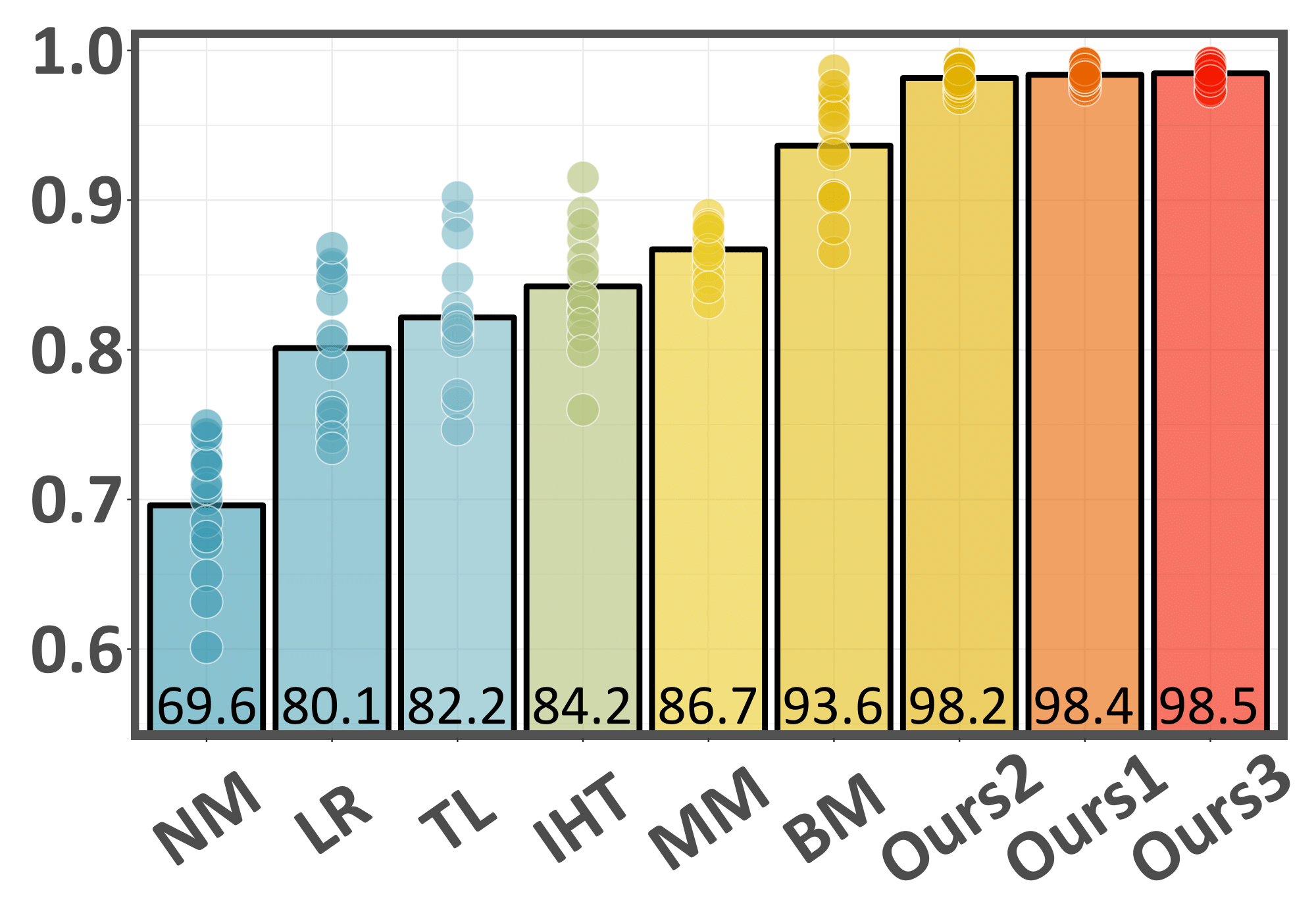} 
  }
  \subfigure[Svmguide2]{
    \includegraphics[width=0.315\textwidth]{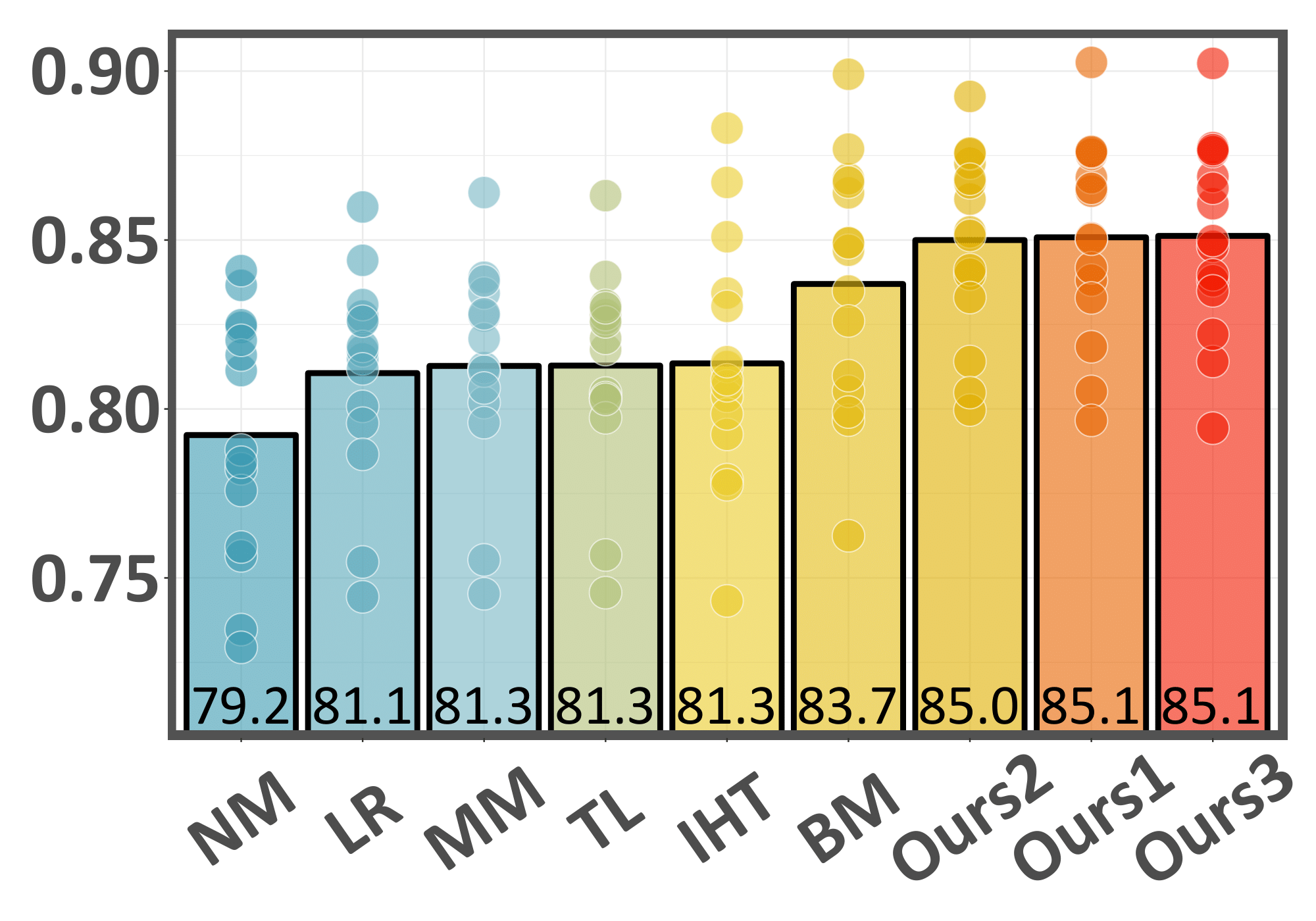} 
  }
  \subfigure[Yeast]{
    \includegraphics[width=0.315\textwidth]{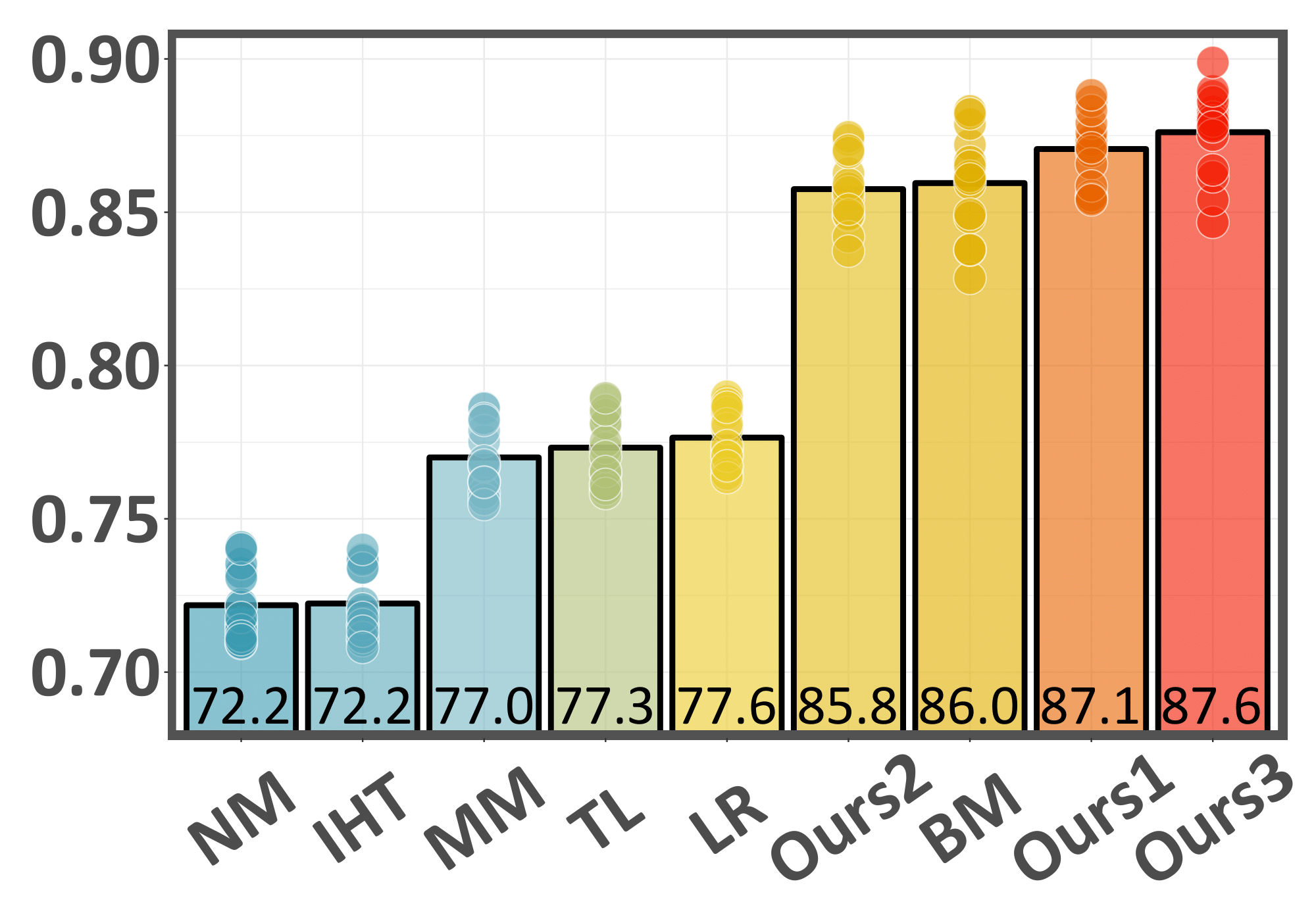} 
  }
  \caption{\label{fig:coarse} \textbf{Coarse-grained Performance Comparison.} We plot the overall $\mathsf{MAUC}^\uparrow$ over the 15 splits against different algorithms. Here each bar presents the performance of a specific algorithm. The bar itself captures the mean performance, and the scatters distributed along a bar are the results over 15 splits.   }
  \end{figure*}

  \begin{figure*}[h]

    \includegraphics[width=0.99\textwidth]{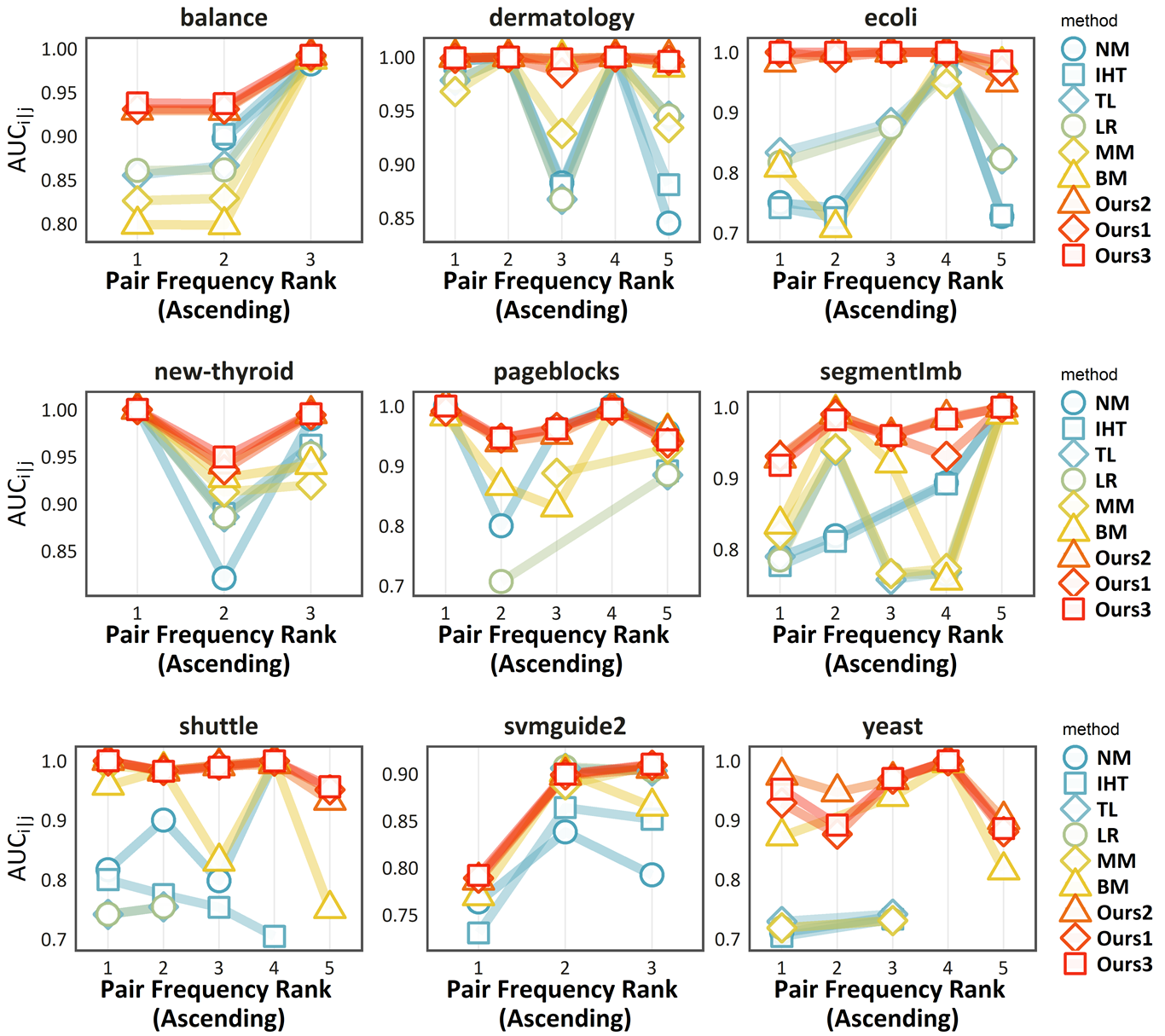} 
   

   \caption{\label{fig:fine}\textbf{Fine-grained Comparison Over the Minority Class Pairs (Traditional Datasets).} 
    The x-axis gives the frequency rank of the class pairs $(i,j)$, \emph{i.e.}, $p_ip_j$, where a larger rank represents a larger frequency.  The y-axis represents the $\aucij$ of the corresponding class pairs. Each line in a plot then captures the minority class pair performance of a given algorithm. To have a clearer look at tendency, we carry out two filtering processes before we visualize the plot: (a) For those datasets which has more than 5 class pairs, \textbf{we only visualize the bottom-5 pairs} in terms of the frequency to turn our focus to the minority pairs in the dataset. (b) To have a clearer look at the difference of the top competitors, we filter out the pairs with a smaller  $\aucij$ than $0.7$.}    
   
   \end{figure*}
\subsection{Datasets}
Now we start the empirical analyses on the real-world datasets. First, we describe all the datasets involved in the forthcoming discussions. Generally speaking, the datasets come from three types of sources: (a) LIBSVM website, (b) KEEL website, and (c) others. Note that for all the datasets prefixed with Imb, we use an imbalanced subset sampled from the original dataset to perform our experiments. Here we only present a brief summary. More detailed contents could be found in Appendix \ref{App:des}.
\begin{enumerate}[itemindent=0pt, leftmargin =14pt]
\item[(a)]  \textbf{LIBSVM Datasets} \footnote{\url{https://www.csie.ntu.edu.tw/~cjlin/libsvmtools/datasets/}}, which includes: \textbf{Shuttle}, \textbf{Svmguide-2}, \textbf{SegmentImb}. \footnote{The goal of the original dataset is to predict image segmentation based on 19 hand-crafted features. To obtain an imbalanced version of the dataset, we sample 17, 33, 26, 13, 264, 231, 165 instances respectively from each of the 7 classes.}

\item[(b)] \textbf{KEEL Datasets}. \footnote{\url{http://www.keel.es/}}, which includes: \textbf{Balance}, \textbf{Dermatology}, \textbf{Ecoli}, \textbf{New Thyroid}, \textbf{Page Blocks}, \textbf{Yeast}.

\item[(c)] \textbf{Other Datasets}:
\begin{enumerate}[ itemindent=0pt, leftmargin =8pt]
\item[(1)] \textbf{CIFAR-100-Imb}. We sampled an imbalanced version of the CIFAR-100 \footnote{\url{https://www.cs.toronto.edu/~kriz/cifar.html}} dataset, which originally contains 100 image classes each with 600 instances.  The 100 classes are encoded as $1,\cdots 100$. 

\item[(2)] \textbf{User-Imb}. The original dataset is collected from TalkingData, a famous third-party mobile data platform from China, which predicts mobile users’ demographic characteristics based on their app usage records. The dataset is collected for the Kaggle Competition named Talking Data Mobile User Demographics. \footnote{\url{https://www.kaggle.com/c/talkingdata-mobile-user-demographics/data}} The raw features include logged events, app attributes, and device information. There are 12 target classes 'F23-', 'F24-26','F27-28','F29-32', 'F33-42', 'F43+', 'M22-', 'M23-26', 'M27-28', 'M29-31', 'M32-38', 'M39+', which describe the demographics (gender and age) of users. In our experiments, we sample an imbalanced subset of the original dataset to leverage a class-skewed dataset. 

\item[(3)]  \textbf{iNaturalist2017.} iNaturalist Challenge 2017 dataset\footnote{https://www.kaggle.com/c/inaturalist-challenge-at-fgvc-2017} is a large-scale image classification benchmark with 675,170 images covering 5,089 different species of plants and animals. We split the dataset into the training set, validation set, and test set at a ratio of 0.7:0.15:0.15. Since directly training deep models in such a large-scale dataset is time-consuming, we instead generate 2048-d features with a ResNet-50 model pre-trained on ImageNet for each image and train models with three fully-connected layers for all methods. We utilize Adam optimizer to train the models, with an initial learning rate of $10^{-5}$. To ensure all categories are covered in a mini-batch, the batch size is set to 8196. Other hyperparameters are the same as those in the CIFAR-100-Imb dataset.


\end{enumerate}
\end{enumerate}
\subsection{Competitors}
\noindent \textbf{Choice of Competitors}. {Specifically, our goal  is to validate that direct MAUC optimization techniques leverage better MAUC performance than other imbalanced learning methods}. Motivated by this, the competitors include: a \textbf{ Baseline} that does not deal with the imbalance distribution; \textbf{Over-sampling and Under-sampling Methods} that tackle imbalance issues by re-sampling; and \textbf{Imbalanced Loss Functions} that tackle imbalance issues by reformulating the loss function. Moreover, we implement our framework with the squared loss, exponential loss, and hinge loss with the acceleration methods proposed in Sec.\ref{sec:d}.    Note that, for the sampling-based methods, we use the code from the python Lib \texttt{Imbalanced-learn} \cite{Imblearn} \footnote{\url{https://imbalanced-learn.readthedocs.io/en/stable/\#}} to implement these competitors. The details are listed as follows:

\begin{enumerate}[ itemindent=0pt, leftmargin =15pt]
  \item \textbf{Standard Baseline: LR}. This is the baseline model where no measures are taken against imbalance issues. For  \textbf{LR}, we adopt a multiclass version of the logistic regression. For traditional datasets, \textbf{LR} is built upon a linear model. For deep learning datasets, \textbf{LR} is built upon a common deep backbone. 

  \item \textbf{Over-sampling Methods}. For these competitors, we first plug in an oversampling method to generate a more balanced dataset, then we use \textbf{LR} to train a model on the new dataset. Here we adopt the following two oversampling methods as our competitors:
  \begin{itemize}
  \item \textbf{BM}. (BorderlineSMOTE) \cite{borderlineSMOTE}: This method is a variant of the SMOTE method (Synthetic Minority Over-sampling Technique), which restricts the oversampling process to the hard minority samples which lie at the borderline of the decision boundary. 
  \item \textbf{MM}. (MWMOTE) \cite{MWMOTE}: MWMOTE is another variant of SMOTE. It first identifies the hard-to-learn informative minority class samples and assigns them weights according to their Euclidean distance from the nearest majority class samples. 
  \end{itemize}
  \item \textbf{Under-sampling Methods}.  For these competitors, we first adopt an undersampling method to generate a more balanced dataset, then we use \textbf{LR} to train a model on the new dataset. Here we adopt the following two methods as our competitors:
  \begin{itemize}
  \item \textbf{IHT}. (InstanceHardnessThreshold) \cite{InstanceHardnessThreshold}. This method works by filtering out the hard and noisy samples from the majority classes based on the instance hardness measure proposed in \cite{InstanceHardnessThreshold}. 
  \item \textbf{NM}. (NearMiss)\cite{NearMiss}: It adopts an under-sampling idea to make majority class samples surround most of the minority class samples.
  \item \textbf{TL}. (TomekLinks) \cite{TomekLinks} It adopts the Tomek Links method to remove redundant samples that fail to contribute to the outline of the decision boundary from the majority class.
  \end{itemize}
  \item \textbf{Imbalanced Loss Functions} (for deep learning): For CIFAR-100-Imb and User-Imb dataset, we also compare our method with some of the recently proposed imbalanced loss functions for deep learning.  
  \begin{itemize}
    \item \textbf{Focal Loss}\cite{FOCAL}: It tackles the imbalance problem by adding a modulating factor to the cross-entropy loss to highlight the hard and minority samples during the training process. 
  
    \item \textbf{CB-CE}: It refers to the loss function that applies the reweighting scheme proposed in \cite{CB-CE-FOCAL} on the cross-entropy loss.

    \item \textbf{CB-Focal}:  It refers to the loss function that applies the reweighting scheme proposed in \cite{CB-CE-FOCAL} on the Focal loss.
    
    \item \textbf{LDAM}\cite{LDAM}: It proposes a Label-distribution-aware margin loss based on the minimum margin per class. Here we adopt the smooth relaxation for cross-entropy loss proposed therein.    
  \end{itemize}
  \item Existing Deep AUC optimization methods:
\begin{itemize}
 \item 
  \textbf{DeepAUC:} \cite{DBLP:conf/iclr/LiuYYY20} is a state-of-the-art stochastic AUC maximization algorithm developed for the deep neural network. It solves the AUC maximization from the saddle point problem. To ensure a fair comparison, we implement the algorithm that extends to the multi-class AUC problem by PyTorch. In addition, there are two optimization configurations in \cite{DBLP:conf/iclr/LiuYYY20}, including Proximal Primal-Dual Stochastic Gradient (PPD-SG) and Proximal Primal-Dual AdaGrad (PPD-AdaGrad). PPD-AdaGrad is employed in our experiment because it usually demonstrates better performance than PPD-SG in most cases.
  
\end{itemize}

  \item Our Methods:
  \begin{itemize}
    \item \textbf{Ours1}: An implementation of our learning framework with the square surrogate loss $\ell_{sq}(\alpha,t) = (\alpha- t)^2 $
  
    \item \textbf{Ours2}: An implementation of our learning framework with the exponential loss $ \ell_{\exp}(\alpha,t) = \exp(-\alpha t)$

    \item \textbf{Ours3}:  An implementation of our learning framework with the hinge loss $ \ell_{hinge}(\alpha,t) = (\alpha- t)_+$
  \end{itemize}
\end{enumerate}

\subsection{Implementation details}
\textbf{Infrastructure}. All the experiments are carried out on a ubuntu 16.04.6 server  equipped with  Intel(R) Xeon(R) CPU E5-2620 v4 cpu and a TITAN RTX GPU. The codes are implemented via \texttt{python 3.6.7}, the basic dependencies are: \texttt{pytorch} (v-1.1.0), \texttt{sklearn} (v-0.21.3), \texttt{numpy} (v-1.16.2). For traditional datasets, we implement our proposed algorithms with the help of the \texttt{sklearn} and \texttt{numpy}. For hinge loss, we use Cython to accelerate the dynamic programming algorithm. For the deep learning datasets, our proposed algorithms are implemented with \texttt{pytorch}.
{\color{white}dsadsa}\\
\noindent\textbf{Evaluation Metric}. Given a trained scoring function $f = (f^{(1)},\cdots, f^{(N_C)})$, all the forthcoming results are evaluated with the MAUC metric (the larger the better). 
\begin{equation*}
\small
\begin{split}
&\mathsf{MAUC}^\uparrow  = \frac{1}{N_C \cdot (N_C - 1)} \cdot \\
&\sumi\suminj \frac{\left|\left\{(\x_1,\x_2): \x_1 \in \mathcal{N}_i, ~\x_2 \in \mathcal{N}_j, f^{(i)}(\x_1) > f^{(i)}(\x_2) \right\}\right|}{\ni\nj}
\end{split}
\end{equation*}

\noindent \textbf{Optimization methods}  As for the optimization methods, we adopt the nestrov-like acceleration method for non-convex optimization algorithm \cite{opt1,opt2,opt3} for training linear model composed with softmax and we adopt adam  to train deep neural networks. We adopt SGD method to train the models for User-Imb dataset, while we use the adam\cite{adam1,adam2} to train the models for the CIFAR-100-Imb dataset.

\textit{The readers are referred to Sec.\ref{app:exp} for more details about how the models are trained in this paper.}

  \begin{table}[htbp]
    \centering
  \caption{\label{tab:perffin}Performance comparison based on $\mathsf{MAUC}^\uparrow$ with Deep Learning}
  \begin{tabular}{l|l|c|c|c}
    \toprule
    \textbf{type} & \textbf{method} & \textbf{CIFAR-100} & \textbf{User} & \textbf{iNaturalist} \\
    \midrule
    Baseline & CE    & \cellcolor[rgb]{ .996,  .91,  .843}$59.80$ & \cellcolor[rgb]{ .937,  .957,  .89}$55.26$ & \cellcolor[rgb]{ .918,  .965,  .976}$66.84$ \\
    \midrule
    \multicolumn{1}{c|}{\multirow{5}[2]{*}{Sampling}} & BM    & \cellcolor[rgb]{ 1,  .965,  .933}$59.07$ & \cellcolor[rgb]{ .827,  .886,  .71}$58.95$ & \cellcolor[rgb]{ .702,  .863,  .906}$78.18$ \\
          & MM    & $58.52$ & \cellcolor[rgb]{ .933,  .957,  .886}$55.35$ & \cellcolor[rgb]{ .914,  .961,  .973}$67.07$ \\
          & IHT   & \cellcolor[rgb]{ .992,  .859,  .749}$60.56 $& \cellcolor[rgb]{ .918,  .945,  .859}$55.87$ & \cellcolor[rgb]{ .914,  .961,  .973}$67.08$ \\
          & NM    & \cellcolor[rgb]{ .992,  .882,  .788}$60.24$ & \cellcolor[rgb]{ .957,  .973,  .925}$54.52 $& \cellcolor[rgb]{ .922,  .965,  .976}$66.74$ \\
          & TL    & \cellcolor[rgb]{ .992,  .878,  .784}$60.28$ & $52.99$ & \cellcolor[rgb]{ .922,  .965,  .976}$66.77$\\
    \midrule
    \multirow{4}[2]{*}{Loss} & FOCAL & \cellcolor[rgb]{ .992,  .894,  .816}$60.03$ & \cellcolor[rgb]{ .933,  .957,  .886}$55.34$ & \cellcolor[rgb]{ .918,  .961,  .973}$66.97$ \\
          & CBCE  & \cellcolor[rgb]{ .992,  .894,  .816}$60.04$ & \cellcolor[rgb]{ .831,  .886,  .71}$58.91$ & \cellcolor[rgb]{ .71,  .867,  .91}$77.71$ \\
          & CBFOCAL & \cellcolor[rgb]{ .992,  .867,  .769}$60.42$ & \cellcolor[rgb]{ .835,  .89,  .718}$58.75 $& \cellcolor[rgb]{ .835,  .925,  .949}$71.12$ \\
          & LDAM  & \cellcolor[rgb]{ .992,  .878,  .788}$60.26$ & \cellcolor[rgb]{ .902,  .933,  .831}$56.47$ & \cellcolor[rgb]{ .796,  .91,  .937}73.18 \\
    \midrule
    \multirow{4}[2]{*}{AUC} & DeepAUC & \cellcolor[rgb]{ .992,  .878,  .784}$60.27$ & \cellcolor[rgb]{ .8,  .867,  .659}$59.93$ & $62.48$ \\
    \cmidrule{2-5} 
   & Ours1 & \cellcolor[rgb]{ .984,  .765,  .584}$\underline{61.90}$ & \cellcolor[rgb]{ .78,  .851,  .624}$\underline{60.64}$ & \cellcolor[rgb]{ .675,  .851,  .898}$\underline{79.67}$ \\
          & Ours2 &\cellcolor[rgb]{ .98,  .749,  .561}$\bm{62.08}$& \cellcolor[rgb]{ .769,  .843,  .608}$\bm{60.94}$& \cellcolor[rgb]{ .573,  .804,  .863}$\bm{84.87}$ \\
          & Ours3 & \cellcolor[rgb]{ .996,  .922,  .863}$59.64$ & \cellcolor[rgb]{ .812,  .875,  .678}$59.52$ & \cellcolor[rgb]{ .682,  .855,  .898}$79.16$\\
    \bottomrule
  \end{tabular}%

 \end{table}%

\subsection{Results and Analysis}

\begin{table*}[htbp]
  \begin{center}
  \normalsize
  \caption{\textbf{Fine-grained Comparison Over the Minority Class Pairs (Deep Learning Datasets).} 
 Again, we provide a finer-grained comparison result on the minority class pairs of the two deep learning datasets.
Here 1st, 2nd, 3rd, 4th, 5th are the \textbf{bottom 5 class pairs $(i,j)$} with smallest $p_ip_j$. \label{tab:finedeep}}
   
\scalebox{0.9}{

  \begin{tabular}{c|l|ccccc|ccccc|}
    \toprule
    \multirow{2}[4]{*}{\textbf{type}} & \textbf{method} & \multicolumn{5}{c|}{\textbf{CIFAR-100-Imb}} & \multicolumn{5}{c|}{\textbf{User-Imb}} \\
\cmidrule{2-12}          & \textbf{bottom pairs} & \textbf{1st} & \textbf{2nd} & \textbf{3rd} & \textbf{4th} & \textbf{5th} & \textbf{1st} & \textbf{2nd} & \textbf{3rd} & \textbf{4th} & \textbf{5th} \\
    \midrule
    Baseline & CE    & \cellcolor[rgb]{ .996,  .922,  .859}$48.82$ & \cellcolor[rgb]{ .988,  .835,  .71}$53.28$ & \cellcolor[rgb]{ .992,  .871,  .769}$50.74$ & \cellcolor[rgb]{ .988,  .843,  .722}$58.44$ & \cellcolor[rgb]{ .992,  .863,  .757}$56.17$ & \cellcolor[rgb]{ .973,  .98,  .949}$49.43$ & \cellcolor[rgb]{ .969,  .98,  .949}$49.51$ & \cellcolor[rgb]{ .937,  .957,  .89}$53.72$ & \cellcolor[rgb]{ .898,  .933,  .827}$58.06$ & \cellcolor[rgb]{ .973,  .98,  .953}$49.73 $\\
    \midrule
    \multicolumn{1}{c|}{\multirow{5}[2]{*}{Sampling}} & BM    & \cellcolor[rgb]{ .992,  .871,  .773}$50.46$ & \cellcolor[rgb]{ .992,  .867,  .765}$50.65$ & \cellcolor[rgb]{ .992,  .855,  .745}$51.97$ & \cellcolor[rgb]{ .996,  .945,  .898}$54.44$ & \cellcolor[rgb]{ .992,  .855,  .745}$56.79 $& \cellcolor[rgb]{ .882,  .922,  .804}$54.64$ & \cellcolor[rgb]{ .894,  .929,  .824}$53.04$ & \cellcolor[rgb]{ .89,  .925,  .812}$56.36$ & \cellcolor[rgb]{ .894,  .929,  .816}$58.44$ & \cellcolor[rgb]{ .776,  .851,  .624}$58.50$ \\
          & MM    & \cellcolor[rgb]{ 1,  .992,  .988}$46.45$ & $38.70$ & $39.07$ & \cellcolor[rgb]{ 1,  .988,  .98}$52.61$ & $44.71$ & \cellcolor[rgb]{ .914,  .945,  .855}$52.77$ & \cellcolor[rgb]{ .965,  .976,  .941}$49.68$ & \cellcolor[rgb]{ .918,  .945,  .859}$54.82$ & \cellcolor[rgb]{ .859,  .902,  .757}$60.42$ & $48.39$ \\
          & IHT   & $46.18$ & \cellcolor[rgb]{ .992,  .855,  .745}$51.60$ & \cellcolor[rgb]{ .988,  .839,  .714}$53.53$ & \cellcolor[rgb]{ 1,  .961,  .933}$53.67$ & \cellcolor[rgb]{ .988,  .831,  .706}$58.63$ & \cellcolor[rgb]{ .855,  .902,  .749}$56.49$ & \cellcolor[rgb]{ .859,  .906,  .761}$54.77$ & \cellcolor[rgb]{ .859,  .906,  .761}$58.02$ & \cellcolor[rgb]{ .835,  .89,  .722}$61.60$ & \cellcolor[rgb]{ .922,  .949,  .867}$52.00$ \\
          & NM    & \cellcolor[rgb]{ .996,  .914,  .851}$49.01$ & \cellcolor[rgb]{ .988,  .812,  .667}$55.40$ & \cellcolor[rgb]{ .988,  .827,  .698}$54.35$ & \cellcolor[rgb]{ .984,  .784,  .62}$\underline{60.78}$ & \cellcolor[rgb]{ .984,  .796,  .639}$61.75$ & \cellcolor[rgb]{ .973,  .98,  .949}$49.38$ & \cellcolor[rgb]{ .933,  .957,  .886}$51.21$ & \cellcolor[rgb]{ .933,  .953,  .882}$53.99$ & \cellcolor[rgb]{ .937,  .957,  .89}$56.05$ & \cellcolor[rgb]{ .914,  .941,  .855}$52.31$ \\
          & TL    & \cellcolor[rgb]{ .996,  .922,  .863}$48.75$ & \cellcolor[rgb]{ .988,  .839,  .718}$52.88$ & \cellcolor[rgb]{ .992,  .875,  .776}$50.43$ & $52.11$ & \cellcolor[rgb]{ .992,  .867,  .765}$55.83$ & $47.55$ & $47.96$ & $50.11$ & $52.35$ & \cellcolor[rgb]{ .945,  .965,  .906}$50.92$ \\
    \midrule
    \multirow{4}[2]{*}{Imbalanced} & FOCAL & \cellcolor[rgb]{ .988,  .847,  .729}$51.25$ & \cellcolor[rgb]{ .988,  .843,  .725}$52.58$ & \cellcolor[rgb]{ .992,  .89,  .808}$48.89$ & \cellcolor[rgb]{ .988,  .851,  .733}$58.17$ & \cellcolor[rgb]{ .988,  .812,  .671}$60.38$ & \cellcolor[rgb]{ .965,  .976,  .937}$49.83$& \cellcolor[rgb]{ .961,  .976,  .933}$49.88$ & \cellcolor[rgb]{ .929,  .953,  .878}$54.16$ & \cellcolor[rgb]{ .886,  .922,  .804}$58.86$ & \cellcolor[rgb]{ .973,  .98,  .949}$49.77$ \\
          & CBCE  & \cellcolor[rgb]{ .996,  .933,  .882}$48.42$ & \cellcolor[rgb]{ .988,  .835,  .71}$53.33$ & \cellcolor[rgb]{ .988,  .851,  .733}$52.53$ & \cellcolor[rgb]{ .996,  .929,  .871}$55.06$ & \cellcolor[rgb]{ .988,  .831,  .706}$58.67$ & \cellcolor[rgb]{ .769,  .843,  .608}$\bm{61.48}$ & \cellcolor[rgb]{ .769,  .843,  .608}$\bm{59.05}$ & \cellcolor[rgb]{ .792,  .863,  .651}$61.60 $& \cellcolor[rgb]{ .784,  .855,  .635}$64.42$ & \cellcolor[rgb]{ .769,  .843,  .608}$\bm{58.82}$ \\
          & CBFOCAL & \cellcolor[rgb]{ .992,  .894,  .812}$49.74$ & \cellcolor[rgb]{ .992,  .859,  .753}$51.18$ & \cellcolor[rgb]{ .992,  .882,  .796}$49.45$ & \cellcolor[rgb]{ .988,  .847,  .733}$58.22 $& \cellcolor[rgb]{ .988,  .835,  .714}$58.29$ & \cellcolor[rgb]{ .773,  .847,  .612}$\underline{61.45}$ & \cellcolor[rgb]{ .773,  .847,  .612}$\underline{58.98}$ & \cellcolor[rgb]{ .796,  .863,  .655}$61.39$ & \cellcolor[rgb]{ .792,  .859,  .647}$64.09$ & \cellcolor[rgb]{ .773,  .847,  .612}$\underline{58.75}$ \\
          & LDAM  & \cellcolor[rgb]{ .996,  .925,  .867}$48.68$ & \cellcolor[rgb]{ .992,  .867,  .769}$50.45$ & \cellcolor[rgb]{ .988,  .847,  .733}${52.57}$ & \cellcolor[rgb]{ .992,  .875,  .78}$57.17$ & \cellcolor[rgb]{ .988,  .816,  .678}$59.96$ & \cellcolor[rgb]{ .953,  .969,  .922}$50.46$ & \cellcolor[rgb]{ .914,  .941,  .851}$52.18$ & \cellcolor[rgb]{ .91,  .941,  .847}$55.20$ & \cellcolor[rgb]{ .855,  .902,  .749}$60.64$ & \cellcolor[rgb]{ .98,  .988,  .965}$49.33$ \\
    \midrule
    \multirow{4}[2]{*}{AUC} & DeepAUC & \cellcolor[rgb]{ .996,  .929,  .875}$48.55$ & \cellcolor[rgb]{ .988,  .839,  .718}$53.03$ & \cellcolor[rgb]{ .988,  .824,  .69}$54.68$ & \cellcolor[rgb]{ .98,  .749,  .561}$\bm{62.06}$ & \cellcolor[rgb]{ .984,  .792,  .635}$61.88$ & \cellcolor[rgb]{ .937,  .957,  .89}$51.56$ & \cellcolor[rgb]{ .847,  .898,  .741}$55.30$ & \cellcolor[rgb]{ .91,  .937,  .843}$55.27$ & \cellcolor[rgb]{ .945,  .965,  .906}$55.45$ & \cellcolor[rgb]{ .804,  .867,  .667}$57.34$ \\
    \cmidrule{2-12} 
    & Ours1 & \cellcolor[rgb]{ .988,  .847,  .733}$51.18$ & \cellcolor[rgb]{ .984,  .788,  .627}$\underline{57.48}$ & \cellcolor[rgb]{ .984,  .773,  .6}$\underline{59.39}$ & \cellcolor[rgb]{ .984,  .796,  .639}$60.33$ & \cellcolor[rgb]{ .984,  .792,  .635}${62.04}$& \cellcolor[rgb]{ .816,  .875,  .686}$58.74 $& \cellcolor[rgb]{ .788,  .859,  .639}$58.20$ & \cellcolor[rgb]{ .78,  .851,  .624}$\underline{62.41}$& \cellcolor[rgb]{ .784,  .855,  .631}$\underline{64.49}$ & \cellcolor[rgb]{ .8,  .863,  .655}$57.58$ \\
          & Ours2 & \cellcolor[rgb]{ .988,  .804,  .659}$\underline{52.57}$ & \cellcolor[rgb]{ .988,  .82,  .682}$54.68$ & \cellcolor[rgb]{ .984,  .796,  .639}$57.31$ & \cellcolor[rgb]{ .988,  .82,  .682}$59.33$ & \cellcolor[rgb]{ .98,  .749,  .561}$\bm{65.38}$ & \cellcolor[rgb]{ .812,  .875,  .678}$59.02$ & \cellcolor[rgb]{ .78,  .851,  .627}$58.54$ & \cellcolor[rgb]{ .769,  .843,  .608}$\bm{62.90}$ & \cellcolor[rgb]{ .769,  .843,  .608}$\bm{65.26}$ & \cellcolor[rgb]{ .8,  .863,  .659}$57.55$ \\
          & Ours3 & \cellcolor[rgb]{ .98,  .749,  .561}$\bm{54.34}$ & \cellcolor[rgb]{ .98,  .749,  .561}$\bm{60.70}$ & \cellcolor[rgb]{ .98,  .749,  .561}$\bm{61.18}$ & \cellcolor[rgb]{ .988,  .851,  .733}$58.17$ & \cellcolor[rgb]{ .984,  .753,  .569}$\underline{65.13}$ & \cellcolor[rgb]{ .804,  .867,  .667}$59.52$ & \cellcolor[rgb]{ .824,  .878,  .698}$56.59$ & \cellcolor[rgb]{ .82,  .878,  .694}$60.09$ & \cellcolor[rgb]{ .8,  .867,  .663}$63.56$& \cellcolor[rgb]{ .824,  .882,  .702}$56.37$ \\
    \bottomrule
    \end{tabular}%

	}

      \end{center}
\end{table*}%

\begin{table}[htbp]  \label{tab:inat}%
  \centering
  \caption{Peformance Comparison on iNaturalist 2017 Dataset.}
  \scalebox{0.8}{
    \begin{tabular}{c|l|ccccc}
    \toprule
    \multirow{2}[4]{*}{\textbf{type}} & \textbf{method} & \multicolumn{5}{c}{iNaturalist} \\
\cmidrule{2-7}          & \textbf{bottom pairs} & \textbf{0-5\%} & \textbf{5-10\%} & \textbf{10-15\%} & \textbf{15-20\%} & \textbf{20-25\%} \\
    \midrule
    Baseline & CE    & \cellcolor[rgb]{ .859,  .937,  .957}$62.62$ & \cellcolor[rgb]{ .859,  .937,  .957}$62.45$ & \cellcolor[rgb]{ .859,  .937,  .957}$62.33$ & \cellcolor[rgb]{ .855,  .933,  .953}$62.55$ & \cellcolor[rgb]{ .847,  .933,  .953}$62.78$ \\
    \midrule
    \multicolumn{1}{c|}{\multirow{5}[2]{*}{Sampling }} & BM    & \cellcolor[rgb]{ .655,  .843,  .89}$78.68$ & \cellcolor[rgb]{ .655,  .843,  .89}$78.20$ & \cellcolor[rgb]{ .651,  .843,  .89}$77.96$ & \cellcolor[rgb]{ .651,  .839,  .89}$78.06$ & \cellcolor[rgb]{ .651,  .839,  .89}$78.06$ \\
          & MM    & \cellcolor[rgb]{ .855,  .933,  .957}$62.83$ & \cellcolor[rgb]{ .855,  .933,  .957}$62.65$ & \cellcolor[rgb]{ .855,  .933,  .953}$62.57$ & \cellcolor[rgb]{ .851,  .933,  .953}$62.78$ & \cellcolor[rgb]{ .843,  .929,  .953}$63.04$ \\
          & IHT   & \cellcolor[rgb]{ .855,  .933,  .953}$63.08$ & \cellcolor[rgb]{ .855,  .933,  .953}$62.82$ & \cellcolor[rgb]{ .855,  .933,  .953}$62.66$ & \cellcolor[rgb]{ .851,  .933,  .953}$62.80$ & \cellcolor[rgb]{ .843,  .929,  .953}$63.07$ \\
          & NM    & \cellcolor[rgb]{ .859,  .937,  .957}$62.66$ & \cellcolor[rgb]{ .859,  .937,  .957}$62.44$ & \cellcolor[rgb]{ .859,  .937,  .957}$62.30$ & \cellcolor[rgb]{ .855,  .933,  .953}$62.45$ & \cellcolor[rgb]{ .851,  .933,  .953}$62.71$ \\
          &   TL    & \cellcolor[rgb]{ .859,  .937,  .957}$62.50$ & \cellcolor[rgb]{ .859,  .937,  .957}$62.32$ & \cellcolor[rgb]{ .859,  .937,  .957}$62.23$ & \cellcolor[rgb]{ .855,  .933,  .953}$62.46$ & \cellcolor[rgb]{ .847,  .933,  .953}$62.75$ \\
    \midrule
    \multirow{4}[2]{*}{Imbalanced} & FOCAL & \cellcolor[rgb]{ .855,  .933,  .953}$62.97$ & \cellcolor[rgb]{ .855,  .933,  .953}$62.69$ & \cellcolor[rgb]{ .855,  .933,  .953}$62.53$ & \cellcolor[rgb]{ .851,  .933,  .953}$62.73$ & \cellcolor[rgb]{ .847,  .929,  .953}$63.01$ \\
          & CBCE  & \cellcolor[rgb]{ .6,  .82,  .875}$\underline{83.09}$ & \cellcolor[rgb]{ .616,  .824,  .878}$\underline{81.22}$ & \cellcolor[rgb]{ .624,  .827,  .882}$\underline{80.18}$ & \cellcolor[rgb]{ .631,  .831,  .882}$\underline{79.50}$ & \cellcolor[rgb]{ .639,  .835,  .886}$\underline{79.02}$ \\
          & CBFOCAL & \cellcolor[rgb]{ .655,  .843,  .89}$78.65$ & \cellcolor[rgb]{ .682,  .855,  .898}$75.90 $& \cellcolor[rgb]{ .698,  .863,  .906}$74.49$ & \cellcolor[rgb]{ .71,  .867,  .91}$73.55$ & \cellcolor[rgb]{ .718,  .871,  .91}$72.91$ \\
          & LDAM  & \cellcolor[rgb]{ .753,  .886,  .922}$70.94$ & \cellcolor[rgb]{ .753,  .886,  .922}$70.48$ & \cellcolor[rgb]{ .753,  .886,  .922}$70.34$ & \cellcolor[rgb]{ .749,  .886,  .922}$70.49$ & \cellcolor[rgb]{ .745,  .882,  .918}$70.84 $\\
    \midrule
    \multirow{4}[4]{*}{AUC} & DeepAUC & $51.31$ & $51.34$ & $51.31$ &$ 51.21$ & $50.92$ \\
\cmidrule{2-7}          & Ours1 & \cellcolor[rgb]{ .635,  .835,  .886}$80.29$ & \cellcolor[rgb]{ .639,  .835,  .886}$79.37$ & \cellcolor[rgb]{ .639,  .835,  .886}$78.98$ & \cellcolor[rgb]{ .639,  .835,  .886}$79.02$ & \cellcolor[rgb]{ .639,  .835,  .886}$79.00$ \\
          & Ours2 & \cellcolor[rgb]{ .573,  .804,  .863}$\bm{85.14}$ & \cellcolor[rgb]{ .573,  .804,  .863}$\bm{84.33}$ & \cellcolor[rgb]{ .573,  .804,  .863}$\bm{83.93}$ & \cellcolor[rgb]{ .573,  .804,  .863}$\bm{83.94}$ & \cellcolor[rgb]{ .573,  .804,  .863}$\bm{83.92}$ \\
          & Ours3 & \cellcolor[rgb]{ .655,  .843,  .89}$78.65$ & \cellcolor[rgb]{ .682,  .855,  .898}$75.90$ & \cellcolor[rgb]{ .698,  .863,  .906}$74.49$ & \cellcolor[rgb]{ .71,  .867,  .91}$73.55$ & \cellcolor[rgb]{ .718,  .871,  .91}$72.91$ \\
    \bottomrule
    \end{tabular}%
  }

\end{table}%

\textbf{Traditional Datasets}. The average performances of 15 repetitions for the nine traditional datasets are shown in Fig.\ref{fig:coarse}, where the scatters show the 15 observations over different dataset splits, and the bar plots show the average performance over 15 repetitions. Consequently, we have the following observations: 1) The best performance of our proposed algorithm consistently surpasses all the competitors significantly on all the datasets. More specifically,  the best algorithm among Ours1, Our2 and Our3 outperforms the  best competitors by a margin of $5.3$, $1.5$, $5.6$, $1.3$, $4.4$, $2.9$, $4.9$, $1.4$, $1.8$ for \emph{Balance}, \emph{Dermatology}, \emph{Ecoli}, \emph{New Thyroid}, \emph{Page Blocks}, \emph{SegmentImb}, \emph{Shuttle}, \emph{Svmguide2}, and \emph{Yeast}, respectively. It turns out that our improvements are significant in most cases. 2) It could always be observed that some of the sampling-based competitors fail to outperform the original LR algorithm. One reason for this phenomenon might be that the sampling-based methods are not directly designed for optimizing  AUC.   Another possible reason is that most of the sampling methods adopt an ova strategy to perform multiclass sampling. This has a similar negative effect, as shown in Thm.\ref{prop:mo} for $\aucova$, where the imbalance issue across class pairs is not taken into consideration. 3) Based on the fact that an imbalanced dataset's performance bottleneck comes from its minority classes, we investigate a closer look at the performance comparison on minority class pairs. Specifically, we visualize the finer-grained comparison for 5 class pairs having the smallest frequency for each data, which is shown in Fig.\ref{fig:fine}. Note that we only present the results that are greater than 0.7 to have a clearer view of the differences across top algorithms. The results show that our improvement over the minority class pairs is even more significant, especially for the Balance, Ecoli, Page Blocks, SegmentImb, Shuttle, and Yeast.\\ 
\noindent  \textbf{Deep Learning Datasets}. The performance comparisons are shown in the Tab.\ref{tab:perffin}. Consequently, when using either the squared loss (Ours1) or the exponential loss (Ours2), our proposed algorithm consistently outperforms all the competitors for all the datasets. Note that DeepAUC has a reasonable performance on the first two datasets. However, it fails in the last dataset. The possible reason here is that DeepAUC requires an extra set of parameters with the size of which depending on $N_C$. Hence, for a large dataset like iNaturalist, training DeepAUC becomes significantly more difficult than other methods. Next, we show the finer-grained comparisons of the minority class pairs in Tab.\ref{tab:finedeep} and Tab.\ref{tab:inat}. For CIFAR-100-Imb and User-Imb, we adopt a similar to the traditional datasets where the performance comparisons in terms of $\aucij$ on the class pairs $(i,j)$ with bottom-5 pairwise frequency $p_ip_j$. For the CIFAR-100-Imb dataset,  the best algorithm among Ours1, Our2, and Our3 significantly outperform their best competitor for all the class pairs except the 4-th one. Moreover, even though the hinge loss (Ours3) does not provide an improvement of $\mathsf{MAUC}^\uparrow$, it at least mitigates the imbalanced issue by providing good performances on the first four minority pairs.  For User-Imb, the corresponding improvements produced by our proposed methods are not that sharp as the previous results. The possible reasons are that: (1) the imbalance degree of User-Imb is relatively moderate; (2) even the minority classes in this dataset have a sufficient amount of samples as high as 400. These two traits automatically alleviate the imbalance issue in this dataset, making the difference between different imbalance-aware methods seem less significant. For the iNaturalist 2017 Dataset, since the tail classes are too rate to differentiate the performance from the different algorithms, we turn to report the result for bottom 0-5\%, 6-10\%, 11-15\%, 16-20\%, 21-25\% class pairs. Again, we can see that our proposed algorithm outperforms the competitors consistently.

\section{Conclusion}
This paper provides an early study of AUC-guided machine learning for multiclass problems. Specifically, we propose a novel framework to learn scoring functions by optimizing the popular M metric where employs surrogate losses for the 0-1 loss to leverage a differentiable objective function. Utilizing consistency analysis, we show that a series of surrogate loss fucntions are fisher consistent with the 0-1 loss based M metric. Moreover, we provide an empirical surrogate risk minimization framework to minimize the $\mauc$ with guaranteed generalization upper bounds. Practically, we propose acceleration methods for three implementations of our proposed framework. Finally, empirical studies on 11 real-world datasets show the efficacy of our proposed algorithms.

\section{Ackownledgment}
This work was supported in part by the National Key R\&D Program of China under Grant 2018AAA0102003, in part by National Natural Science Foundation of China: 61620106009, 62025604, 61861166002, 61931008, 61836002 and 61976202, in part by the Fundamental Research Funds for the Central Universities, in part by the National Postdoctoral Program for Innovative Talents under Grant BX2021298, in part by Youth Innovation Promotion Association CAS, and in part by the Strategic Priority Research Program of Chinese Academy of Sciences, Grant No. XDB28000000.

\bibliographystyle{abbrv}
\bibliography{mauc-fin}
\ifCLASSOPTIONcaptionsoff
\newpage
\fi

\vspace{-0.2cm}\begin{IEEEbiography}[{\includegraphics[width=1in,height=1.25in,clip,keepaspectratio]{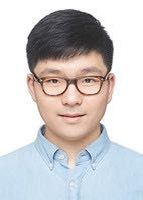}}]{\textbf{Zhiyong Yang} received the M.Sc. degree in computer
science and technology from University of Science
and Technology Beijing (USTB) in 2017, and the Ph.D. degree from University of Chinese Academy of Sciences (UCAS) in 2021. He is currently a postdoctoral research fellow with the University of
Chinese Academy of Sciences. His research interests
lie in machine learning and learning theory, with special focus on AUC optimization,
meta-learning/multi-task learning, and learning theory for recommender systems. He has authored or coauthored several academic papers in top-tier international conferences and journals
including T-PAMI/ICML/NeurIPS/CVPR. He served as a reviewer for several top-tier journals and conferences such as T-PAMI, ICML, NeurIPS and ICLR.
}
\end{IEEEbiography}
\vspace{-0.2cm}\begin{IEEEbiography}[{\includegraphics[width=1in,height=1.25in,clip,keepaspectratio]{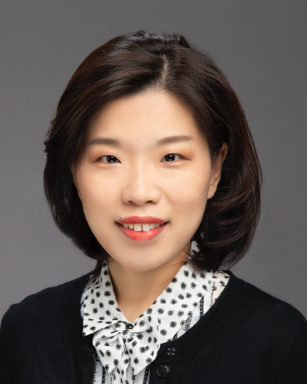}}]{\textbf{Qianqian Xu}  received the B.S. degree in computer science from China University of Mining and Technology in 2007 and the Ph.D. degree in computer science from University of Chinese Academy of Sciences in 2013. She is currently an Associate Professor with the Institute of Computing Technology, Chinese Academy of Sciences, Beijing, China. Her research interests include statistical machine learning, with applications in multimedia and computer vision. She has authored or coauthored 40+ academic papers in prestigious international journals and conferences (including T-PAMI, IJCV, T-IP, NeurIPS, ICML, CVPR, AAAI, etc), among which she has published 6 full papers with the first author's identity in ACM Multimedia. She served as member of professional committee of CAAI, and member of online program committee of VALSE, etc.}
\end{IEEEbiography}

\vspace{-0.2cm}\begin{IEEEbiography}[{\includegraphics[width=1in,height=1.25in,clip,keepaspectratio]{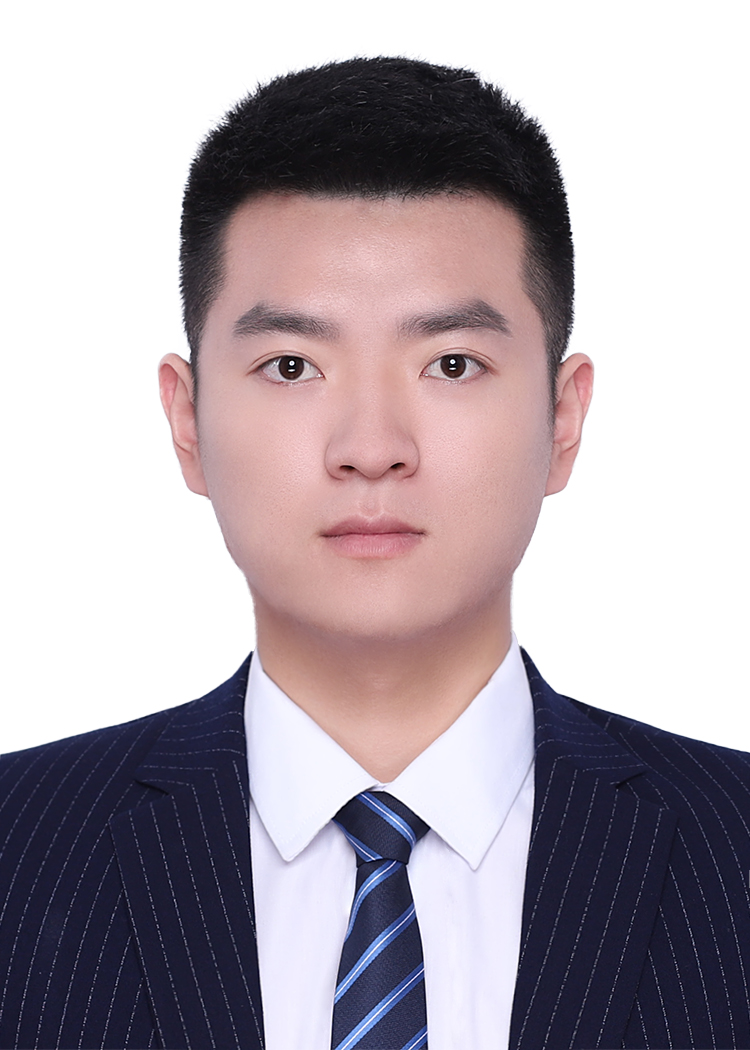}}]{Shilong Bao received the B.S. degree in College of Computer Science and Technology from Qingdao University in 2019. He is currently pursuing the M.S. degree with University of Chinese Academy of Sciences. His research interest is machine learning and data mining.}
\end{IEEEbiography}

\vspace{-0.5cm}\begin{IEEEbiography}[{\includegraphics[width=1in,height=1.25in,clip,keepaspectratio]{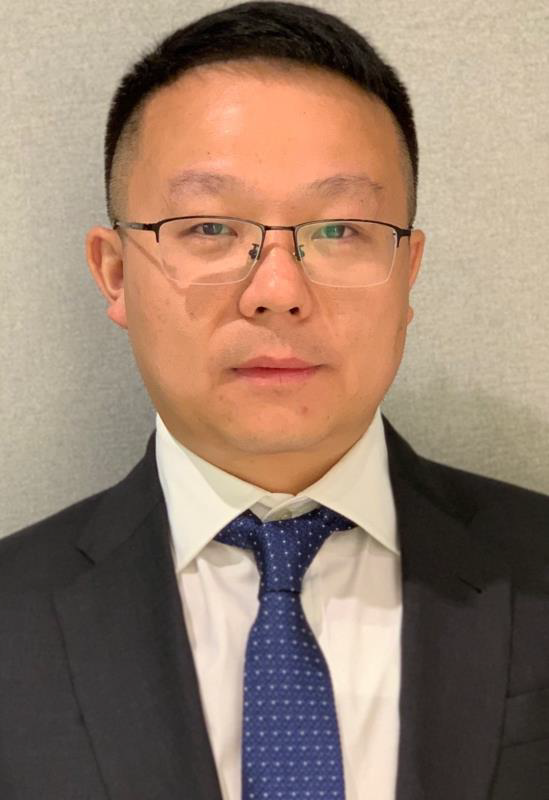}}]{\textbf{Xiaochun Cao}, Professor of the Institute of Information Engineering, Chinese Academy of Sciences. He received the B.E. and M.E. degrees both in computer science from Beihang University (BUAA), China, and the Ph.D. degree in computer science from the University of Central Florida, USA, with his dissertation nominated for the university level Outstanding Dissertation Award. After graduation, he spent about three years at ObjectVideo Inc. as a Research Scientist. From 2008 to 2012, he was a professor at Tianjin University. He has authored and coauthored over 100 journal and conference papers. In 2004 and 2010, he was the recipients of the Piero Zamperoni best student paper award at the International Conference on Pattern Recognition. He is a fellow of IET and a Senior Member of IEEE. He is an associate editor of IEEE Transactions on Image Processing, IEEE Transactions on Circuits and Systems for Video Technology and IEEE Transactions on Multimedia.}
\end{IEEEbiography}

\vspace{-0.2cm}\begin{IEEEbiography}[{\includegraphics[width=1in,height=1.25in,clip,keepaspectratio]{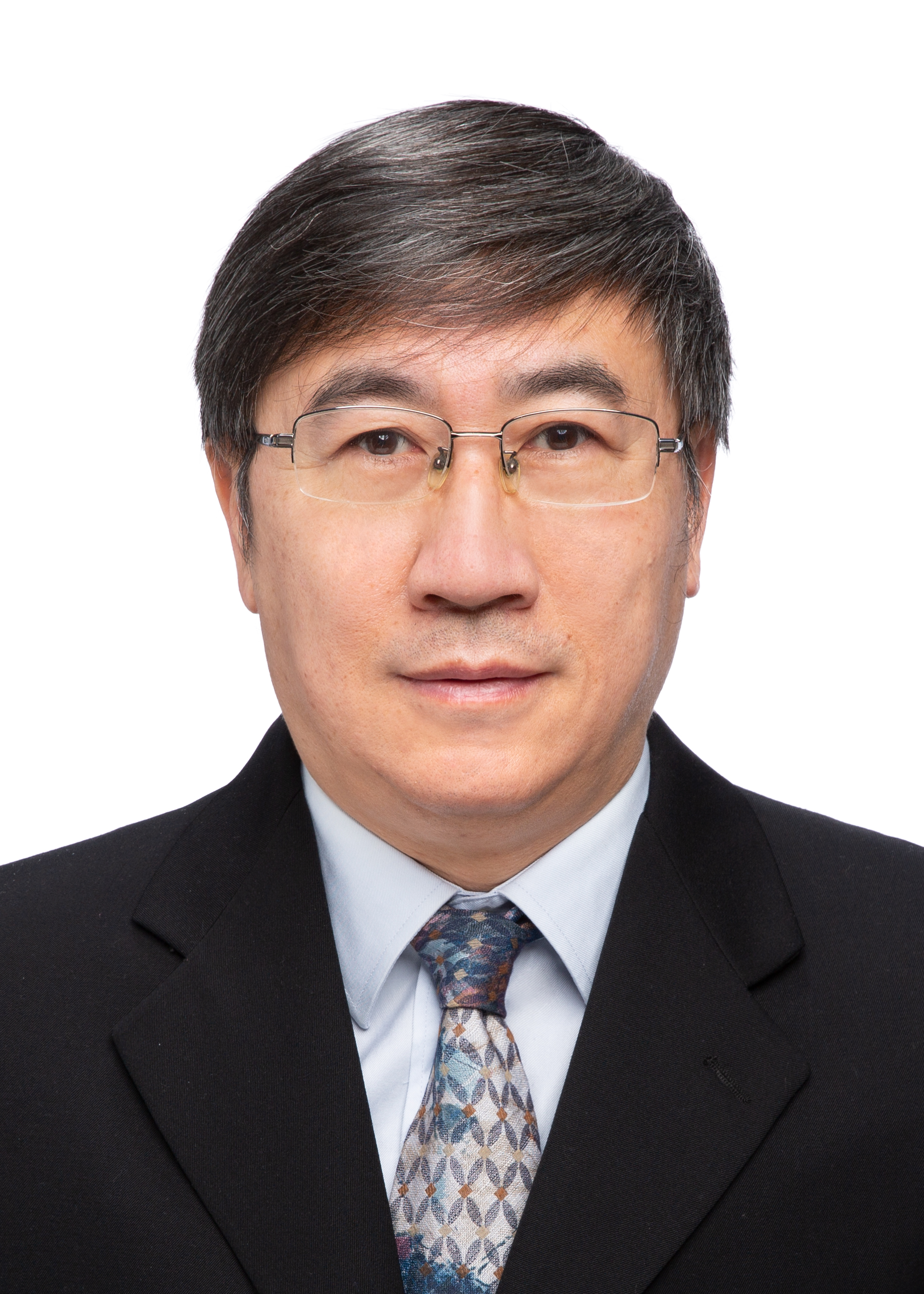}}]{\textbf{Qingming Huang} is a chair professor in the University of Chinese Academy of Sciences and an adjunct research professor in the Institute of Computing Technology, Chinese Academy of Sciences. He graduated with a Bachelor degree in Computer Science in 1988 and Ph.D. degree in Computer Engineering in 1994, both from Harbin Institute of Technology, China. His research areas include multimedia computing, image processing, computer vision and pattern recognition. He has authored or coauthored more than 400 academic papers in prestigious international journals and top-level international conferences. He is the associate editor of IEEE Trans. on CSVT and Acta Automatica Sinica, and the reviewer of various international journals including IEEE Trans. on PAMI, IEEE Trans. on Image Processing, IEEE Trans. on Multimedia, etc. He is a Fellow of IEEE and has served as general chair, program chair, track chair and TPC member for various conferences, including ACM Multimedia, CVPR, ICCV, ICME, ICMR, PCM, BigMM, PSIVT, etc.}
\end{IEEEbiography}
\newpage
\onecolumn
\appendices

\textcolor{white}{dasdsa}\\

\section*{\textcolor{blue}{\Large{Contents}}}

\startcontents[sections]
\printcontents[sections]{l}{1}{\setcounter{tocdepth}{3}}
\newpage

\section{Comparison Properties Between $AUC^{ova}$ and $AUC^{ovo}$ }\label{sec:appa}
In this section, we will provide a comparison result for $AUC^{ova}$ and $AUC^{ovo}$, which suggests that  employing  $AUC^{ova}$ does not consider the  imbalance issue across class pairs. Then we will provide a practical   example to compare these two metrics.

\begin{rthm1}
Given the label distribution as $\mathbb{P}[y=i] = p_i>0$ and any multiclass scoring function $f$. The following properties hold:
\begin{enumerate}
\item[(a)] We have that: 
\begin{equation}
	\aucova(f) = \nicefrac{1}{N_C}\sumi\suminj (\nicefrac{p_j}{1-p_i}) \cdot \aucij(f^{(i)}).
\end{equation}
\item[(b)] 
\begin{equation*}
\begin{split}
&\aucova(f)
=\aucovo(f),~ \text{when} ~~  p_i =\nicefrac{1}{N_C}, ~~~i=1,2,\cdots, N_C.
\end{split}
\end{equation*}
\item[(c)] We  have $\aucova(f) = 1$ if and only if $\aucovo(f) = 1.$
\end{enumerate}
\end{rthm1}
\begin{pf}
{\color{white}dsdsadsa}\\
\textbf{Proof of (a)}. :For any event $\mathcal{C}$, we have:

\begin{equation*}
\begin{split}
\mathbb{P}\left[\mathcal{C},\mathcal{E}^{(i)}\right]  = \suminj \mathbb{P}\left[\mathcal{C}|\mathcal{E}^{(ij)}\right]\mathbb{P}\left[\mathcal{E}^{(ij)}\right] .
\end{split}
\end{equation*}

  \begin{equation*}
  \begin{split}
      \aucinj &= \mathbb{P}\left[(\yif{1} - \yif{2}) \cdot (f^{(i)}(\x_1) - f^{(i)}(\x_2))> 0 |\mathcal{E}^{(i)}\right] + \frac{1}{2}\mathbb{P}\left[ f^{(i)}(\x_1) - f^{(i)}(\x_2)= 0 |\mathcal{E}^{(i)}\right]\\
      & = \suminj \frac{\mathbb{P}\left[\mathcal{E}^{(ij)}\right]}{\mathbb{P}\left[\mathcal{E}^{(i)}\right]} \cdot \bigg( \mathbb{P}\left[(\yif{1} - \yif{2}) \cdot (f^{(i)}(\x_1) - f^{(i)}(\x_2))> 0 |\mathcal{E}^{(ij)}\right] \\
      & ~~~~+\frac{1}{2}\mathbb{P}\left[ f^{(i)}(\x_1) - f^{(i)}(\x_2)= 0 |\mathcal{E}^{(ij)}\right] \bigg)\\
      & = \suminj \frac{\mathbb{P}\left[\mathcal{E}^{(ij)}\right]}{\mathbb{P}\left[\mathcal{E}^{(i)}\right]} \aucij \\
      & = \suminj \frac{p_j}{1-p_i} \aucij
  \end{split}
    \end{equation*}
 It then follows that :
 \begin{eqnarray*}
 \aucovo&&= \dfrac{1}{N_C}\sumi \aucinj \\
 &&  = \dfrac{1}{N_C} \sumi \suminj \frac{p_j}{1-p_i} \aucij.
 \end{eqnarray*} 
\textbf{Proof of (b)}.
When $p_i = \frac{1}{N_C},~ \forall i =1,2,\cdots, N_C$, we have  $\frac{p_j}{1-p_i} = \frac{p_i}{1-p_j} = \frac{1}{N_C -1} $, which completes the proof with the conclusion of (a).\\
\textbf{Proof of (c)}. 
Since $\aucovo$ and $\aucova$ are convex combinations of $\aucij$, and that $\aucij \in [0,1]$, we have \[\aucovo = 1 \leftrightarrow \aucij = 1, ~\forall i \neq j \leftrightarrow \aucova =1 .\]
\qed
\end{pf}

\section{Consistency Analysis}
In this section, we will provide all the proofs for Consistency Analysis.

\subsection{The Bayes Optimal Scoring Function}
First, we show the derivation of the Bayes optimal scoring function set under the $\mauc$ criterion.
\begin{rthm2}
Given $\eta_i(\cdot) = \mathbb{P}[y = i|x]$, $p_i= \mathbb{P}[y = i]$,  we have the following consequences:
\begin{enumerate}
\item[(a)] $f= \{f^{(i)}\}_{i= 1,2,\cdots, N_C} \in \mathcal{F}^{N_C}_\sigma$ is a Bayes optimal scoring function under the $\mauc$ criterion, if:
\begin{equation*}
\left(f^{(i)}(\bm{x}_1) - f^{(i)}(\bm{x}_2)\right) \cdot  \left(\pione - \pitwo \right) > 0, ~ \forall \pione \neq \pitwo,
\end{equation*}
where:
\begin{equation*}
\pione =  \sum_{j \neq i} \frac{\eta_i(\bm{x}_1)\eta_j(\bm{x}_2)}{2p_ip_j}, ~~ \pitwo =  \sum_{j \neq i} \frac{\eta_j(\bm{x}_1)\eta_i(\bm{x}_2)}{2p_ip_j}.\\ 
\end{equation*}

\item[(b)] Define $\sigma(\cdot)$ as the sigmoid function, $s_i(\x)= \eta_i(\x)/p_i$ and $s_{\backslash i}(\x) = \sum_{j \neq i } s_j(\x)$,  then a Bayesian scoring function could be given by:
 \begin{equation}
 f^{\star(i)}(\x) = \begin{cases}
 \sigma\left(\dfrac{s_i(\x)}{s_{\backslash i}(\x)}\right), & 0 \le \eta_i(\x) < 1\\
 1,& \eta_i(\x) = 1. 
 \end{cases}
 \end{equation}
\end{enumerate}

\end{rthm2}
\vspace{3ex}

\begin{pf} {\color{white}dasdsa}\\
Proof of \textbf{(a)}\\
First let us summarize the notations we will use in the forthcoming derivation in Tab.\ref{tab:note}.

\begin{table}[htbp]
\centering
 \caption{\label{tab:note}Notations and descriptions.}
 \begin{tabular}{ll}
  Notation &Description  \\
  \toprule
$\mathcal{Y}_{ij}$ & is the event $[y_1 = i, y_2 = j]$ \\
\midrule
$\mathcal{E}_{i,j}$ &  is the event  $\mathcal{Y}_{ij} ~\text{or}~ \mathcal{Y}_{ji}$ \\
\midrule
$y^{(i)}_{1}$&  $=1$ if $y_1 = i$, otherwise $y^{(i)}_{1}  = 0$\\
\midrule
$y^{(i)}_{2}$&  $=1$ if $y_2 = i$, otherwise $y^{(i)}_{2}  = 0$\\
\midrule
$\x_{1,2}$& refers to the couple $(\x_1, \x_2)$\\
\midrule
$y_{1,2}$& refers to the couple $(y_1, y_2)$\\
\midrule

$\mathcal{G}_{i}(y_{1,2}, \x_{1,2})$& is event that $ ( f^{(i)}(\x_1) - f^{(i)}(\x_2) ) \cdot  (y^{(i)}_{1} - y^{(i)}_{2}) <0 $\\
\midrule
$\mathcal{G}_{i}\left((1, 0), \x_{1,2}\right)$& is event that  $( f^{(i)}(\x_1) - f^{(i)}(\x_2) ) \cdot (1 - 0) <0 $\\
\midrule
$\mathcal{G}_{i}\left((0, 1), \x_{1,2}\right)$& is event that $ ( f^{(i)}\x_1) - f^{(i)}(\x_2) ) \cdot  (0 - 1) <0 $\\
\midrule
$\mathcal{G}_{0,i}(\x_{1,2})$&  is the event that  $ f^{(i)}(\x_1) = f^{(i)}(\x_2)$ \\
\midrule
$\eta_i(\x)$& $\mathbb{P}( y = i | \x )$\\
\midrule
$\pi^{(i)}(\x_1, \x_2)$ & $\sum_{j\neq i}  (\frac{\eta_i(\x_1)  \eta_j(\x_2)} {2p_ip_j})$\\
\midrule
$\pi^{(i)}(\x_2, \x_1)$ &  $\sum_{j\neq i}  (\frac{\eta_j(\x_1)  \eta_i(\x_2)} {2p_ip_j})$ \\
 \bottomrule
 \end{tabular}
\end{table}

First of all, we prove that the condition holds for:
\begin{equation*}
  f \in \arginf_{f \in \mathcal{M}} R(f)
\end{equation*}

where
\begin{equation*}
  \mathcal{M} = \{f = (f^{(1)},\cdots, f^{(N_C)}),  f^{(i)} ~\text{is a measurable function}.\}
\end{equation*}

First of all, we can remove the dependence on the conditioning of $\mathcal{E}_{i,j}$. We have
\begin{equation}\label{eq:cons1}
\begin{split}
N_C \cdot (N_C -1 ) \cdot \mauc &= \sum_{i = 1}^{N_C} \sum_{j \neq i} \bigg( \mathbb{P}\big[\mathcal{G}_{i}(y_{1,2}, \x_{1,2}) | \mathcal{E}_{i,j}  \big] + \frac{1}{2}  \cdot  \mathbb{P} \big[\mathcal{G}_{0,i}(\x_{1,2}) | \mathcal{E}_{i,j} \big] \bigg) \\
&= \sum_{i = 1}^{N_C} \sum_{j \neq i} \dfrac{\mathbb{P}\big[\mathcal{G}_{i}(y_{1,2}, \x_{1,2}), \mathcal{E}_{i,j}\big]}{\mathbb{P} \left[ \mathcal{E}_{i,j} \right]}   + \frac{1}{2}  \dfrac{\mathbb{P}\big[ \mathcal{G}_{0,i}(\x_{1,2}), \mathcal{E}_{i,j} \big]}{\mathbb{P} \left[ \mathcal{E}_{i,j} \right]}\\ 
&= \sum_{i = 1}^{N_C} \sum_{j \neq i} \dfrac{\mathbb{P}\big[\mathcal{G}_{i}(y_{1,2}, \x_{1,2}), \mathcal{E}_{i,j}\big]}{2p_ip_j}   + \frac{1}{2}  \dfrac{\mathbb{P}\big[ \mathcal{G}_{0,i}(\x_{1,2}), \mathcal{E}_{i,j} \big]}{{2p_ip_j} }
\end{split}
\end{equation}

Now we only need to expand the joint possibility by expectation, which leads to:

\begin{equation}\label{eq:cons2}
\begin{split}
\dfrac{\mathbb{P}\big[ \mathcal{G}_{i}(y_{1,2}, \x_{1,2}), \mathcal{E}_{i,j} \big]}{2p_ip_j} &=  \dfrac{\expe_{ \x_{1,2}, y_{1,2} } \big[ \bm{I}[ \mathcal{G}_{i}(y_{1,2}, \x_{1,2}) ] \cdot \bm{I}[ \mathcal{E}_{i,j} ] \big]} { 2p_ip_j}\\
&= \dfrac{\expe_{\x_{1,2}} \bigg[\expe_{y_{1,2}|\x_{1,2} } \big[ \bm{I}[ \mathcal{G}_{i}(y_{1,2}, \x_{1,2}) ] \cdot \bm{I}[ \mathcal{E}_{i,j} ]  \big] \bigg]}{2p_ip_j}\\
&=  \dfrac{\expe_{\x_{1,2}} \bigg[  \eta_i(\x_1) \cdot \eta_j(\x_2)   \cdot \bm{I}[ \mathcal{G}_{i}\left((1, 0), \x_{1,2}\right) ] +   \eta_j(\x_1) \cdot \eta_i(\x_2)  \cdot \bm{I}[ \mathcal{G}_{i}\left((0, 1), \x_{1,2}\right) ]    \bigg]}{2p_ip_j}.
\end{split}
\end{equation}
Similarly, we have:

\begin{equation}\label{eq:cons3}
 \dfrac{\mathbb{P}\big[ \mathcal{G}_{0,i}(\x_{1,2}), \mathcal{E}_{i,j} )\big]}{2p_ip_j} = \dfrac{\expe_{\x_{1,2}} \bigg[\eta_i(\x_1) \cdot \eta_j(\x_2)   \cdot \bm{I}\big[ \mathcal{G}_{0,i}(\x_{1,2})\big] +   \eta_j(\x_1) \cdot \eta_i(\x_2)  \cdot \bm{I}\big[ \mathcal{G}_{0,i}(\x_{1,2})\big]\bigg]}{2p_ip_j}       
\end{equation}

Above all, by combining Eq.(\ref{eq:cons1})-Eq.(\ref{eq:cons3}), we reach that :

\begin{equation}
\begin{split}
N_C \cdot (N_C -1 ) \cdot \mauc  = &\sum_{i = 1}^{N_C} \expe_{x_{1,2}} \bigg[ \pi^{(i)}(\x_1, \x_2)  \cdot \bm{I}\big[ \mathcal{G}_{i}\left((1, 0), \x_{1,2}\right) \big]  +  \pi^{(i)}(\x_2, \x_1) )\cdot \bm{I}\big[  \mathcal{G}_{i}\left((0, 1), \x_{1,2}\right)  \big] \\
&+ \frac{1}{2} \cdot \big( \pi^{(i)}(\x_1, \x_2) + \pi^{(i)}(\x_2, \x_1) \big) \cdot I\big[ \mathcal{G}_{0,i}(\x_{1,2})\big] \bigg].
\end{split}
\end{equation}

Fix a sub-scoring function $f^{(i)}$ and $\x_{1,2}$, we only need to minimize the following quantity $L^{(i)}$ to reach the Bayes optimal solution:

\begin{equation}
 L^{(i)} =   \pi^{(i)}(\x_1, \x_2)  \cdot \bm{I}\big[ \mathcal{G}_{i}\left((1, 0), \x_{1,2}\right) \big]  +  \pi^{(i)}(\x_2, \x_1) )\cdot \bm{I}\big[  \mathcal{G}_{i}\left((0, 1), \x_{1,2}\right)  \big] + \frac{1}{2} \cdot \big( \pi^{(i)}(\x_1, \x_2) + \pi^{(i)}(\x_2, \x_1) \big) \cdot I\big[ \mathcal{G}_{0,i}(\x_{1,2})\big]
\end{equation} 

\noindent Obviously 
,
\begin{equation}L^{(i)}(\x_1,\x_2) = 
\begin{cases}
\pitwo, &f^{(i)}(\x_1) -f^{(i)}(\x_2) > 0 ,\\
\\
\dfrac{\pione+\pitwo}{2}, &f^{(i)}(\x_1) -f^{(i)}(\x_2) = 0,\\
\\
\pione, & f^{(i)}(\x_1) -f^{(i)}(\x_2) < 0.
\end{cases}
\end{equation}

\noindent To obtain the Bayes optimal scoring function, we must minimize $L(\x_1,\x_2)$ for any $\x_1,\x_2 \in \mathcal{X}$.

\noindent When $\pione = \pitwo$, $L(\x_1,\x_2)$ stays as a constant, which is irrelevant to the choice of $f^{(i)}$.

\noindent When $\pione < \pitwo$, we must have $f^{(i)}(\x_1) -f^{(i)}(\x_2) < 0$ to minimize $L(\x_1,\x_2)$.

\noindent  When $\pione > \pitwo$, we must have $f^{(i)}(\x_1) -f^{(i)}(\x_2) > 0$ to minimize $L(\x_1,\x_2)$.

\noindent Above all, by choosing $f^{(i)}$ such that $\forall \bm{x}_1, \bm{x}_2 \in \mathcal{X}, \pione \neq \pitwo$
\begin{equation*}
\left(f^{(i)}(\bm{x}_1) - f^{(i)}(\bm{x}_2)\right) \cdot  \left(\pione - \pitwo \right) > 0,
\end{equation*}
one can reach the minimum of $R^{(i)}(f^{(i)})$ as $\exx{\min\{\pione,\pitwo\}}.$

This ends the proof of the condition for:
\begin{equation*}
  f \in \arginf_{f \in \mathcal{M}} R(f)
\end{equation*}

To extend the conclusion to:
\begin{equation*}
  f \in \arginf_{f \in \mathcal{F}^{N_C}_\sigma} R(f)
\end{equation*}

One only need to notice that if $f^{(i)}(\x_1) -f^{(i)}(\x_2) > 0$, then for any strictly increasing measurable function $\varphi$, we have:
$\varphi(f^{(i)}(\x_1)) -\varphi(f^{(i)}(\x_2)) > 0$. Obvisously, a similar result holds for $f^{(i)}(\x_1) -f^{(i)}(\x_2) < 0$. This implies if $f \in \mathcal{M}$ is Bayes optimal, then
 \[\varphi \circ f = (\varphi\circ f^{(1)},\cdots,\varphi\circ f^{(N_C)} )\]
 is also Bayes optimal with $\varphi \circ f \in \mathcal{F}^{N_C}_\sigma$. The proof is then completed.
\\

Proof of \textbf{(b)}\\
Fixing class $i$, if $\eta_i(\x_k) \neq 1,~ k=1,2$, then by dividing $s_{\backslash i}(\x_1)\cdot s_{\backslash i}(\x_2)$ in Eq.(\ref{eq:optbay}) in Thm.\ref{prop:bay}, we find that $f^{\star(i)}(\x)$ could be chosen as $\sigma(s_i(\x) / s_{\backslash i}(\x))$.\\
Otherwise, if $\eta_i(\x_1) = 1, \eta_i(\x_2) <1 $ (note that $\pione \neq \pitwo$) then $s_{\backslash i}(\x_1) = 0$. To reach  $f^{\star(i)}(\x_1) > f^{\star(i)}(\x_2)$, we can set $f^{\star(i)}(\x_1) =1$, $f^{\star(i)}(\x_2) =\sigma\left(s_i(\x_2) / s_{\backslash i}(\x_2)\right)$.\\ 
In the last case, we have $\eta_i(\x_1) < 1, \eta_i(\x_2) =1 $, we can reach that an optimal solution could be  $f(\x_1) = \sigma\left(s_i(\x_1) / s_{\backslash i}(\x_1)\right), f(\x_2) =1$.\\
Since $\x_1$, $\x_2$ is arbitrarily chosen, a Bayesian scoring function could be then set as: $f^{\star(i)}(\x) = \sigma(s_i(\x) / s_{\backslash i}(\x))$($<1$) if $\eta_i(\x) < 1$; $f^{\star(i)}(\x) = 1$ otherwise.
\qed
\end{pf}
\vspace{3ex}

\subsection{Consistency Analysis of Surrogate Losses}\label{sec:appb2}

Based on Thm.\ref{prop:bay}, we provide the following result for some popular surrogate losses. 
\begin{rthm3}
The surrogate loss $\ell$ is consistent with $\mauc$ if it is differentiable, convex, nonincreasing within $[-1, 1]$ and $\ell'(0) < 0$.
\end{rthm3}
\begin{pf}

First, recall that we have the following definition for the surrogate risk:
  \begin{equation*}
    \begin{split}
    &R_\ell(f) = \sum_{i} \frac{R^{(i)}_\ell(f^{(i)})}{N_C(N_C-1)}\\
    &R^{(i)}_\ell(f^{(i)}) = \sum_{j\neq i} \ezz\left[{\ell(\Delta(\yi)\Delta{f}^{(i)})|\eij}\right],
    \end{split}
    \end{equation*}
where $R^{(i)}_\ell(f^{(i)})$ is the corresponding risk for $f^{(i)}$.

  Denote:
\begin{equation*}
  f^\star = \arginf_{f \in \mathcal{F}_\sigma} R_\ell(f)
\end{equation*}
with $f^\star= (f^{\star(1)}, f^{\star(2)},\cdots, f^{\star(N_C)})$.

Moreover, it is obvious that:
  \begin{equation}
  \inf_{f \in \mathcal{F}_\sigma} R_{\ell}(f) = \nc \cdot \sum_{i} \inf_{f^{(i)} \in \mathcal{F}_\sigma } R^{(i)}_{\ell}(f^{(i)}).
  \end{equation}

In this sense, each binary scoring function $f^{\star(i)}$ could be solved independently for different subproblems, \emph{i.e.} 

\begin{equation*}
  f^{\star(i)} = \arginf_{f^{(i)} \in \mathcal{F}_\sigma}  R^{(i)}_{\ell}(f^{(i)})
\end{equation*}

  Similar to the derivation in Thm.\ref{prop:bay},  we could rewrite $\Rl$ as:
  \begin{equation*}
    \Rl(f^{(i)}) =  \frac{1}{2}  \int_{\mathcal{X}}\int_{\mathcal{X}} \left[\epx \cdot \enxp \cdot \ell(\fif{\x} -\fif{x'}) +  \epxp \cdot \enx  \cdot \ell(\fif{\x'} - \fif{\x}) \right] d\mathbb{P}(\x) d\mathbb{P}(\x')
  \end{equation*}
where 
\[\epx  = \frac{\eta_i(\x)}{p_i}, ~\enx  = \suminj\frac{\eta_j(\x)}{p_j}.\]

Denote $\mathsf{Bayes}^{(i)}_\sigma$ as the set of $f^{(i)}$, where $f$ is a Bayes optimal socring functions, \emph{i.e.}
\begin{equation*}
  \mathsf{Bayes}^{(i)}_\sigma = \{f^{(i)}: \left(f^{(i)}(\x_1) - f^{(i)}(\x_2) \right)  \cdot  (\pione - \pitwo) > 0, ~\forall (\x_1,\x_2) ~s.t.~ \pione \neq \pitwo\}
\end{equation*}

Now proof the theorem with the following steps.

\noindent\rule[0.15\baselineskip]{\columnwidth}{1pt}

\begin{clm}\label{clm:1}
  $f^{\star(i)} \in \mathsf{Bayes}^{(i)}_\sigma$
\end{clm}
\noindent\rule[0.15\baselineskip]{\columnwidth}{1pt}

We prove the claim by contradiction.  To do this,  we assume that
claim does not hold in the sense that  $\exists \x_1, \x_2$ s.t. $\fifs{\x_1} \le \fifs{\x_2}$ but $\pione > \pitwo$.

\noindent \textbf{Case 1: $\enxone >0, \enxtwo >0$}

Define an intermediate function $\delta_h(\gamma) = \Rl(f^{\star(i)} +\gamma h)$, we must have  $\delta'_h(0) = 0$, $\forall h$.

Given any pair of instance $(\x_1, \x_2) \in \mathcal{X}$, with $\x_1\neq \x_2$, and for all $\ell$ satisfying the sufficient condition, we prove that $\fifs{\x_1} > \fifs{\x_2}$ must imply that $\pione > \pitwo$ by contradiction.

Picking 
\[h_1(x)  = \frac{\bm{I}\left[x =x_1\right]}{\eta_-(\x)},\]

using $\delta'_{h_1}(0) = 0$,  we have:
\begin{equation}\label{eq:opt1}
  \int_{X\backslash\x_1} \ratone \cdot \enx \cdot \ldiff{\x_1}{\x} - 
  \epx \cdot \ldiff{\x}{\x_1} d\mathbb{P}(\x) =0
\end{equation}

Picking 
\[h_2(x)  = \frac{\bm{I}\left[x =x_2\right]}{\eta_-(\x)},\]

using $\delta'_{h_2}(0) = 0$,  we have:
\begin{equation}\label{eq:opt2}
  \int_{X\backslash\x_2} \rattwo \cdot \enx \cdot \ldiff{\x_2}{\x} - 
  \epx \cdot \ldiff{\x}{\x_2} d\mathbb{P}(\x) =0
\end{equation}

Taking \eqref{eq:opt1} - \eqref{eq:opt2}, we have:

\begin{equation}\label{eq:base}
\begin{split}
  &\underbrace{\int_{X\backslash\{\x_1,\x_2\}}  \epx \cdot \left(\ldiff{\x}{\x_2} -\ldiff{\x}{\x_1} \right)d\mathbb{P}(\x)}_{\bm{(a)}} \\ 
  &+\underbrace{\int_{X\backslash\{\x_1,\x_2\}}  \frac{\enx }{\enxtwo\enxone}  \left(\pione \cdot \ldiff{\x_1}{\x} - \pitwo \cdot \ldiff{\x_2}{\x} \right)d\mathbb{P}(\x)}_{\bm{(b)}}\\ 
  &+\underbrace{(\frac{\mathbb{P}(\x_1)}{\enxone} + \frac{\mathbb{P}(\x_2)}{\enxtwo}) \cdot \left(\pione \cdot \ldiff{\x_1}{\x} - \pitwo \cdot \ldiff{\x_2}{\x} \right)}_{\bm{(c)}}\\ 
  &=0
\end{split}
\end{equation}
We now construct a contradiction with the equation above.

First notice that
\begin{equation*}
  \ldiff{\x}{\x_2} \le  \ldiff{\x}{\x_1}
\end{equation*}
since by assumption $\ell$ is nonconvex, \emph{i.e.}, $\ell'$ is nondecreasing. This implies $\bm{(a)} \le 0$. 

With a similar spirit, we have:

\begin{equation*}
  \ldiff{\x_1}{\x} \le  \ldiff{\x_2}{\x} \le 0.
\end{equation*}
Together with $\pione > \pitwo$, we have $\bm{(b)}\le 0$.

Now, we prove that $\bm{(c)} < 0$:

\noindent \textbf{Case a}: If $\fifs{\x_1} = \fifs{\x_2}$, we have :
\begin{equation*}
  \bm{(c)} = (\pione - \pitwo)\cdot \ell'(0) < 0
\end{equation*}
since $\ell(0) <0$.

\noindent \textbf{Case b} If $\fifs{\x_1} < \fifs{\x_2}$, we have:
\begin{equation*}
  \ldiff{\x_1}{\x_2} \le  \ell'(0) < 0,
\end{equation*}
\begin{equation*}
  \ldiff{\x_2}{\x_1} \le 0,
\end{equation*}
and $\pitwo > \pione$, we have:$\bm{(c)} <0$

Putting all togther, we have that $\bm{(a) + (b) +(c)} <0$, which contradicts with eq.\eqref{eq:base}. Since the choice of $(\x_1,\x_2)$ is arbitrary, this shows that
\[f^{\star(i)} \in \mathsf{Bayes}^{(i)}_\sigma, ~\text{if} ~\enxone >0, \enxtwo >0 \]

\noindent \textbf{Case 2: $\enxone =0, \enxtwo >0$}

Since :

\begin{equation*}
  \sum_i  \mathbb{P}\left[\x_1, y= i\right]  =  \sum_i \eta_i(\x_1) \mathbb{P}(\x_1) = \mathbb{P}(\x_1)
\end{equation*}
We have :
\begin{equation*}
   \sum_i \eta_i(\x_1)  = 1.
\end{equation*}

This further implies that $\eta_i(\x_1)$ reaches its maximum 1, which means $\epxone >  \epxtwo$.

Similar to the proof of Case 1, we can set $h_1(\x) = \bm{I}\left[ \x = \x_1 \right],~ h_2(\x) = \bm{I}\left[ \x = \x_2 \right]$ and obtain the following equation.

\begin{equation}\label{eq:base}
  \begin{split}
    &\underbrace{\int_{X\backslash\{\x_1,\x_2\}}  \enx \cdot  \left(\epxone \cdot \ldiff{\x}{\x_1} - \epxtwo \cdot \ldiff{\x}{\x_2} \right)  d\mathbb{P}(\x)}_{\bm{(a)}} \\ 
    &+\underbrace{\int_{X\backslash\{\x_1,\x_2\}}  \epx  \cdot \enxtwo \cdot  \ldiff{\x_2}{\x} d\mathbb{P}(\x)}_{\bm{(b)}}\\ 
    &+\underbrace{(\mathbb{P}(\x_1) + \mathbb{P}(\x_2)) \cdot \epxone \cdot \enxtwo \cdot \ldiff{\x_1}{\x_2} }_{\bm{(c)}}\\ 
    &=0
  \end{split}
  \end{equation}
Similarly, one can show that $\bm{(a) + (b) +(c)} <0$, which contradicts with the optimal condition.

\[f^{\star(i)} \in \mathsf{Bayes}^{(i)}_\sigma, ~\text{if} ~\enxone =0, \enxtwo >0 \]

\noindent \textbf{Case 3: $\enxone >0, \enxtwo =0$.} In this sense, we have  $\pione < \pitwo$, which contradicts with the assumption that $\pione > \pitwo$. \textbf{Thus this is impossible to observe.} 

\noindent \textbf{Case 4: $\enxone =0, \enxtwo =0$.} We have  $\pione = \pitwo$, which contradicts with the assumption that $\pione > \pitwo$.  \textbf{Thus this is impossible to observe.} 

Then the proof of Claim \ref{clm:1} is completed.

 \noindent\rule[0.15\baselineskip]{\columnwidth}{1pt}

 \begin{clm}\label{clm:2}
 
\begin{equation}\label{eq:subopt}
  \inf_{f^{(i)} \notin  \mathsf{Bayes}_\sigma } \Rl(f^{(i)}) >   \inf_{f^{(i)} \in  \mathcal{F}_\sigma } \Rl(f^{(i)})
\end{equation}

\end{clm}
\noindent\rule[0.15\baselineskip]{\columnwidth}{1pt}

Claim \ref{clm:2} follows that $\arginf_{f^{(i)} \in  \mathcal{F}_\sigma } \Rl(f^{(i)}) = \mathsf{Bayes}^{(i)}_\sigma.$

\noindent\rule[0.15\baselineskip]{\columnwidth}{1pt}

\begin{clm}\label{clm:3}
 
  For any sequence $\{f_t\}_{ t\in \mathbb{N}_+}$, such that $f_i \in \mathcal{F}_\sigma$, 

  \[\Rl(f_t) \rightarrow \Rl(f^{\star(i)}) ~\text{implies}~ R^{(i)}(f_t) \rightarrow \inf_{f \in \mathcal{F}_\sigma} R^{(i)}(f)  \] 
  
  \end{clm}
  \noindent\rule[0.15\baselineskip]{\columnwidth}{1pt}

Since $\arginf_{f \in \mathcal{F}_\sigma} R^{(i)}(f) =  \mathsf{Bayes}^{(i)}_\sigma$, we only need to prove that $\lim_{t\rightarrow \infty} f_t \in  \mathsf{Bayes}^{(i)}_\sigma$. Define $\delta^{(i)}$ as:
\begin{equation*}
  \delta^{(i)} = \inf_{f^{(i)} \notin  \mathsf{Bayes}^{(i)}_\sigma } \Rl(f^{(i)}) -  \inf_{f^{(i)} \in  \mathcal{F}_\sigma } \Rl(f^{(i)}) >0
\end{equation*}
Suppose that $\lim f_t \notin \mathcal{F}_\sigma$, then for large enough $T$, we have:
\begin{equation*}
  \Rl(f_T) - \Rl(f^{\star(i)}) > \delta^{(i)},
\end{equation*}
which contradicts with the fact that $\Rl(f_t) \rightarrow \Rl(f^{\star{(i)}})$. This shows that Eq.\eqref{eq:subopt} holds.

Since the argument above is based on an arbitrarily chosen class $i$, Eq.\eqref{eq:subopt} holds for all i.
\noindent\rule[0.15\baselineskip]{\columnwidth}{1pt}

\begin{clm}\label{clm:3}
  Given a sequence $\{f\}_{t}$ with $f_t = (f^{(1)}_t,\cdots, f^{(N_C)}_t)$, we have:

  \[R_\ell(f_t) \rightarrow R_\ell(f) ~\text{implies}~ R(f_t) \rightarrow \inf_{f \in \mathcal{F}_\sigma} R(f)  \]

  \end{clm}

  \noindent\rule[0.15\baselineskip]{\columnwidth}{1pt}

This directly follows that:

\begin{equation*}
  R(f_t) = \frac{\sum_{i=1}^{N_C} R^{(i)}(f^{(i)}_t) }{N_C\cdot(N_C-1)},~   R_\ell(f_t) = \frac{\sum_{i=1}^{N_C} R^{(i)}_\ell(f^{(i)}_t) }{N_C\cdot(N_C-1)}.
\end{equation*}
and Claim \ref{clm:3}.

The proof is then completed according to Def.\ref{def:cons}. \qed
\end{pf}

\vspace{2ex}

\section{Unbiased Estimation of $R_\ell$}\label{sec:appc}
In this section, we provide a derivation of an unbiased estimation of  $R_\ell$.
\begin{rprop1}  $\Rhatf$ as:
\[\Rhatf = \sumi \suminj \sumxone \sumxtwo \ninj \ell(f^{(i)}(\x_m) - f^{(i)}(\x_n) ).\]
  $\Rhatf$ is an unbiased estimation of ${R}_{\ell}(f)$, in the sense that:
${R}_{\ell}(f) = \expe\limits_{\mathcal{S}}(\Rhatf)$, where $\mathcal{N}_i$ denotes the set of all samples in $\mathcal{S}$ with $y = i$.
\end{rprop1}
\begin{pf}
\begin{equation*}
\expe_{\mathcal{S}}(\Rhatf) = \expe_{{\bm{Y}}}\left[\expe_{\bm{X}|{\bm{Y}}}(\Rhatf)\right]
\end{equation*}
Moreover, we have 
\begin{equation*}
\begin{split}
 &\expe_{\bm{X}|{\bm{Y}}}(\Rhatf)\\
& ~~~=   \frac{1}{N_C(N_C-1)}\sumi \suminj \sum_{\x_1 \in \mathcal{N}_i} \sum_{\x_2 \in \mathcal{N}_j} \ninj \exx \left[\ell(f^{(i)},\x_1,\x_2)|y_1 = i, y_2 = j\right]\\
&~~~~ = \frac{1}{N_C(N_C-1)}\sumi \suminj  \exx\left[\ell(f^{(i)},\x_1,\x_2)|y_1 = i, y_2 = j\right].
\end{split}
\end{equation*}

Since  $\expe_{\bm{x}_1,\bm{x}_2}\left[\ell(f^{(i)},\x_1,\x_2)|y_1 = i, y_2 = j\right]$ does not depend on the distribution of $\bm{Y}$, we have:
\begin{equation*}
\expe_{{\bm{Y}}}\left[\expe_{\bm{X}|{\bm{Y}}}(\Rhatf)\right] = \expe_{\bm{X}|{\bm{Y}}}(\Rhatf).
\end{equation*}
Now it only remains to prove that 
\[\exx\left[\ell\left(f^{(i)}(\x_1)-f^{(i)}(\x_2)\right)|y_1 = i, y_2 = j\right] = \ezz\left[\ell\left(\Delta(y^{(i)}) \Delta(f^{(i)})\right)|\eij\right]. \]

To see this, we have:

\begin{equation*}
\begin{split}
&\ezz\left[\ell\left(\Delta(y^{(i)}) \Delta(f^{(i)})\right)|\eij\right] \\ \\
& ~~{=}  \ezz{\left[\pione \ell\left(f^{(i)}(\x_1)-f^{(i)}(\x_2)\right) + \pitwo \ell\left(f^{(i)}(\x_2)-f^{(i)}(\x_1)\right)\right]}\\ \\
& ~~= 2\exx{\left[\pione \ell\left(f^{(i)}(\x_1)-f^{(i)}(\x_2)\right)\right]}\\ \\ 
& ~~{=} \exx{\left[\frac{\eta^{(i)}(\x_1)\eta^{(j)}(\x_2)}{p_ip_j} \ell\left(f^{(i)}(\x_1)-f^{(i)}(\x_2)\right)\right]} \\ \\
& ~~{=} \exx\left[\ell\left(f^{(i)}(\x_1)-f^{(i)}(\x_2)\right)|y_1 = i, y_2 = j\right].
\end{split}
\end{equation*}
 The proof is thus completed. 
\qed
\end{pf}

\section{Preliminary for Generalization Analyses}\label{app:gen_pre}
\subsection{Concentration Inequalities}
In this subsection, we provide a brief introduction to the concentration bounds that are employed throughout the next two sections.
\subsubsection{Bounded Difference Property}
\begin{defi}[Bounded Difference Property]\label{def:bdp}
 Given independent random variables $X_1,\cdots, X_n$, with $X_i \in \mathbb{X}$, A function $f(X_1,X_2,\cdots, X_n)$ is said to has the bounded difference property if there exist non-negative constants $c_1, c_2,\cdots, c_n$, such that:
 \begin{equation}
   \sup_{x_1,x_2,\cdots,x_n, x'_i} \left|f(x_1,\cdots,x_n) - f(x_1,\cdots, x_{i-1},x'_i,\cdots,x_n)\right| \le c_i, ~ \forall  1 \le i \le n.
 \end{equation}
\end{defi}
For all functions satisfying the Bounded Difference Property, we have the following Bounded Difference Inequality over the generated momentum functions.
\begin{lem}[Bounded Difference Inequality]\label{lem:bdp} \cite[\color{org}{Prop.6.1, Thm.6.2}]{concen}
Assume that $Z= f(X_1,\cdots,X_n)$, with $X_i$s being independent, satisfies the bounded difference property with constants $c_1,c_2,\cdots,c_n$. Denote 
\begin{equation}
v = \frac{1}{4}\sum_{i=1}^n c_i^2
\end{equation}
then we have:
\begin{equation}
  \log\expe\left[\exp\left(\lambda(Z-\expe[Z])\right)\right] \le \frac{\lambda^2 v^2}{2},
\end{equation}
holds for all $\lambda > 0$.
\end{lem}
\subsubsection{Maximal Inequality}\label{lem:max}
\begin{lem}[Maximal Inequality] \cite[\color{org}{Sec.2.5}]{concen} Let $Z_1,\cdots,Z_n $ be real-valued random variables where a $v>0$ exists, such that for every $i=1,2,\cdots,n$, we have $\log\big(\expe\big[\exp\left(\lambda Z_i\right)\big]\big)\le \frac{\lambda v^2}{2}$, then we have:
  \begin{equation*}
    \expe\left[\max_{i=1,2,\cdots,n} Z_i\right] \le \sqrt{2v\log n}.
  \end{equation*}
  
\end{lem}
\subsubsection{Mcdiarmid Inequality}

First let us review the fundamental concentration inequality that is adopted in our proof.
\begin{lem}[Mcdiarmid inequality]\label{lem:mc} \cite{mc}
Let $X_1,\cdots,X_m$ be independent random variables all taking values in the set $\mathcal{X}$. Let $f: \mathcal{X} \rightarrow \mathbb{R}$ be a function of
  $X_1,\cdots,X_m$ that satisfies:
  \[\sup_{\boldsymbol{x},\boldsymbol{x}'}|f(x_1,\cdots,x_i,\cdots, x_m) - f(x_1, \cdots,x'_i \cdots,x_,) | \le c_i,\]
  with $\boldsymbol{x} \neq \boldsymbol{x}'$.
 Then for all $\epsilon >0$,
  \[\mathbb{P}[ \mathbb{E}(f) - f \ge \epsilon ] \le \exp\left(\dfrac{-2\epsilon^2}{\sum_{i=1}^mc_i^2} \right). \]
\end{lem}

\subsection{Properties of Rademacher Averages}
The following lemmas are instrumental for our derivation of $\mauc$ Rademacher Complexity upper bounds in Lem.\ref{lem:rlin} and Lem.\ref{lem:Rdn}.

First the Khinchin-Kahane inequality bridges the $\ell_p$ norm average to the $\ell_q$ norm average, given that $1<p<q<\infty$.
\begin{lem}[Khinchin-Kahane inequality]\label{lem:kk}\cite{tala}
$\forall n \in \mathbb{N}$, let $\sigma_1, \cdots, \sigma_n$ be a family of i.i.d. Rademacher variables. Then for any $1<p<q<\infty$, we have:
\begin{equation*}
\bigg(\expe_{\sigma} \left[\big|\sum_{i=1}^n\sigma_if_i\big|^q\right]\bigg)^{\nicefrac{1}{q}} \le \left(\frac{q-1}{p-1}\right)^{\nicefrac{1}{2}} \cdot \bigg(\expe_{\sigma} \left[\big|\sum_{i=1}^n\sigma_if_i\big|^p\right]\bigg)^{\nicefrac{1}{p}}.
\end{equation*}
\end{lem}

Dealing with the Rademacher complexities for  vector-valued function, we need the vector version of the well-known Talagrand contraction lemma, see \cite{vec1,vec2} for a proof.
\begin{lem}[vector version Talagrand contraction lemma]\label{lem:vec-con}
Let $\mathcal{X}$ be any set, $n \in \mathcal{N}$, $(\x_1; \cdots; \x_N) \in X_n$, let $\mathcal{F}$ be a class of k-component-functions $f = (f^{(1)},\cdots,f^{(k)}): X \rightarrow \ell_2^k$ and let $h:\ell_2^k \rightarrow \mathbb{R}$ have Lipschitz norm $\phi$. Then,
\[\expe_{\sigma}\left[\sup_{f\in\mathcal{F}} \sum_{i=1}^N \sigma_i h(f(\x_i))\right] \le \sqrt{2}\phi\expe_{\sigma}\left[\sup_{f\in\mathcal{F}} \sum_{i=1}^N\sum_{k=1}^K \sigma_{i,k} (f^{(k)}(\x_i))\right],\]
where $\sigma_1 \cdots,\sigma_N$, and $\sigma_{1,1}, \cdots, \sigma_{1,K}, \cdots, \sigma_{N,K}$ are two sequences of independent Rademacher random variables.
\end{lem}

The following Lemma is a simple extension of Lem.1 in \cite{deep-gen} to double indexed Rademacher random sequence, which is the key to the derivation of Lem.\ref{lem:Rdn}.
\begin{lem}\label{lem:exp}Let $s$ be a 1-Lipschitz, positive-homogeneous activation function which is applied element-wise. Then for any class of vector-valued functions $\mathcal{F}$, and any convex and monotonically increasing function $g : \mathbb{R} \rightarrow [0,\infty)$, we have:
\begin{equation*}
\expe_{\sigma} \left[\sup_{f \in \mathcal{F}, ||W||_F \le R_W}  g\left(\left\|\sum_{i=1}^N \sum_{c=1}^{N_C} \sigma_{i,c} \cdot   s(\bm{W}f(\x_i)) \right\| \right)\right] \le 2 \expe_{\sigma} \left[ \sup_{f \in \mathcal{F}} g\left(R_W \cdot \left\|\sum_{i=1}^N \sum_{c=1}^{N_C} \sigma_{i,c} \cdot f(\x_i) \right\| \right)\right]
\end{equation*}
where $\{\sigma_{i,c}\}$ is a sequence of double indexed Rademacher random variables.
\end{lem}

\subsection{Lipschitz Properties of Softmax}
In this section, we provide an $\ell_2$  Lipschitz Constant for softmax component function.
\begin{lem}[Lipschitz Constant for softmax components] \label{lem:soft} Given $\mathcal{X} \in \mathbb{R}^K$, the function $\mathsf{soft}_i,~ i = 1,2,\cdots,K$ is a mapping $\x = (x_1,\cdots, x_k) \in \mathcal{X} \rightarrow [0,1]$, defined as:
\[\mathsf{soft}_i(\x)  = \dfrac{\exp(x_i)}{\sum_{j=1}^K \exp(x_j)},\]
then $\mathsf{soft}_i(\cdot)$ is $\frac{\sqrt{2}}{2}$-Lipschitz continuous with respect to vector $\ell_2$ norm.
\end{lem}
\begin{pf}
We only  need to prove that 
\[\sup_{\x \in \mathcal{X}} \big\| \nabla_{\x}\mathsf{soft}_i(\x)\big\|_2 \le \frac{\sqrt{2}}{2}. \]
For any $\x \in \mathcal{X}$, we have:

\[\frac{\partial \mathsf{soft}_i(\x)}{\partial x_j} =  \mathsf{soft}_i(\x) \cdot \left(I[i = j] -\mathsf{soft}_{j}(\x) \right), ~ i,j= 1,\cdots,K  .\]
Then we have:
\begin{equation*}
\begin{split}
\big\| \nabla_{\x}\mathsf{soft}_i(\x)\big\|_2 = \left( \mathsf{soft}^2_i(\x)\cdot \sum_{j\neq i}\mathsf{soft}^2_j(\x)+ \mathsf{soft}_i(\x)^2\cdot \left(1-\mathsf{soft}_{j}(\x)\right)^2\right)^{1/2}.
\end{split}
\end{equation*}
Since $ \mathsf{soft}^2_i(\x) \le  \mathsf{soft}_i(\x)$, $\left(1-\mathsf{soft}_{j}(\x)\right)^2 \le \left(1-\mathsf{soft}_{j}(\x)\right)$, and $\sum_{j\neq i}\mathsf{soft}^2_j(\x) \le (1 - \mathsf{soft}_i(\x))$, we have:
\begin{equation*}
\big\| \nabla_{\x}\mathsf{soft}_i(\x)\big\|_2 \le \bigg(2 \cdot \mathsf{soft}_i(\x) \cdot \left(1 - \mathsf{soft}_i(\x)\right)\bigg)^{1/2} \le\frac{\sqrt{2}}{2},
\end{equation*}
which ends the proof.

\end{pf}

\section{$\mauc$ Rademacher Complexity and Its Properties}\label{app:gen_bound}

\subsection{$\mauc$ Symmetrization }

In this section, we provide the derivation for $\mauc$ Symmetrization, which is the key technology for Thm.\ref{thm:gen}. The main idea is that exchanging instances, but not the terms, does not change the value of \[\ess{\left[ \supf \left(\js - \jsp\right)\right]}.\]
\begin{lem}[$\mauc$ symmetrization]\label{lem:perm}
The following inequality holds:
\[\ess{\left[ \supf \left(\js - \jsp\right)\right] } \le  \frac{4  \sR}{N_C(N_C-1)},\]
with the labels $\bm{Y}$ fixed for $\mathcal{S}$ and $\mathcal{S}'$.
\end{lem}
\begin{pf} Denote $\lfxy{\bm{x}}{\xp}  = \ell(f^{(i)}(\x_m)  - f^{(i)}(\x_n'))$ and define 
\begin{equation}
\begin{split}
T^{i,j,m,n} =&\frac{\sigmai_m + \sigmaj_n}{2} \lfxy{\xp}{\xp} +\frac{\sigmai_m - \sigmaj_n}{2} \lfxy{\xp}{\x} \\
 &-\frac{\sigmai_m - \sigmaj_n}{2} \lfxy{\x}{\xp}  -\frac{\sigmai_m + \sigmaj_n}{2} \lfxy{\x}{\x}, 
\end{split}
\end{equation}
we first show that
\begin{equation}
\begin{split}
& \ess{ \left[\supf \left(\jsp - \js\right)\right] } \\
&~ = \nc \ess{\expe_{\sigma} \left[\supf \sumauc T^{i,j,m,n}\right]}.
\end{split}
\end{equation}
Given $\mathcal{S}= \left\{(\bm{x}_i,y_i)\right\}_{i=1}^m$,  $\mathcal{S}'= \left\{(\bm{x}'_i,y_i)\right\}_{i=1}^m$, where we sort the instances such that $y_i \le y_j$ for $i \le j$ for the sake of convenience, since the instances are drawn independently, we employ the fact that: 
\begin{equation}
\ess{\left[ \supf \left(\jsp - \js\right) \right] }  =  \ess{\left[  \supf \left(\jspt - \jst\right) \right] }
\end{equation}
where $\ts$ and $\tsp$ are permuted datasets obtained from $\mathcal{S}$ and $\mathcal{S}'$ respectively after exchanging $ 0 \le x \le N$ samples of $(\bm{x}_i,y_i)$ and $(\bm{x}'_i,y'_i)$ sharing the same index.

\noindent Next, we show that
for any sequence of \textit{i.i.d} Rademacher random variables,  $\sigma = (\sigma^{(1)}_1, \cdots,\sigma^{(1)}_{n_1},\cdots,\sigma^{(2)}_{1} ,\cdots, \sigma^{(N_C)}_{n_C})$ and any $\mathcal{S}$, and $\mathcal{S}'$, there exists a pair of permuted datasets $\tilde{S}^\sigma$ and  $\tilde{S}'^\sigma$ such that  
\begin{equation}
\supf \left[\sumauc T^{i,j,m,n}\right] 
 = \supf \left[ \jfunp{\sigma} - \jfun{\sigma}. \right] 
\end{equation} 
Specifically, we show this by induction.

\noindent \textbf{Base Case}. We assume that $\mathcal{S} = \left\{ (\bm{x}_1,1), (\bm{x}_2,2)\right\}$ $\mathcal{S}' = \left\{(\bm{x}'_1,1), (\bm{x}'_2,2)\right\}$, $\sigma = (\sigma^{(1)}_1,\sigma^{(2)}_1)$. Here we define 
\[T^1 = T^{1,2,1,1}, T^2 = T^{2,1,1,1} .\] We have the following cases:
\begin{itemize}
\item[(a)] $\sigma^{(1)}_1 =1,\sigma^{(2)}_1 = 1$. We have:
\begin{equation}
\begin{split}
& T^1 = \ell(f^{(1)},\bm{\xp}_1,\bm{\xp}_2) - \ell(f^{(1)},\bm{\x}_1,\bm{\x}_2),\\
& T^2 = \ell(f^{(2)},\bm{\xp}_2,\bm{\xp}_1) - \ell(f^{(2)},\bm{\x}_2,\bm{\x}_1),\\
\end{split}
\end{equation}
This shows that 
\begin{equation}
\supf \left[T^1 +T^2\right] = \supf \left[\jsp - \js\right].
\end{equation}
This suggests that $\tilde{\mathcal{S}}_{\sigma} = \mathcal{S}$, $\tilde{\mathcal{S}'}_{\sigma} = \mathcal{S}'$.

\item[(b)] $\sigma^{(1)}_1 =1,\sigma^{(2)}_1 =-1$. \begin{equation}
\begin{split}
& T^1 = \ell(f^{(1)},\bm{\xp}_1,\bm{\x}_2) - \ell(f^{(1)},\bm{\x}_1,\bm{\xp}_2),\\
& T^2 = \ell(f^{(2)},\bm{\x}_2,\bm{\xp}_1) - \ell(f^{(2)},\bm{\xp}_2,\bm{\x}_1),\\
\end{split}
\end{equation}
This shows that 
\begin{equation}
\supf \left[T^1 +T^2\right] = \supf \left[\jfunp{\sigma} - \jfun{\sigma}\right].
\end{equation}
where $\tilde{\mathcal{S}}_{\sigma} and \tilde{\mathcal{S}'}_{\sigma}$ are obtained by exchanging $(\bm{x}_2,2)$ and  $(\bm{x}'_2,2)$

\item[(c)] $\sigma^{(1)}_1 =-1,\sigma^{(2)}_1 =1$. One can show that the corresponding $\tilde{\mathcal{S}}_{\sigma},\tilde{\mathcal{S}'}_{\sigma}$  are obtained by exchanging $(\bm{x}_1,1)$ and  $(\bm{x}'_1,1)$.
\item[(d)] $\sigma^{(1)}_1 =-1,\sigma^{(2)}_1 =-1$. One can show that the corresponding $\tilde{\mathcal{S}}_{\sigma},\tilde{\mathcal{S}'}_{\sigma}$  are obtained by completely exchanging $\mathcal{S}$ and  $\mathcal{S}'$.
\end{itemize}
The arguments above complete the proof for base case.

\noindent \textbf{Recursion}. Given \[\mathcal{S}^- =\{(\bm{x}_i,y_i)\}_{i=1}^{k},\mathcal{S}^{-'} =\{(\bm{x}'_i,y_i)\}_{i=1}^{k}, \forall \sigma^- \in \{-1,1\}^k,\]
we assume that $\mathcal{S}^-,\mathcal{S}^{-'}$ and $\sigma^-$ satisfy our conclusion, which is realized by $\tilde{\mathcal{S}}^-_{\sigma^-}$ and $\tilde{\mathcal{S}}^{-'}_{\sigma^-}$. We now show that given a new pair $(\x_{n}, i)$ , $(\x'_{n}, i)$ and $\sigma^{(i)}_{n} \in \{-1,1\}$ such that
\[\mathcal{S} = \mathcal{S}^-\cup\{(\x_{n}, i)\}, \mathcal{S}' = \mathcal{S}^{-'} \cup \{(\x'_{n}, i)\}, \sigma = (\sigma^-,\sigma^{(i)}_{new}) \]
still satisfies our conclusion. Obviously only $T^{i,*,n,*}$ and $T^{*,i,*,n}$ are involved with the new sample, where $*$ simply refers to all the possible choices.
 
\noindent For the case that $\sigma^{(i)}_{n}=-1$, we have the following cases:
First, we examine the terms $T^{i,*,n,*}$ by taking any specific $T^{i,j,n,m}$.
\begin{itemize}
\item[(a)] $\sigma^{(j)}_m = 1$, we have:
\[T^{i,j,n,m} = \ell(f^{(i)},\bm{\x}_n,\bm{\xp}_m) - \ell(f^{(i)},\bm{\xp}_n,\bm{\x}_m).\]
 \item [(b)] $\sigma^{(j)}_m = -1$, we have:
\[T^{i,j,n,m} = \ell(f^{(i)},\bm{\x}_n,\bm{\x}_m) - \ell(f^{(i)},\bm{\xp}_m,\bm{\xp}_n).\]
 \end{itemize}
 Moreover, for any  $T^{j,i,n,m}$, we have:
\begin{itemize}
\item[(a)] $\sigma^{(j)}_m = 1$, we have:
\[T^{j,i,m,n} = \ell(f^{(j)},\bm{\xp}_m,\bm{\x}_n) - \ell(f^{(j)},\bm{\x}_m,\bm{\xp}_n).\]
 \item [(b)] $\sigma^{(j)}_m = -1$, we have:
\[T^{i,j,n,m} = \ell(f^{(j)},\bm{\x}_m,\bm{\x}_n) - \ell(f^{(j)},\bm{\xp}_m,\bm{\xp}_n).\]
 \end{itemize}
 This shows that 
 \[\tilde{\mathcal{S}}_{\sigma} = \tilde{\mathcal{S}}^-_{\sigma^-} \cup \{(\x'_{n}, i)\}, ~ \tilde{\mathcal{S}}'_{\sigma} = \tilde{\mathcal{S}}^{-'}_{\sigma^-} \cup \{(\x_{n}, i)\}. \]
 Similarly if $\sigma^{(i)}_n = 1$, we have 
  \[\tilde{\mathcal{S}}_{\sigma} = \tilde{\mathcal{S}}^-_{\sigma^-} \cup \{(\x_{n}, i)\}, ~ \tilde{\mathcal{S}}'_{\sigma} = \tilde{\mathcal{S}}^{-'}_{\sigma^-} \cup \{(\x'_{n}, i)\}. \]
  In either case, we know that $\tilde{\mathcal{S}}_{\sigma}$ and $\tilde{\mathcal{S}}'_{\sigma}$ are obtained by exchanging the corresponding instances of $\mathcal{S}$ and $\mathcal{S}'$.\\
  By the arguments in the base case and the recursion, we complete the proof with:  
  \begin{equation}
\supf \left[\sumauc T^{i,j,m,n}\right] 
 = \supf \left[ \jfunp{\sigma} - \jfun{\sigma} \right]. 
\end{equation} 
It immediately suggests that 
  \begin{equation}
  \begin{split}
 &\ess{\expe_{\sigma} \bigg[\supf \sumauc T^{i,j,m,n}\bigg] } \\
 &~~ = \expe_{\sigma}  \ess{\left[ \supf  \left(\jfunp{\sigma} - \jfun{\sigma}\right) \right] }\\
 &~~ = \frac{1}{2^N} \cdot  \sum_{\sigma} \ess{\left[ \supf \left(\jfunp{\sigma} - \jfun{\sigma}\right) \right]}\\
  &~~ = \frac{1}{2^N} \cdot  \sum_{\sigma} \ess{\left[\supf  \left(\jsp - \js\right) \right]}\\
 &~~ = \frac{2^N}{2^N} \cdot   \ess{\left[\supf  \left(\jsp - \js\right) \right]}\\
 &~~ = \ess{ \left[ \supf \left(\jsp - \js\right) \right]}.
  \end{split}
\end{equation} 
The proof then follows that
\begin{equation}
\ess{\expe_{\sigma} \left[\supf \sumauc T^{i,j,m,n}\right] } \le 4\sR.
\end{equation}
\end{pf}

{\color{white}ddsadsa} \\

\subsection{A General Result}
In this section, we will provide a proof for the general result for $\mauc$ generalization bound, which is based on Lem.\ref{lem:mc}, and Lem.\ref{lem:perm}.
\begin{rthm4} 
Given dataset $\mathcal{S} = \{(\bm{x}_i,y_i)\}_{i=1}^m$, where the instances are sampled independently, for all multiclass scoring functions $f \in \mathcal{H}$, if $dom\ell = [0,B]$, $\forall \delta \in (0,1)$, the following inequalities hold with probability at least $1 - \delta$:
\begin{equation*}
R_\ell(f) \le \js + \frac{4\empsR}{N_C(N_C-1)}  +  \frac{10B }{N_C}  \cdot \xi(\bm{Y}) \cdot \sqrt{\frac{\log(\nicefrac{2}{\delta})}{2N}}
\end{equation*}
where $\xi(\bm{Y}) = \sqrt{\sum_{i=1}^{n_C}\nicefrac{1}{\rho_i}}.$
\end{rthm4}

\begin{pf}
Given $\mathcal{S}'$ as another dataset being independent with $\mathcal{S}$, since $\js = \mathbb{E}_{\mathcal{S}'}\jsp$, using the Jensen's inequality, we have:
\begin{equation*}
\begin{split}
 \es{\left[\supf \left(\es{\js} - \js \right) \right]} &= \es {\supf \expe_{\mathcal{S}'}{ \left[\js-\jsp\right]  }}  \\
&\le \ \ \ \  \ess{ \left[\supf \left(\js-\jsp \right) \right]  }
\end{split}
\end{equation*}
From Lem.\ref{lem:perm}, we have:
\begin{equation}\label{eq:rade}
\begin{split}
\ess{\left[ \supf \left(\js - \jsp\right)\right] } \le 4\nc\sR.
\end{split}
\end{equation}

Given $\mathcal{S}$,  define $\mathcal{S}_m = ( \mathcal{S} ~ \backslash ~ \{(\bm{x}_m, y_m)\}) \cup {\{(\bm{x}'_m, y_m)\}}$.
We now proceed to derive the bounded difference property for the loss function. Suppose that $y_m = i$, we have:

\begin{equation*}
\begin{split}
d_i &=  |\supf(\es{\js} - \js ) - \supf(\mathbb{E}_{\mathcal{S}_i}{\jsi} - \jsi )|\\ 
&\le \supf |\js - \jsi|\\
& = \nicefrac{1}{N_C(N_C-1)} \supf \bigg[\suminj \sumxtwo \ninj |\ell(f^{(i)},\x_m,\x_n) - \ell(f^{(i)},\x'_m,\x_n)  | \\
&~~~~+ \suminj \sum_{\x_n \in \Nj} \ninj |\ell(f^{(j)},\x_n,\x_m) - \ell(f^{(j)},\x_n,\x'_m)  |   \bigg]\\
& \le \ \ \  \ \frac{2B}{N_C\ni}
\end{split}
\end{equation*}

Thus we have $\sum_{i=1}^N d_i^2 = 4 \nicefrac{B^2}{N_C^2} \cdot \sum_{i=1}^{n_C}\nicefrac{1}{n_i} $.
According to Eq.(\ref{eq:rade}) and Lem.\ref{lem:mc}, we know that fixing the labels $\bm{Y}$, with probability at least $1 - \nicefrac{\delta}{2}$, the following inequality  holds:

\begin{equation}\label{eq:half}
\es{\js} \le \js + \frac{4}{N_C(N_C-1)} \sR +  \frac{2 B }{N_C}\xi(\bm{Y})  \cdot \sqrt{\frac{\log(\nicefrac{2}{\delta})}{2N}}
\end{equation}

 Using an analogous argument, we can prove that, fixing the labels $\bm{Y}$, with probability at least $1 - \nicefrac{\delta}{2}$, the following inequality  holds:
\begin{equation}\label{eq:half2}
\frac{\sR}{N_C(N_C-1)} \le \frac{\empsR}{N_C(N_C-1)} + \frac{2 B }{N_C} \xi(\bm{Y}) \cdot \sqrt{\frac{\log(\nicefrac{2}{\delta})}{2N}}.
\end{equation}
Combining Eq.(\ref{eq:half}) and Eq.(\ref{eq:half2}), fixing  $\bm{Y}$, we have with probability at least $1 - \delta$ that:
\begin{equation}\label{eq:gen_xy}
\es{\js} \le \js + \frac{4}{N_C(N_C-1)} \empsR +  \frac{10B }{N_C} \xi(\bm{Y}) \cdot \sqrt{\frac{\log(\nicefrac{2}{\delta})}{2N}}
\end{equation}

Following the arguments in Thm.8 in \cite{genal}, we have Eq.\eqref{eq:gen_xy} holds with probability (over $\bm{X}$ and $\bm{Y}$) at least $1 - \delta$.
\end{pf}
\subsection{Sub-Gaussian Property of the $\mauc$ Rademacher Complexity}
\begin{defi}[Sub-Gaussian Stochastic Process] A stochastic process $\theta \mapsto X_\theta$ with indexing set $\mathcal{T}$ is sub-Gaussian to a pseudo-metric $d$ on $\mathcal{T}$, if for all $\theta, \theta' \in \mathcal{T}$ and all $\lambda \in \mathbb{R}$, 
\begin{equation}
\expe\left[\exp\left(\lambda \cdot (X_\theta - X_{\theta'})\right)\right] \le \exp\left(\frac{\lambda^2 d(\theta, \theta')^2}{2}\right).
\end{equation}

\end{defi}

\noindent Define $\tfs = \empcompsim$, and recall that \[T^{i,j,m,n} = \sigmaterm,\] with the function $f$ chosen from an index set (in this case a function class) $\mathcal{F}$. We see that ${\tfs}_{f \in \mathcal{F}}$ is essentially a stochastic process of the Rademacher random variables $\sigma$, which is revealed by the following lemma:

\begin{lem}\label{lem:pro} Given an input feature set $\mathcal{D}_{\mathcal{X}} = \{\bm{x}_1, \cdots, \bm{x}_N\}$, $\mathcal{Z}_{\mathcal{D}}$ is defined as 
\[ \big\{(\bm{x}, y): \bm{x} \in \mathcal{D}_{\mathcal{X} }, y \in \{1,2,\cdots, N_C\} \big\},\]
 which is the cartesian product between $\mathcal{D}_{\mathcal{X}}$ and the label space.  Moreover, $\forall \bm{z}= (\bm{x}, i) \in \mathcal{Z}_{\mathcal{D}}$, $s(\bm{z}) = s^{(i)}(\bm{x})$.
With the notations above, if $\ell$ is $\phil$-Lipschitz continuous, we have that,  $\{C_G\cdot\tfs\}_{f \in \mathcal{H}}$ with $f = (f^{(1)},\cdots, f^{N_C})$ and $f^{(i)} = \mathsf{soft}^{(i)}\circ \s$ is a sub-Gaussian stochastic process in the sense that :
\begin{equation}
\expe_{\sigma}\Bigg[\exp\bigg(\lambda \cdot C_G \cdot \big(\tfsd\big)\bigg)\Bigg] \le \exp\left(\frac{\lambda^2 d_\infty(\cone{\s}, \ctwo{\st})^2}{2}\right).
\end{equation}
where
\begin{equation}
	C_{G} = \dfrac{1} {\phil \cdot (N_C-1) \cdot \xi(\bm{Y}) \cdot \sqrt{\frac{1}{N}}},~~~d_{\infty, \mathcal{S}}(\s, \st) = \max_{\bm{z} \in \mathcal{Z}_{\mathcal{D}}} |\s(\bm{z}) - \st({\bm{z}}
	)|.
\end{equation}

\end{lem}
\begin{pf}
Denote $\sigone$ as a set of Rademacher random variables in the following form:

 \[{\sigone}=(\sigma^{(1)}_1, \cdots, \sigma^{(1)}_{n_1}, \cdots ,\sigma^{(i)}_{1}, \cdots, \cone{\sigma^{(i)}_{k}}, \cdots, \sigma^{(i)}_{n_1}, \cdots, \sigma^{(N_C)}_{n_{N_C}})\]
 and denote $\sigtwo$ as another set of Rademacher random variables with all entries of which  the same as $\sigma$ except that $\color{red}\sigma^{(i)}_k$ is replaced with $\color{c2}\tilde{\sigma}^{(i)}_k$, i.e,

 \[ \sigtwo=(\sigma^{(1)}_1, \cdots, \sigma^{(1)}_{n_1}, \cdots ,\sigma^{(i)}_{1}, \cdots, {\color{c2}\tilde{\sigma}^{(i)}_{k}}, \cdots, \sigma^{(i)}_{n_1}, \cdots, \sigma^{(N_C)}_{n_{N_C}}).\] We have the following bounded difference property for every possible choice of $(i,k)$:

\begin{equation*}
\begin{split}
diff_i &=   \left| \left(\tfsd\right) - \left( \tfsdi \right) \right|\\
& =  \bigg|\left(\frac{{\color{blue}{\sigma^{(i)}_k}}-{\color{red}\tilde{\sigma}^{(i)}_k}}{2}\right) \cdot \bigg[\suminj \sumxtwo  \cdot \ninj \left(\ell(\cone{f}^{(i)},\x_m,\x_n) - \ell(\ctwo{\funt}^{(i)},\x_m,\x_n) \right) \\
&~~~~+ \suminj \sum_{\x_n \in \Nj}  \ninj \left(\ell(\cone{f}^{(j)},\x_n,\x_m) - \ell(\ctwo{\funt}^{(j)},\x_n,\x_m) \right)\bigg]\bigg|\\
& \le \ \ \  \ \frac{2\cdot (N_C-1)}{\ni} \max_{i,~\x_m \in \Ni,~ \x_n \notin \Ni} \left|\ell(\cone{f}^{(i)},\x_m,\x_n) - \ell(\ctwo{\funt}^{(i)},\x_m,\x_n)\right|\\
&\le  \ \ \  \  \frac{4\cdot (N_C-1)}{\ni} \cdot \phil \cdot \max_{i,~\x \in \mathcal{D}_{\mathcal{X}} } \left|\cone{f}^{(i)}(\x) - \ctwo{\funt}^{(i)}(\x)\right|\\
&= \ \ \  \ \frac{4\cdot (N_C-1)}{\ni} \cdot \phil \cdot \max_{i,~\x \in \mathcal{D}_{\mathcal{X}} } \left|\int_0^1 \left<\nabla\mathsf{soft}^{(i)}\left(\tau \cdot \bm{\ctwo{\st}}(\x) + (1-\tau) \cdot \cone{\s}(\x)\right), ~\left(\cone{\s}(\x) - \ctwo{\st}(\x)\right)\right> d\tau\right|\\
&\le\ \ \ \  \frac{4\cdot (N_C-1)}{\ni} \cdot \phil \cdot \left(\sup_{\x}||\nabla\mathsf{soft}^{(i)}(\x)||_1\right) \cdot  \max_{\bm{z} \in \mathcal{Z}_{\mathcal{D}} }|\cone{\s}(\bm{z}) - \ctwo{\st}(\bm{z})| \\
& \le\ \ \ \ \frac{2\cdot (N_C-1)}{\ni} \cdot \phil \cdot \max_{\bm{z} \in \mathcal{Z}_{\mathcal{D}} }  |\cone{\s}(\bm{z}) - \ctwo{\st}(\bm{z})|.
\end{split}
\end{equation*}
It is now easy to see that $C_G \cdot \big(\tfsd\big)$ satisfies the bounded difference property (Def.\ref{def:bdp}). According to Lem.\ref{lem:bdp}, we could choose $v$ therein as:
\begin{equation*}
v = \frac{1}{4} \sum_{i=1}^N C_G^2 \cdot diff_i^2 = \max_{\bm{z} \in \mathcal{Z}_{\mathcal{D}} } |\bm{\cone{s}}(\bm{z}) - \ctwo{\st}(\bm{z})|^2
\end{equation*} 
This suffices to complete the proof based on Lem.\ref{lem:bdp}.
\qed

\end{pf}

\subsection{Chaining Bounds}
Now with the sub-Gaussian property proved in Lem.\ref{lem:pro}, we can derive chaining upper bounds  for the $\mauc$ Rademacher complexity based on the notion of covering number. Before presenting the results, let us start with the definition of $\epsilon$-covering and the covering number.
\begin{rdef3}[$\epsilon$-covering] \cite{tala} Let $(\mathcal{H}, d)$ be a (pseudo)metric space, and $\Theta \in \mathcal{H}$. $\{h_1,\cdots, h_K\}$ is said to be an  $\epsilon$-covering of $\Theta$ if $\Theta \in \bigcup_{i=1}^K\mathcal{B}(h_i,\epsilon)$, \emph{i.e.}, $\forall \theta \in \Theta$, $\exists i ~s.t.~ d(\theta, h_i) \le \epsilon$.
\end{rdef3}

\begin{rdef4}[Covering Number] \cite{tala} Based on the notations in Def.\ref{def:ecover}, the covering number of $\Theta$ with radius $\epsilon$ is defined as:
  \begin{equation*}
  \mathfrak{C}(\epsilon,\Theta, d) = \min\left\{n: \exists \epsilon-\text{covering over $\Theta$ with size n}\right\}.
  \end{equation*}
\end{rdef4}

\begin{rthm6}[Chaining Bounds for $\mauc$ Rademacher Complexity] Suppose that the score function $s^{(i)}$ maps $\mathcal{X}$ onto a bounded interval $[-R_s, R_s]$, the following properties hold for $\empsR$:
\begin{enumerate}
\item[(a)] For a decreasing precision sequence $\{\epsilon_k\}_{k=1}^K$, with $\epsilon_{k+1} = \frac{1}{2} \epsilon_{k},~ k=1,2,\cdots, K-1$ and $\epsilon_0 \ge R_s$, we have:
\begin{equation*}
  \begin{split}
  \empsR \le N_C \cdot (N_C -1) \cdot \phil\cdot \epsilon_K
  + 6\cdot \sum_{k=1}^K \epsilon_k \phil(N_C-1) \cdot \xi(\bm{Y})\sqrt{\frac{\log(\mathfrak{C}(\epsilon_k,\mathcal{F}, d_{\infty, \mathcal{S}}))}{N}} 
  \end{split}
  \end{equation*}
\item[(b)] There exists a universal constant $C$, such that:
\[\empsR \le C\cdot \inf_{ R_s \ge \alpha \ge 0} \left(N_C \cdot (N_C -1)\cdot \phil \cdot \alpha + \phil \cdot (N_C-1) \cdot \xi(\bm{Y}) \cdot \int_{\alpha}^{R_s} \sqrt{\frac{\log(\mathfrak{C}(\epsilon,\mathcal{F}, d_{\infty, \mathcal{S}}))}{N}} d\epsilon  \right)\]
\end{enumerate}
\end{rthm6}

\begin{pf}
First of all, we have:
\begin{equation}
\begin{split}
2C_G \empsR  =& C_G \expe_{\sigma}\left[\sup_{\conef \in \mathcal{H}} T_{\conef}(\sigma) + \sup_{\ctwo{\funt} \in \mathcal{H}} T_{\ctwo{\funt} }(\sigma) \right]\\
 &=C_G \expe_{\sigma}\left[\sup_{\conef \in \mathcal{H}} T_{f}(\sigma) +\sup_{\ctwo{\funt}  \in \mathcal{H}} T_{\ctwo{\funt} }(-\sigma) \right]\\ 
& = C_G \expe_{\sigma}\left[\sup_{\conef \in \mathcal{H}} T_{\conef}(\sigma) +\sup_{\ctwo{\funt}  \in \mathcal{H}} -T_{\ctwo{\funt} }(\sigma) \right]\\
 & = C_G\expe_{\sigma}\left[\sup_{\conef,\ctwo{\funt}  \in \mathcal{H}} \left(T_{\conef}(\sigma) - T_{\ctwo{\funt}}(\sigma) \right)\right] 
\end{split}
\end{equation}
Now define $\hat{\mathcal{H}}$ as an $\epsilon$-cover of $\mathcal{H}$ with the pseudo-metric $d_{\infty,\mathcal{S}}$. Choose ${\conehf}, {{\ctwohft}} \in \hat{\mathcal{H}}$, such that $d_{\infty,\mathcal{S}}(\conef, {\conehf})\le \epsilon$ and  $d_{\infty,\mathcal{S}}(\ctwo{\funt}, {\ctwohft})\le \epsilon$. We then have the following result:
\begin{equation}
\begin{split}
&C_G\expe_{\sigma}\left[\sup_{\conef,\ctwoft \in \mathcal{H}} \left(T_{\conef}(\sigma) - T_{\ctwoft}(\sigma) \right)\right] \\
&=C_G\expe_{\sigma}\left[\sup_{\conef,\ctwoft \in \mathcal{H}} \left(T_{\conef}(\sigma) - T_{\conehf}(\sigma)\right) + \left(T_{\conehf}(\sigma) -  T_{\ctwohft}(\sigma)\right) +  \left(T_{\ctwohft}(\sigma) -  T_{\ctwoft}(\sigma) \right)\right]\\
&\le 2C_G\expe_{\sigma}\left[\sup_{d_{\infty,\mathcal{N}}(\cones,\conehs)\le \epsilon} \left(T_{\conef}(\sigma) - T_{\conehf}(\sigma)\right)\right] +  C_G\expe_{\sigma}\left[\sup_{\conehs,\ctwohst \in \hat{\mathcal{H}}_{\epsilon}} \left(T_{\conehf}(\sigma) - T_{\ctwohft}(\sigma)\right)\right]
\end{split}
\end{equation} 
Furthermore, let $\hat{\mathcal{H}}_k$ be an $\epsilon_k$-cover of $\mathcal{H}$. For each $k$, pick $\conehsk$, $\ctwohstk$ such that $d_{\infty, \mathcal{S}}(\conehsk, \cones) \le \epsilon_k$ and $d_{\infty, \mathcal{S}}(\ctwohstk, \ctwost) \le \epsilon_k$. Specifically, we take  $\epsilon_K = \epsilon$, $\epsilon_0 \ge R_s$,  which allows us to choose $\conehsn{0} = \ctwohstn{0}$,  $\epsilon_{k+1} = \frac{1}{2} \epsilon_k$, $\conehfn{K} = \conehf$, and $\ctwohftn{K} = \ctwohft$  . With $\conehfk = \mathsf{soft} \circ \conehsk$ and $\ctwohftk = \mathsf{soft} \circ \ctwohstk$. Then, for $\cones, \ctwost \in \hat{\mathcal{H}}_{\epsilon}$, we can decompose  $T_{\conehf}(\sigma)$  as 
\[T_{\conehf}(\sigma) = T_{\conehfn{K}}(\sigma)= T_{\conehfn{0}}(\sigma) + \sum_{i=1}^K \left(T_{\conehfk}(\sigma) - T_{\conehfn{k-1}}(\sigma)\right)~~~ \text{and}~~~T_{\ctwohft}(\sigma) = T_{\ctwohftn{K}}(\sigma)= T_{\ctwohftn{0}}(\sigma) + \sum_{i=1}^K \left(T_{\ctwohftk}(\sigma) - T_{\ctwohftn{k-1}}(\sigma)\right). \]
Based on the decompositions, we have:
\begin{equation}\label{eq:chain1}
\begin{split}
C_G\expe_{\sigma}\left[\sup_{\conehs,\ctwohst \in \hat{\mathcal{H}}_{\epsilon}} \left(T_{\conehf}(\sigma) - T_{\ctwohft}(\sigma)\right)\right] \le  2 \cdot C_G \cdot \sum_{i=1}^K \expe_{\sigma}\left[\supcover \left(T_{\hat{f}_k}(\sigma) - T_{\hat{f}_{k-1}}(\sigma)\right)\right].
\end{split}
\end{equation}
According to maximal inequality (Lem.\ref{lem:max}), we have:
\begin{equation}\label{eq:chain2}
C_G \expe_{\sigma}\left[\supcover \left(T_{\hat{f}_k}(\sigma) - T_{\hat{f}_{k-1}}(\sigma)\right)\right] \le  3\epsilon_{k} \sqrt{2\log{|\hat{\mathcal{H}}_{k}|\cdot |\hat{\mathcal{H}}_{k-1}|}} \le 6\epsilon_{k}\sqrt{\log(\mathfrak{C}(\epsilon_k,\mathcal{F}, d_{\infty, \mathcal{S}}))}.
\end{equation}

Following a similar derivation to Lem.\ref{lem:pro}, we have:
\begin{equation}\label{eq:chain3}
\expe_{\sigma}\left[\sup_{d_{\infty,\mathcal{N}}(\cones,\conehs)\le \epsilon_K} \left(T_{\conef}(\sigma) - T_{\conehf}(\sigma)\right)\right] \le N_C \cdot (N_C -1) \cdot \phil \cdot \epsilon_K
\end{equation}
Putting all together, we have:
\begin{equation*}
\begin{split}
\empsR &= \frac{1}{2} \cdot \expe_{\sigma}\left[\sup_{\conef,\ctwoft \in \mathcal{H}} \left(T_{\conef}(\sigma) - T_{\ctwoft}(\sigma) \right)\right]  \le N_C \cdot (N_C -1) \cdot \phil \cdot  \epsilon_K + \left(\frac{1}{C_G}\right) 6 \sum_{i=1} \epsilon_{k} \sqrt{\log(\mathfrak{C}(\epsilon_k,\mathcal{F}, d_{\infty, \mathcal{S}}))} \\
& \le  N_C \cdot (N_C -1) \cdot \phil\cdot \epsilon_K +  6 \sum_{i=1}^K \epsilon_{k}\phil\cdot (N_C-1) \xi(\bm{Y})\sqrt{\frac{\log(\mathfrak{C}(\epsilon_k,\mathcal{F}, d_{\infty, \mathcal{S}}))}{N}}
\end{split}
\end{equation*}
This ends the proof of (a).\\
Given the result of (a), we now turn to proof (b). First note that $\epsilon_k  = 2(\epsilon_{k} - \epsilon_{k+1})$ and that $\log(\mathfrak{C}(\epsilon,\mathcal{F}, d_{\infty, \mathcal{S}})$ is non-increasing with respect to $\epsilon$.
We then have:
\begin{equation*}
\begin{split}
 &N_C \cdot (N_C -1) \cdot \phil\cdot \epsilon_K +  6 \sum_{i=1}^K \epsilon_{k} \phil\cdot (N_C-1) \cdot \xi(\bm{Y})\sqrt{\frac{\log(\mathfrak{C}(\epsilon_k,\mathcal{F}, d_{\infty, \mathcal{S}}))}{N}} \\
& \le 2 N_C \cdot (N_C -1) \cdot  \phil\cdot \epsilon_{K+1} +  12 \sum_{i=1}^K (\epsilon_{k} - \epsilon_{k+1}) \cdot (N_C-1)\phil\cdot \xi(\bm{Y})\sqrt{\frac{\log(\mathfrak{C}(\epsilon_k,\mathcal{F}, d_{\infty, \mathcal{S}}))}{N}}
\end{split}
\end{equation*}
Furthermore, we have:
\begin{equation*}
\begin{split}
\empsR &\le 12\phil \left( N_C \cdot (N_C -1) \cdot \epsilon_{K+1} +  \sum_{i=1}^K (\epsilon_{k} - \epsilon_{k+1})  \cdot (N_C-1) \cdot \xi(\bm{Y})\sqrt{\frac{\log(\mathfrak{C}(\epsilon_k,\mathcal{F}, d_{\infty, \mathcal{S}}))}{N}}\right)\\
&\le 12 \phil \bigg( N_C \cdot (N_C -1) \cdot \epsilon_{K+1} + \int_{\epsilon_{K+1}}^{R_s}   (N_C-1) \xi(\bm{Y})\sqrt{\frac{\log(\mathfrak{C}(\epsilon,\mathcal{F}, d_{\infty, \mathcal{S}}))}{N}} d\epsilon \\
&~~~+\int_{R_s}^{\epsilon_0}  (N_C-1) \xi(\bm{Y})\sqrt{\frac{\log(\mathfrak{C}(\epsilon,\mathcal{F}, d_{\infty, \mathcal{S}}))}{N}} d\epsilon  \bigg) \\
&= 12 \phil \left(N_C \cdot (N_C -1) \cdot \epsilon_{K+1} + \int_{\epsilon_{K+1}}^{R_s}  \xi(\bm{Y})\sqrt{\frac{\log(\mathfrak{C}(\epsilon,\mathcal{F}, d_{\infty, \mathcal{S}}))}{N}} d\epsilon \right)
\end{split}
\end{equation*}

It then becomes clear that the universal constant should be chosen as $C = 12$. Set $\alpha = \epsilon_{K+1}$.  By proper choices of $K$ and $\epsilon_0$ and let $\alpha = \frac{\epsilon_0}{2^{K+1}}$, we see that the inequality holds for all $\alpha \in [0, R_s]$, which thus completes the proof. 
\qed
\end{pf}
According to Thm.\ref{thm:chain}, we have the following theorem as a result of a new minorization technique which appears in a recent work \cite{optbound}. 
\begin{rthm7}[Transformation Upper Bound] Given the Hypothesis class \[ \mathsf{soft} \circ \mathcal{F} = \bigg\{\bm{g}(\x) = \mathsf{soft}(\bm{s}(\x)):~ \s \in \mathcal{F}\bigg\}, \]  where $\mathsf{soft}(\cdot)$ is the softmax function. Suppose that $\s(x) \subseteq [-R_s, R_s]^{N_C}$ and $\ell$ is $\phil$-Lipschitz continuous, the following inequality holds:
  \begin{equation}
  \begin{split}
  &\frac{\empsRF}{N_C(N_C-1)} \le \phil\bigg( 2^9 \cdot \frac{1}{N_C} \cdot  \sqrt{N_C}\cdot\xi(\bm{Y})\cdot \log^{3/2}\left(e \cdot R_s \cdot N \cdot N_C\right) \cdot \hat{\mathfrak{R}}_{N\cdot N_C}(\Pi \circ \mathcal{F})+ \sqrt{\frac{1}{N}}\bigg), 
  \end{split}
  \end{equation}
  where the Rademacher complexity  $\hat{\mathfrak{R}}_{N\cdot N_C}(\Pi \circ \mathcal{F})$ is defined as:
  \begin{equation}
  \begin{split}
  &\hat{\mathfrak{R}}_{N\cdot N_C}(\Pi \circ \mathcal{F})  \\
  &~~= \expe_{\sigma}\left[\sup_{f = (f^{(1)}, \cdots, f^{(N_C)})\in \mathcal{F} } \frac{1}{N\cdot N_C} \sum_{j=1}^{N_C} \sum_{i=1}^N \sigma^{(i)}_j \cdot f^{(j)}(\x_i)\right],
  \end{split}
  \end{equation}
  where $\{\sigma^{(i)}_j\}_{(i,j)}$ is a sequence of independent Rademacher random variables.
\end{rthm7}
\begin{pf}
With the help of the chaining bound in Thm.\ref{thm:chain}{\color{blue}{-a)}}, the proof follows an analogous spirit as  \cite[\color{org}{Prop.1}]{optbound}, with the parameters therein set as $\lambda = 1, ~\theta = 0, ~q = N_C, ~n = N.$
\end{pf}

 The result in Thm.\ref{thm:trans} directly relates the complicated pairwise Rademacher complexity $\empsRF$ to the instance-wise Rademacher complexity $\hat{\mathfrak{R}}_{N\cdot N_C}$. This makes the derivation of the generalization upper bound much easier.  Once a bound on the ordinary Rademacher complexity over the functional class $\mathcal{F}$ is available, we can directly plug it in to this theorem and find a resulting bound over the $\mauc$ complexity.

\section{Practical Generalization Bounds}\label{sec:appch}

\subsection{Practical Results for Linear Models}\label{sec:appd6}
In this section, we provide concrete generalization bounds for two practical models.
\begin{rthm5}
  Based on the assumptions of Thm.\ref{thm:gen}, we have the   Given the hypothesis space for $\ell_p$ norm penalized linear model as:
  \begin{equation*}
  \begin{split}
  \mathcal{H}^{Lin}_{p,\gamma} = \big\{&f=(f^{(1)},\cdots,f^{(n_C)}): f^{(i)}(\x) = \bm{W}^{(i)}\bm{x},\\
   &||\bm{W}^{(i)}||_p \le \gamma \big\},
  \end{split}
  \end{equation*}
  if $\ell$ is $\phil$-Lipschitz continuous, and the input features are sampled from $\mathcal{X} \subset \mathcal{R}^{d}$, and for all $\bm{x} \in \mathcal{X}$, we have $||\bm{x}||^2_2 \le R$ with $0<p<\infty$, $\frac{1}{p} + \frac{1}{\bar{p}} =1$, for all $f\in \Hlin$, we have the following inequality holds with probability at least $1-\delta$:
  \begin{equation*}
  {R}_{\ell}(f) \le \hat{R}_{\ell, \mathcal{S}}(f)  +  \mathcal{I}_{{Lin}}\bigg(\chi(\bm{Y}),\xi({\bm{Y}}),\delta\bigg) \cdot \sqrt{\frac{1}{N}},
  \end{equation*}
  where 
  \begin{equation*}
  \begin{split}
  & \mathcal{I}_{{Lin}}\bigg(\chi(\bm{Y}),\xi({\bm{Y}}),\delta\bigg) =  \frac{4 \Rx\phil\gamma}{N_C-1} \cdot  \sqrt{\frac{2(\bar{p} -1)}{N_C}} \cdot \rchi(\bm{Y})  \\
  &+  \dfrac{5B }{N_C } \cdot \sqrt{2\log\left(\dfrac{2}{\delta}\right)} \cdot \xi(\bm{Y}).
  \end{split}
  \end{equation*}
\end{rthm5}

\begin{pf}
Follows Thm.\ref{thm:gen}, Lem.\ref{lem:rlin} 
\end{pf}

\subsection{Practical Results for Deep Fully-connected Neural Networks}
\noindent \textbf{Settings}. We denote an $L$-layer deep fully-connected neural network with $N_C$-way output as :
\[ f(\x) =  \bm{W} f_{\bm{\omega},L}(\bm{x}) =\bm{W} s(\bm{\omega}_{L-2}\cdots s(\bm{\omega}_1\bm{x})),\]
\noindent The notations are as follows: $s(\cdot)$ is the activation function; $n_{h_j}$ is the number of hidden neurons for the $j$-th layer;  $\bm{\omega}_{j} \in \mathbb{R}^{n_{h_{j+1}} \times n_{h_{j}}}, j = 1,2,\cdots, L-2$ are the weights for the first $L-1$ layers; $\bm{W} \in \mathbb{R}^{n_{h_{L-1}}\times N_C }$ is the weight for the output layer. Moreover, the $i$-th output of the network is defined as :
  \[ f^{(i)}(\x) =  \bm{W}^{(i)^\top} f_{\bm{\omega},L}(\bm{x}).\]
  \noindent where $\bm{W}^{(i)^\top}$ is the i-th row of $\bm{W}$.  In the next theorem, we focus on a specific hypothesis class for such networks where the product of weight norms 
  \[\Pi_{\bm{W},\bm{\omega}} =  ||\bm{W}||_F \cdot\prod_{j=1}^{L-2}||\bm{\omega}_j||_F \] 
\noindent are no more than $\gamma$. We denote such a hypothesis class as $\Hdnn$:
  \begin{equation*}
    \begin{split}
    \Hdnn =\bigg\{&  f: f^{(i)}(\bm{x}) = \bm{W}^{(i)^\top} f_{\bm{\omega},L}(\bm{x}), ||f^{(i)}||_\infty \le R_s,~i =1,\cdots,N_C, ~ \Pi_{\bm{W},\bm{\omega}} \le \gamma\bigg\},
    \end{split}
    \end{equation*}
    To obtain the final output, we perform a softmax operation over $f(\x)$. Thus, a valid model under this setting could be chosen  from the following hypothesis class:
    \begin{equation*}
      \begin{split}
      \mathsf{soft} \circ \Hdnn =\bigg\{  g:& g^{(i)} = \dfrac{\exp(f^{(i)}(\x))}{\sum_{j=1}^{N_C} \exp(f^{(j)}(\x))},~ f \in \Hdnn  \bigg \}.
      \end{split}
      \end{equation*}
\begin{rthm8}[\textbf{Practical Generalization Bounds for Deep Models}]
Based on the assumptions of Thm.\ref{thm:gen}
 where  $s(\cdot)$ is a 1-Lipshiptz and positive homogeneous activation function.  If $\ell$ is $\phil$-Lipschitz continuous, and the input features are sampled from $\mathcal{X} \subset \mathbb{R}^{d}$, and for all $\bm{x} \in \mathcal{X}$, we have $||\bm{x}||^2_2 \le \Rx$,  for all $ f \in \mathsf{soft} \circ \Hdnn$, we have the following inequality holds with probability at least $1-\delta$:
\begin{equation*}
R_\ell(f) \le \js  + \min\bigg(\mathcal{I}_{{DNN,1}},~  \mathcal{I}_{{DNN,2}}\bigg) \cdot \sqrt{\frac{1}{N}}
\end{equation*}
where $\rchi(\bm{Y}) =\sqrt{\sumi \suminj \nicefrac{1}{\rho_i\rho_j}},~~ \xi(\bm{Y}) = \sqrt{\sumi \nicefrac{1}{\rho_i}}
, ~~\rho_i = \dfrac{\ni}{N},$
\begin{equation*}
\begin{split}
\mathcal{I}_{{DNN,1}} =  C_1\frac{\sqrt{2}}{2} \phil \cdot \frac{\rchi(\bm{Y})}{N_C-1} + \bigg( \frac{\sqrt{2}C_1\Rx\phil\gamma}{2}  \cdot C_3 +  \dfrac{C_2B }{{N_C}} \cdot \sqrt{2\log\left(\dfrac{2}{\delta}\right)}\bigg) \cdot \xi(\bm{Y}),
\end{split}
\end{equation*}
\begin{equation*}
\begin{split}
\mathcal{I}_{{DNN,2}}  =& C_1\phil\bigg( \frac{2^9}{N_C} \cdot \xi(\bm{Y})\cdot \log^{3/2}\left(K\cdot N \cdot N_C\right) \gamma \cdot \Rx \cdot (\sqrt{2\log(2)L}+1)+ 1\bigg) + C_2 \frac{ B\cdot\sqrt{{\log(\nicefrac{2}{\delta})}}\cdot \xi(\bm{Y})}{N_C} 
\end{split}
\end{equation*}
$C_1$, $C_2$ are universal constants as thm.\ref{thm:gen}, $K = e\cdot R_s$, $C_3 =\frac{\sqrt{L\log2} + \sqrt{N_C}}{\sqrt{N_C-1}}$.
\end{rthm8}

  \begin{pf}
    Follows Thm.\ref{thm:gen}, Lem.\ref{lem:Rdn}, Thm.\ref{thm:trans} and the fact that  $\hat{\mathfrak{R}}_{N\cdot N_C}(\Pi \circ \mathcal{F})\le \frac{\Rx\gamma (\sqrt{2\log(2)L}+1)}{N\cdot N_C}$ from Thm.1 in \cite{deep-gen}.
    \end{pf}

    \subsection{Practical Results for Deep Convolutional Neural Networks}

    Now we use the result in Thm.\ref{thm:chain} to derive a generalization bound for a general class of deep neural networks where fully-connected layers and convolutional layers coexist. In a nutshell, the result is essentially an application of Thm.\ref{thm:chain} to a recent idea appeared in \cite{convbound}. \\
    \textbf{General Setting.} Now we are ready to introduce the setting of the deep neural networks employed in the forthcoming theoretical analysis, which is adopted from \cite{convbound}. We focus on the deep neural networks with $N_{conn}$ fully-connected layers and $N_{conv}$ convolutional layers. The $i$-th convolutional layer has a kernel $\bm{K}^{(i)} \in \mathbb{R}^{k_i \times k_i \times c_{i-1} \times c_i}$. Recall that convolution is a linear operator. For a Given kernel $\bm{K}$, we denote its associated matrix as $op(\bm{K})$, such that $\bm{K}(\bm{x}) = op(\bm{K}) \bm{x}$. Moreover, we assume that, each time, the convolution layer is followed by a componentwise non-linear activation function and an optional pooling operation. We assume that the activation functions and the pooling operations are all $1$-Lipschitz. For the $i$-th fully-connected layer,  we denote its weight as $\bm{V}^{(i)}$. Above all, the complete parameter set of a given deep neural network could be represented as $\bm{P}= \{\bm{K}^{(1)},\cdots, \bm{K}^{(N_{conv})}, \bm{V}^{(1)}, \cdots, \bm{V}^{(N_{conn})}\}$. Again, we also assume that the loss function is $\phil$-Lipschitz and $Range\{\ell\} \subseteq [0,B]$. Finally, given two deep neural networks with parameters $\coneP$ and $\ctwoP$, we adopt a metric $d_{NN}(\cdot,\cdot)$ to measure their distance:
    \begin{equation*}
    d_{NN}(\coneP,\ctwoP) = \sum_{i=1}^{N_{conv}} ||op(\coneK) -op(\ctwoK)||_2 + \sum_{i=1}^{N_{conn}} ||\coneV - \ctwoV||_2
    \end{equation*} \\
    \textbf{Constraints Over the Parameters}. First, we define  $\mathcal{P}^{(0)}_{\nu}$ as the class for \underline{initialization of the parameters}:
    \begin{equation*}
    \mathcal{P}^{(0)}_{\nu} = \bigg\{\bm{P}: \left(\max_{i\in \{1,\cdots, N_{conv}\}}||op(\bm{K}^{(i)})||_2 \right) \le 1+\nu, ~ \left(\max_{j \in \{1,\cdots, N_{conn}\}} ||\bm{V}^{(j)}||_2\right) \le 1+\nu \bigg\}
    \end{equation*}
    Now we further assume that the learned parameters should be chosen from a class denoted by $\mathcal{P}_{\beta, \nu}$, where the distance between the learned parameter and the fixed initialization residing in $\mathcal{P}^{(0)}_{\nu}$ is no bigger than $\beta$:
    \begin{equation*}
    \mathcal{P}_{\beta, \nu} = \bigg\{\bm{P}: d_{NN}(\coneP,\ctwoP_0) \le \beta, ~\ctwoP_0 \in \mathcal{P}^{(0)}_\nu \bigg\}.
    \end{equation*}     

    The following result then provides a generalization upper bound based on Thm.\ref{thm:gen} and Lem.\ref{lem:gen_conv}.
    \begin{rthm9}
    Based on the setting in Lem.\ref{lem:gen_conv}, given dataset $\mathcal{S} = \{(\bm{x}_i,y_i)\}_{i=1}^m$, where the instances are sampled independently, for all multiclass scoring functions $f \in \mathsf{soft} \circ \mathcal{F}_{\beta,\nu}$, if $Range\{\ell\}\subseteq[0,B]$, $\forall \delta \in (0,1)$, the following inequalities hold with probability at least $1 - \delta$:
    \begin{equation*}
    \begin{split}
    R_\ell(f) \le& \js + C_1 \cdot \left(\frac{\tilde{C}\cdot\phil  \cdot R_s \cdot \xi(\bm{Y})}{N_C} \cdot \sqrt\frac{N_{par}\left(\nu N_{L} + \beta + \log \left(3{\beta\Rx{N}}\right) \right)}{{N}}\right)+ C_2 \cdot \frac{B\cdot\xi(\bm{Y})}{N_C}   \cdot \sqrt{\frac{\log(\nicefrac{2}{\delta})}{{N}}}
    \end{split}
    \end{equation*}
    where $C_1, C_2, \tilde{C}$ are universal constants.
    \end{rthm9}
\subsection{Essential Lemmas}
\subsubsection{$\mauc$ Rademacher Complexity for p-norm penalized Linear Models }

\begin{lem}[$\mauc$ Rademacher Complexity for Linear Model]\label{lem:rlin} We have:
\begin{equation}
\empsRLin \le \Rx\phil\gamma \cdot  \sqrt{2(\bar{p} -1) \cdot N_C  } \cdot \rchi(\bm{Y}) \sqrt{\frac{1}{N}} \\
\end{equation}
where 
$\rchi(\bm{Y}) =\sqrt{\sumi \suminj \frac{1}{\rho_i,\rho_j}}
, ~~\rho_i = \frac{\ni}{N}.$
\end{lem}

\begin{pf}
\begin{equation*}
\begin{split}
\empsRLin &= \expe_\sigma \left[  \supf \empcomp \right]\\
\\
&=  \expe_\sigma  \left[ \supw \sumi \suminj \sumxone \sumxtwo \frac{\sigmai_m + \sigmaj_n }{2} \cdot \ninj \cdot  \ell(\bm{W}^{(i)^\top}(\x_m-\x_n)) \right] \\
\\
& \le  \sumi \suminj \sumxtwo   \expe_{\sigma^{(i)}}\left[  \supwi    \sumxone \frac{\sigmai_m}{2} \cdot \ninj \cdot  \ell(\bm{W}^{(i)^\top}(\x_m-\x_n)) \right] \\ \\
&~~ + \sumi \sumxone    \expe_{\sigma^{(j)}}\left[  \supwi   \suminj  \sumxtwo \frac{\sigmaj_n}{2} \cdot \ninj \cdot  \ell(\bm{W}^{(i)^\top}(\x_m-\x_n)) \right] \\ \\
\end{split}
\end{equation*}
\begin{equation*}
\begin{split}
\empsRLin & \overset{(*)}{\le} \phil\sumi \suminj \sumxtwo   \expe_{\sigma^{(i)}}   \left[   \supwi\sumxone \frac{\sigmai_m}{2} \cdot \ninj \cdot  \bm{W}^{(i)^\top}(\x_m-\x_n) \right] \\ \\
&~~ + \phil\sumi  \sumxone     \expe_{\sigma^{(j)}}\left[   \supwi   \suminj \sumxtwo \frac{\sigmaj_n}{2} \cdot \ninj \cdot  \bm{W}^{(i)^\top}(\x_m-\x_n) \right] 
\end{split}
\end{equation*}
That is to say,

\begin{equation}
\begin{split}
\empsRLin & {\le} \phil\cdot\gamma \sumi  \suminj \sumxtwo   \expe_{\sigma^{(i)}}  \left[   \left\| \sumxone \frac{\sigmai_m}{2} \cdot \ninj \cdot  (\x_m-\x_n) \right\|_{\bar{p}}\right] \\ \\
&~~ + \phil\cdot\gamma \sumi \sumxone     \expe_{\sigma^{(j)}}  \left[   \left\| \suminj  \sumxtwo \frac{\sigmaj_n}{2} \cdot \ninj \cdot  (\x_m-\x_n) \right\|_{\bar{p}} \right] \\ \\
& \overset{(**)}{\le} \phil\cdot\gamma \sumi  \suminj \sumxtwo       \bigg(\expe_{\sigma^{(i)}} \left[\left\| \sumxone \frac{\sigmai_m}{2} \cdot \ninj \cdot  (\x_m-\x_n) \right\|_{\bar{p}}^{\bar{p}} \right]\bigg)^{\nicefrac{1}{\bar{p}}} \\ \\
&~~ + \phil\cdot\gamma \sumi \sumxone         \expe_{\sigma^{(j)}}  \left[\bigg(\left\|  \suminj  \sumxtwo \frac{\sigmaj_n}{2} \cdot \ninj \cdot  (\x_m-\x_n) \right\|_{\bar{p}}^{\bar{p}}\bigg)\right]^{\nicefrac{1}{\bar{p}}},  \\ \\
\end{split}
\end{equation}
where $(*)$ is due to the Talagrand contraction lemma, $(**)$ follows that fact that $()^{\nicefrac{1}{\bar{p}}}$ is concave for $\bar{p} > 1$ and Jensen's Inequality. 
\begin{equation*}
\begin{split}
\empsRLin & \overset{(***)}{\le} \phil\cdot\gamma \sqrt{\bar{p}-1} \sumi  \suminj \sumxtwo      \expe_{\sigma^{(i)}} \left[ \left\| \sumxone \frac{\sigmai_m}{2} \cdot \ninj \cdot  (\x_m-\x_n) \right\|_{2}^{2}\right]^{\nicefrac{1}{2}} \\ \\
&~~ + \phil\cdot\gamma \cdot \sqrt{\bar{p}-1} \sumi \sumxone        \expe_{\sigma^{(j)}} \left[ \left\|   \suminj  \sumxtwo \frac{\sigmaj_n}{2} \cdot \ninj \cdot  (\x_m-\x_n) \right\|_{2}^{2}\right]^{\nicefrac{1}{2}}  \\ \\
 & \overset{(****)}{\le} \phil\cdot\gamma \sqrt{\bar{p}-1} \sqrt{N\cdot N_C}          \bigg(\sumi\suminj \sumxtwo\expe_{\sigma^{(i)}} \left[ \left\| \sumxone \frac{\sigmai_m}{2} \cdot \ninj \cdot  (\x_m-\x_n)\right\|_2^2 \right]  \\ \\
 &~~~~~+ \sumxone \expe_{\sigma^{(j)}} \left[ \left\|   \suminj  \sumxtwo \frac{\sigmaj_n}{2} \cdot \ninj \cdot  (\x_m-\x_n) \right\|_{2}^{2}\right] \bigg)^{\nicefrac{1}{2}}\\ \\
& \le  \Rx \phil\gamma  \sqrt{2\frac{\bar{p}-1}{N}} \cdot \sqrt{N_C \sumi \cdot \suminj \frac{1}{\rho_i\rho_j}},
\end{split}
\end{equation*}
where $(***)$ is due to Lem.\ref{lem:kk}, and $(****)$ is due to $\forall \x \in \mathbb{R}^{N}$ $\|\x\|_1 \le \sqrt{N} \|\x\|_2$, and the last inequality follows that :
\begin{equation*}
\begin{split}
&\expe_{\sigma^{(i)}}\left[ \left\| \sumxone \frac{\sigmai_m}{2} \cdot \ninj \cdot  (\x_m-\x_n)\right\|_2^2\right] \le \sumxone \frac{1}{4n_i^2n_j^2} \|\x_m-\x_n\|_2^2 \le  \sumxone  \frac{\Rx^2}{n_i^2n_j^2}\\
&\expe_{\sigma^{(j)}} \left [\left\|\suminj \sumxtwo\frac{\sigmaj_n}{2} \cdot \ninj \cdot  (\x_m-\x_n) \right\|_{2}^{2} \right] \le \suminj \sumxtwo \frac{1}{4n_i^2n_j^2} \|\x_m-\x_n\|_2^2   \le  \suminj\sumxtwo  \frac{\Rx^2}{n_i^2n_j^2}
\end{split}
\end{equation*}

\end{pf}

\subsubsection{$\mauc$ Rademacher Complexity for A class of Deep Fully-connected Neural Networks }
\begin{lem}\label{lem:Rdn}
 we have:
\begin{equation*}
\begin{split}
&\empsRDnn \le \frac{\sqrt{2}}{2} \phil \Rx   \gamma N_C \xi(\bm{Y}) \bigg( \sqrt{\frac{L\log2(N_C-1)}{N}}+ \sqrt{\frac{N_C(N_C-1)}{N}}\bigg) \\
&~~  +  \frac{\sqrt{2}}{2} \phil N_C \rchi(\bm{Y}) \sqrt{\frac{1}{N}}
\end{split}
\end{equation*}
where $\rchi(\bm{Y}) =\sqrt{\sumi \suminj \frac{1}{\rho_i,\rho_j}}, \xi(\bm{Y}) = \sqrt{\sumi \frac{1}{\rho_i}}
, ~~\rho_i = \frac{\ni}{N}. $
\end{lem}

\begin{pf}
Similar to Lem.\ref{lem:rlin}, we can achieve the bound:
\begin{equation*}
\begin{split}
\empsRDnn &\le \phil\sumi \suminj \sumxtwo   \expe_{\sigma^{(i)}} \left[ \sup_{g \in \Hsdnn}    \sumxone \frac{\sigmai_m}{2} \cdot \ninj \cdot  ( g^{(i)}(\bm{x}_m)- g^{(i)}(\bm{x}_n) ) \right] \\ \\
&~~ + \phil\sumi  \sumxone     \expe_{\sigma^{(j)}} \left[  \sup_{g \in \Hsdnn}    \suminj \sumxtwo \frac{\sigmaj_n}{2} \cdot \ninj \cdot ( g^{(i)}(\bm{x}_m)- g^{(i)}(\bm{x}_n) ) \right] \\ \\
\end{split}
\end{equation*}
By the sub-additivity of supremum, we have: 
\begin{equation}
\begin{split}
\empsRDnn&  \le \underbrace{\phil\sumi (N_C-1) \cdot   \expe_{\sigma^{(i)}} \left[ \sup_{g \in \Hsdnn}    \sumxone \frac{\sigmai_m}{2} \cdot \frac{1}{\ni} \cdot  ( g^{(i)}(\bm{x}_m) ) \right]}_{(a)} \\ \\
&~~ + \underbrace{\phil\sumi      \expe_{\sigma^{(j)}}  \left[ \sup_{g \in \Hsdnn}   \suminj \sumxtwo \frac{\sigmaj_n}{2} \cdot \frac{1}{\nj} \cdot (  g^{(i)}(\bm{x}_n) ) \right]}_{(b)} \\ \\
&~~ + \underbrace{\phil\sumi \suminj \sumxtwo   \expe_{\sigma^{(i)}}\left[  \sup_{g \in \Hsdnn}    ( g^{(i)}(\bm{x}_n) )\sumxone \frac{\sigmai_m}{2} \cdot \ninj    \right]}_{(c)} \\ \\
&~~ + \underbrace{\phil\sumi  \sumxone     \expe_{\sigma^{(j)}}  \sup_{g \in \Hsdnn}   \left[(g^{(i)}(\bm{x}_m) ) \cdot \suminj \sumxtwo \frac{\sigmaj_n}{2} \cdot \ninj   \right]}_{(d)}
\end{split}
\end{equation}
We first derive the bound for $(a)+(b)$.\\

According Lem.\ref{lem:vec-con} and Lem.\ref{lem:soft}, we know that:

\begin{equation*}
\begin{split}
(a)+(b)  \le \phil\sumi (N_C-1) \cdot   \expe_{\sigma^{(i)}}\bigg[  &\suphd    \sumxone  \sum_{c=1}^{N_C}\frac{\sigmai_{m,c}}{2} \cdot \frac{1}{\ni} \cdot  ( f^{(c)}(\bm{x}_m) ) \bigg] \\ \\
&~~ + \phil\sumi      \expe_{\sigma^{(j)}} \left[  \suphd     \suminj \sumxtwo \sum_{c=1}^{N_C} \frac{\sigmaj_{n,c}}{2} \cdot \frac{1}{\nj} \cdot (  f^{(c)}(\bm{x}_n) ) \right]
\end{split}
\end{equation*}

Using the Lem.\ref{lem:exp} recursively towards hierarchical structure of DNN, we have:
\begin{equation*}
\begin{split}
 &\expe_{\sigma^{(i)}} \left[\suphd  \sumxone \sum_{c=1}^{N_C} \frac{\sigmai_{m,c}}{2}\cdot \ninj \cdot  ( f^{(c)}(\bm{x}_m) ) \right] \\
 &~~~~~~ \le  \frac{1}{\lambda} \log \left( \exp \left( \lambda \cdot \expe_{\sigma^{(i)}}\left[ \suphd  \sumxone \sum_{c=1}^{N_C} \frac{\sigmai_{m,c}}{2}\cdot \ninj \cdot  ( f^{(c)}(\bm{x}_m) ) \right] \right) \right) \\
  &~~~~~~ \le  \frac{1}{\lambda} \log \left( \left(  \expe_{\sigma^{(i)}} \left[ \suphd \exp   \left(\lambda \cdot \sumxone  \sum_{c=1}^{N_C} \frac{\sigmai_{m,c}}{2}\cdot \ninj \cdot  ( f^{(c)}(\bm{x}_m) )\right) \right] \right) \right) \\
 &~~~~~~\le \frac{1}{\lambda}\log\bigg(2^L \cdot  \expe_{\sigmai} \left[  \exp\left(\lambda\gamma  \left\|\sumxone \sum_{c=1}^{N_C} \frac{\sigmai_{m,c}}{2}\cdot \ninj \cdot  \x_m\right\| \right) \right] \bigg) \\
 \end{split}
\end{equation*}

\begin{equation*}
\begin{split}
&\expe_{\sigma^{(j)}} \left[   \suphd \suminj \sumxtwo \sum_{c=1}^{N_C} \frac{\sigmaj_{n,c}}{2} \cdot \ninj \cdot (  f^{(c)}(\bm{x}_n) )  \right] \\
&~~~~~~\le  \frac{1}{\lambda}\log\bigg(2^L \cdot  \expe_{\sigmaj} \left[ \exp\left(\lambda\gamma  \left\|\suminj \sumxtwo \sum_{c=1}^{N_C} \frac{\sigmaj_{n,c}}{2}\cdot \ninj \cdot  \x_n\right\| \right)\right] \bigg).
\end{split}
\end{equation*}

This suggests that:
\begin{equation*}
\begin{split}
(a+b)  \le 
\phil\sumi \frac{1}{\lambda} \log\bigg(2^L \cdot  \expe_{\sigma} \bigg[  \exp\bigg(\lambda\gamma& (N_C-1)  \bigg\|\sumxone \sum_{c=1}^{N_C}  \frac{\sigmai_{m,c}}{2}\cdot \frac{1}{\ni} \cdot  \x_m\bigg\| \\
&+ \lambda\gamma    \left\|\suminj \sumxtwo \sum_{c=1}^{N_C}  \frac{\sigmaj_{n,c}}{2}\cdot \frac{1}{\nj} \cdot  \x_n\right\|  \bigg) \bigg] \bigg),
\end{split}
\end{equation*}
which is due to the property that $\expe_{\bm{a},\bm{b}}(f(\bm{a})\cdot g(\bm{a})) =  \expe_{\bm{a}}(f(\bm{a})) \cdot \expe_{\bm{b}}(g(\bm{b}))$, if $\bm{a}$ and $\bm{b}$ are independent.
Denote 
\begin{equation}
\begin{split}
\bm{Z}_i = \gamma (N_C -1)  \bigg\| \sumxone  \sum_{c=1}^{N_C} & \frac{\sigmai_{m,c}}{2}\cdot \frac{1}{\ni} \cdot  \x_m\bigg\| 
+ \gamma   \left\|\suminj \sumxtwo \sum_{c=1}^{N_C}  \frac{\sigmaj_{n,c}}{2}\cdot \frac{1}{\nj} \cdot  \x_n\right\|  
\end{split}
\end{equation}
as a random variable with respect to the Rademacher random variables. We have: 
\begin{equation*}
\begin{split}
(a) + (b) &\le \phil\left(\sumi \frac{L\log2+ \log(\expe_{\sigma}\exp(\lambda(\bm{Z}_i-\mathbb{E}(\bm{Z}_i)))}{\lambda} +\sumi \expe_{\sigma}(\bm{Z}_i) \right).\\ 
\end{split}
\end{equation*}

Similar to the proof of Lem.\ref{lem:rlin}, we can bound
\begin{equation}
\sumi \expe_{\sigma}(\bm{Z}_i)\le \frac{\sqrt{2}}{2}  \gamma \cdot\Rx   \cdot (N_C)^{3/2} \left((N_C-1) \sumi \cdot \suminj \frac{1}{\rho_i}\right)^{1/2} \cdot \frac{1}{\sqrt{N}}.
\end{equation}
Utilizing the bounded difference inequality (Lem.\ref{lem:bdp}) of $\bm{Z}_i$, we come to the following result:
\begin{equation}
\expe_{\sigma}\exp\left(\lambda(\bm{Z}_i-\mathbb{E}(\bm{Z}_i)\right) \le \exp\left( \frac{\lambda^2 v_i}{2}\right), ~~ v_i \le \frac{N_C}{4} \frac{(N_C - 1)^2}{\ni}\gamma^2\Rx^2 + \frac{N_C}{4}\suminj \frac{\gamma^2\Rx^2}{\nj},
\end{equation}
which leads to the bound:
\begin{equation}
\left(\sumi \frac{L\log2+ \log(\expe_{\sigma}\exp(\lambda(\bm{Z}_i-\mathbb{E}(\bm{Z}_i)))}{\lambda}\right) \le \left(\sumi \frac{L\log2+ ( \dfrac{\lambda^2 v_i}{2})}{\lambda}\right) 
\end{equation}
By choosing $\lambda = \sqrt{\frac{2N_CL\log2}{\sumi v_i}}$, we reach the optimal bound as:
\begin{equation}
\begin{split}
\sumi \frac{L\log2+ \log(\expe_{\sigma}\exp(\lambda(\bm{Z}_i-\mathbb{E}(\bm{Z}_i)))}{\lambda}& \\ \\
&~~~\le \sqrt{{2N_CL\log2}\cdot {\sumi v_i}} \\
&~~~~ =  R \cdot \gamma \cdot \sqrt{\frac{L\log2}{2N}} \cdot \sqrt{\sumi \frac{N_C-1}{\rho_i}}.
\end{split}
\end{equation}
Putting all together, we have
\begin{equation}
\begin{split}
  (a) +(b) \le \frac{\sqrt{2}}{2} \phil  \Rx  \gamma \cdot  (N_C(N_C-1))^{1/2} \xi(\bm{Y})  \left(  \sqrt{\frac{N_CL\log2}{N}}  + N_C\sqrt{\frac{1}{N}}  \right)
\end{split}
\end{equation}

For (c) and (d), we have:

\begin{equation}\label{eq:dc}
\begin{split}
(c) &\le \phil\sumi \suminj \sumxtwo   \expe_{\sigma^{(i)}} \left[ \left(\suphd  \left|f^{(i)}(\bm{x}_n) )\right|\right)  \cdot \left| \sumxone \frac{\sigmai_m}{2} \cdot \ninj    \right|\right] \\ \\
& \overset{(*)}{\le} \phil\cdot \sumi \suminj \sumxtwo   \expe_{\sigma^{(i)}} \left[   \left| \sumxone \frac{\sigmai_m}{2} \cdot \ninj    \right|\right]
\end{split}
\end{equation}
Similarly, we have :
\begin{equation}\label{eq:dd}
\begin{split}
(d) \overset{(**)}{\le} \phil\sumi \sumxone  \expe_{\sigma^{(j)}} \left[   \left|  \suminj \sumxtwo    \frac{\sigmaj_m}{2} \cdot \ninj    \right|\right],
\end{split}
\end{equation}
where (*) and (**) are due to the fact that the softmax function is upper bounded by 1, i.e.,

 \[\suphd |f^{(i)}_{W,L}(\bm{x}_n) | \le 1 ~~ \text{and} ~~ \suphd |f^{(i)}_{W,L}(\bm{x}_m) | \le 1.\]

 This finishes the proof for Eq.(\ref{eq:dc}) and Eq.(\ref{eq:dd}).\\

 Now we can reach that:
\begin{equation}
\begin{split}
(c) +(d) & \le \phil\cdot  \sumi \suminj \sumxtwo   \expe_{\sigma^{(i)}} \left[   \left| \sumxone\sum_{c=1}^{N_C} \frac{\sigmai_{m,c}}{2} \cdot \ninj    \right|\right] \\ \\
&~~~~+ \phil\cdot \sumi \sumxone   \expe_{\sigma^{(j)}} \left[   \left|  \suminj \sumxtwo  \sum_{c=1}^{N_C}  \frac{\sigmai_{n,c}}{2} \cdot \ninj    \right|\right] \\  \\
&\le \phil\cdot \sumi \suminj \sumxtwo   \sqrt{\expe_{\sigma^{(i)}} \left[   \left( \sumxone  \sum_{c=1}^{N_C} \frac{\sigmai_{m,c}}{2} \cdot \ninj    \right)^2\right]} \\ \\
&~~~~+ \phil\cdot  \sumi \sumxone  \sqrt{\expe_{\sigma^{(j)}} \left[   \left(  \suminj \sumxtwo   \sum_{c=1}^{N_C}  \frac{\sigmai_{n,c}}{2} \cdot \ninj    \right)^2\right]}
\end{split}
\end{equation}
Similar to the proof of \ref{lem:rlin}, we have:
\begin{equation}
(c) +(d)  \le \frac{\sqrt{2}}{2} \cdot  \phil   \cdot \rchi(\bm{Y}) \cdot N_C \sqrt{\frac{1}{N}}
\end{equation}

\end{pf}

Now we provide an alternative upper bound by the chaining technique.

\begin{lem}\label{Lem:N_Cdeep}
Under the setting of Thm.\ref{thm:gen_3}.

 For all $ f \in \mathsf{soft} \circ \Hdnn$, we have the following inequality holds with probability at least $1-\delta$:
  \begin{equation*}
  R_\ell(f) \le \js  +  \mathcal{I}_{{DNN,2}}  \cdot \sqrt{\frac{1}{N}}
  \end{equation*}
  where $\xi(\bm{Y}) = \sqrt{\sumi \nicefrac{1}{\rho_i}}
  , ~~\rho_i = \dfrac{\ni}{N},$
  \begin{equation*}
  \begin{split}
    \mathcal{I}_{{DNN,2}}   =& C_1\phil\bigg( \frac{2^9}{N_C} \cdot \xi(\bm{Y})\cdot \log^{3/2}\left(K\cdot N \cdot N_C\right) \cdot \gamma \cdot \Rx \cdot (\sqrt{2\log(2)L}+1)+ 1\bigg) + C_2 \frac{ B\cdot\sqrt{{\log(\nicefrac{2}{\delta})}}\cdot \xi(\bm{Y})}{N_C} 
  \end{split}
  \end{equation*}
  $C_1$, $C_2$ are universal constants as thm.\ref{thm:gen}, $K = e\cdot R_s$.
\end{lem}

\begin{pf}
The result follows from Thm.\ref{thm:trans}, and the fact that 
\[\hat{\mathfrak{R}}_{N\cdot N_C}(\Pi \circ \mathcal{F}) \le \frac{\Rx\gamma\cdot(\sqrt{2\log(2)L}+1)}{\sqrt{N_C\cdot N}},
\]
which follows from Thm.1 in \cite{deep-gen}
\end{pf}

\subsubsection{$\mauc$ Rademacher Complexity for A  Class of Deep Convolutional Neural Networks}

\begin{lem}\label{lem:gen_conv} Denote the hypothesis class,
\[ \mathsf{soft} \circ \mathcal{F}_{\beta,\nu} = \bigg\{\bm{g}(\x) = \mathsf{soft}(\bm{s}_{\bm{P}}(\x)):~ \bm{s}_{\bm{P}} \in\mathcal{F}_{\beta,\nu}\bigg\}, \]

\[\mathcal{F}_{\beta,\nu} = \{s_{\bm{P}}:\mathbb{R}^{N_{N_L-1}} \rightarrow \mathbb{R}^{N_C}| ~ \bm{P} \in \mathcal{P}_{\beta, \nu}, Range(\bm{s}_P) \subseteq [-R_s, R_s]^{N_C} \}\]

Moreover, define $\tilde{N} = \dfrac{1}{\sum_{i=1}^{N_C}\dfrac{1}{n_i}}$ we assume that $\sup_{x\in\mathcal{X}} ||vec(\bm{x})|| \le \Rx$ and 
\[R_s >1/\min\left\{\sqrt{{N}}, \frac{\xi(\bm{Y})}{N_C} \cdot \sqrt{N_{par}\left(\nu N_L + \beta + \log (3\Rx\cdot \beta \cdot {N}) \right) } \right\},\]  
we have:
\begin{equation}
\empsRFb \le \tilde{C} \left(\phil \cdot (N_C-1) \cdot R_s \cdot \xi(\bm{Y}) \cdot \sqrt\frac{N_{par}\left(\nu N_{L} + \beta + \log\left(3{\beta\Rx{N}}\right) \right)}{{N}} \right),
\end{equation}
where $N_L = N_{conv} + N_{conn}$, $N_{par}$ is total number of parameters in the neural network, and $\tilde{C}$ is a universal constant.
\end{lem}\label{lem:gen_conv}
\begin{pf}
From \cite[\color{org}{Lem.3.4}]{convbound}, we know that:
\begin{equation*}
\max_{z \in \mathcal{Z}_{\mathcal{D}}}|\bm{\cone{s}}_{\coneP}(\bm{z}) - \ctwo{\st}_{\ctwoP}(\bm{z})| \le \sup_{x \in \mathcal{X}}||\bm{\cone{s}}_{\coneP}(\bm{x}) - \ctwo{\st}_{\ctwoP}(\bm{x})||\le \Rx \cdot \beta\cdot\left(1+\nu + \frac{\beta}{N_L}\right)^{N_L} \cdot d_{NN}(\coneP, \ctwoP).
\end{equation*}
Furthermore, according to  \cite[\color{org}{Lem.A.8}]{convbound}, we have:
\begin{equation*}
\log(\mathfrak{C}(\epsilon,\mathcal{F}_{\beta,\nu}, d_{\infty, \mathcal{S}}) \le N_{par} \cdot \log\left(\frac{3C_L}{\epsilon}\right),
\end{equation*}
where $C_L = \Rx \cdot \beta \exp(\nu N_L + \beta)$. Following Thm.\ref{thm:chain}{\color{blue}-b)}, we have:
\[\empsR \le C\cdot \inf_{ R_s \ge \alpha \ge 0} \left(N_C \cdot(N_C-1) \cdot \phil \cdot \alpha + \phil \cdot (N_C-1) \cdot  \int_{\alpha}^{R_s} \sqrt\frac{N_{par}\left(\nu N_{L} + \beta + \log (3\beta\Rx/\epsilon) \right)}{\tilde{N}} d\epsilon  \right)\]
Since $R_s > 1/\sqrt{N}$, we can choose $\alpha = \sqrt{\frac{1}{{N}}}$, the result follows from the inequality that:
\begin{equation*}
\int_{\alpha}^{R_s} \sqrt\frac{N_{par}\left(\nu N_{L} + \beta + \log (3\beta\Rx/\epsilon) \right)}{\tilde{N}} d\epsilon  \le  R_s \cdot \xi(\bm{Y}) \cdot \sqrt\frac{N_{par}\left(\nu N_{L} + \beta + \log \left(3{\beta\Rx{N}}\right) \right)}{{N}}.
\end{equation*}
Together with the fact that  $R_s >1/ \left(\frac{\xi(\bm{Y})}{N_C} \cdot \sqrt{N_{par}\left(\nu N_L + \beta + \log (3\Rx\cdot \beta \cdot {N}) \right) } \right),$ we have :
\begin{equation}
  \empsRFb \le 2 {C} \left(\phil \cdot (N_C-1) \cdot R_s \cdot \xi(\bm{Y}) \cdot \sqrt\frac{N_{par}\left(\nu N_{L} + \beta + \log\left(3{\beta\Rx{N}}\right) \right)}{{N}} \right).
\end{equation}
The proof is completed by choosing $\tilde{C} = 2 C$.
\qed
\end{pf}
\noindent 

\section{Efficient Computation}\label{sec:appe}
 In this section, we will  propose acceleration algorithms for loss and gradient evaluation for three well-known losses: exponential loss $\ell_{exp}(\alpha,t) = \exp(-\alpha t)$, squared loss $\ell_{sq}(t) = (\alpha- t)^2$, and hinge loss  $\ell_{hinge}(\alpha,t) = \max(\alpha-t,0)$.

Generally, the empirical surrogate risk functions $\hat{R}_{\ell} $ have the following abstract form:
\begin{equation*}
\hat{R}_{\ell} = \sumi \suminj \sumxone \sumxtwo \ninj \cdot   \ell(f^{(i)}(\x_m) - f^{(i)}(\x_n)),
\end{equation*}
\begin{equation*}
\text{where}~~ \fii = g_i(
\bm{W}\hth), \bm{W}= [\bm{w}^{(1)},\cdots, \bm{w}^{(N_C)}]^\top.
\end{equation*}
We adopt this general form since it covers a lot of popular models. For example, when $g_i(\cdot)$ is defined as the activation function of the last layer of a neural network (say a softmax function), $\bm{w}^{(i)}$ are the weights of the last layer, and $\hth[\cdot]$ is the neural net where the last layer is excluded, $\fii$ becomes a deep neural network architecture with the last layer designed as a fully-connected layer. As for another instance, if both $g_i(\x)= \x$ and $\hth =\x$, then we reach a simple linear multiclass scoring function. Note that the scalability with respect to sample size only depends on the choice of $\ell$. The choice of $g_i(\cdot)$ and  $\hth[\cdot]$ only affects the instance-wise chain rule. This allows us to provide a general acceleration framework once the surrogate loss is fixed. 

\textbf{Common Notations}. Now we introduce the common notations we adopted in this section. We assume that $\bm{w}^{(i)}$ be a vector such that $\bm{w}^{(i)} \in \mathbb{R}^{n_h \times 1}$ . We let $\bm{W}$ be the matrix that concatenates  $\bm{w}^{(i)}$ column-wisely, \emph{i.e.} ,
\begin{equation}\label{eq:weq}
\bm{W} = [\bm{w}^{(1)},\cdots, \bm{w}^{(N_C)} ]^\top, ~~\bm{W} \in \mathbb{R}^{N_C \times n_h}.
\end{equation} 
 For the sake of convenience, we rearrange $\bm{\theta}$ as a vector in $\mathbb{R}^{d_\theta \times 1}$ such that it concatenates all the parameters in the model except $\bm{W}$. For example, in a deep neural network, $\bm{\theta}$ concatenates all the parameters except the weights of the last layer. Note that this does not affect the result of the loss and gradient evaluation. A simple reshape suffices to transfer our calculation of gradient to its expected result if $\bm{\theta}$ has complicated structures. $\hth$ should be considered as a vector such that $\hth \in \mathbb{R}^{n_h \times 1}$.  $\bm{H}$ is then defined as the matrix that concatenates all the $\hth[\x]$ from every instance in the sense that: 
\begin{equation}\label{eq:heq}
\bm{H} = [\hth[\bm{x}_1],\cdots,\hth[\bm{x}_{N}]]^\top, ~~ \bm{H} \in \mathbb{R}^{n_h \times N}.
\end{equation}

For intermediate partial derivatives, we denote:
\begin{equation}\label{eq:paritala}
\begin{split}
\pfij[\x] = \frac{\partial(\fjj)}{\partial \bm{w}^{(i)^\top} \hth[\x] }
\end{split}
\end{equation}
Furthermore, we have two compact expressions as:
\begin{equation}\label{eq:parital}
\begin{split}
&\bm{\partial}_{i,j} = [\pfij[\x_1], \cdots, \pfij[\x_N]]^\top, ~~ \bm{\partial}_{i,j} \in \mathbb{R}^{N \times 1}, \\ 
&\bm{\partial}_j(\x_k) =  [\partial_1[\fjj[\x_k]], \cdots,\partial_{N_C}(\fjj[\x_k]) ]^\top, ~~ \bm{\partial}_j(\x_k) \in \mathbb{R}^{N_C \times 1}.
\end{split}
\end{equation}
Another variable is also essential when applying the chain rule:
\begin{equation}\label{eq:udfin}
\bm{U}^{(j)} = [ \nabla_{\bm{\theta}}\hth[\x_1]\bm{W}^\top\bm{\partial}_j(\x_1), \cdots, \nabla_{\bm{\theta}}\hth[\x_N]\bm{W}^\top\bm{\partial}_j(\x_N)], ~~ \bm{U}^{(j)} \in \mathbb{R}^{d_\theta \times N}.
\end{equation}
Finally, we provide a compact notation of the weights $\frac{1}{\ni\nj}$ as:
\begin{equation}\label{eq:dlabel}
\bm{D}^{(i)} = [\frac{1}{n_in_{y_1}},\cdots, \frac{1}{n_in_{y_N}} ]^\top ,~~  \bm{D}^{(i)} \in \mathbb{R}^{N \times 1}.
\end{equation}

\subsection{Exponential Loss}
\textbf{Loss Evaluation}. For the exponential loss, we can simplify the computations by a factorization scheme:
\begin{equation*}
\begin{split}
\hat{R}_{exp} &= \sumi\sumxone  \suminj \sumxtwo \ninj \cdot  \exp\left(\alpha \cdot \left(\dfi \right)\right),\\
& = \sumi\underbrace{\left(\sumxone \exp(\alpha\cdot \fixm)\right)}_{(a_i)} \cdot \underbrace{\left(\suminj \sumxtwo \ninj \cdot  \exp(-\alpha\cdot\fixn)\right)}_{(b_i)}
\end{split}
\end{equation*}
From the derivation above, the loss evaluation could be done by first calculating $(a_i),(b_i)$ separately and then performing the multiplication, which only takes $O(NN_CT_\ell)$.  This is a significant improvement compared with the original $O(\sumi\suminj n_in_j N_C T_\ell)$ result.

\textbf{Gradient Evaluation}. Due to the factorization scheme, we can also accelerate the gradient evaluation in a similar spirit. According to. Eq.(\ref{eq:weq})-Eq.(\ref{eq:udfin}), we have:
\begin{equation*}
\begin{split}
\nabla_{\bm{w}^{(i)}}(a_j) &= \sumxone \alpha  \exp(\alpha \cdot \fjxm) \cdot \pfij[\x_m] \cdot \hth[\x_m] \\
 &=\alpha\bm{H}(\mathcal{I}_i \odot \hat{\bm{y}}^{(i),+}_{exp} \odot \bm{\partial}_{i,j} ).\\
  \nabla_{\bm{w}^{(i)}}(b_j)   &= -\suminj \sumxtwo \ninj \alpha \cdot \exp(-\alpha \cdot \fjxn) \cdot \pfij[\x_n] \cdot  \hth[\x_n]\\
  &= -\alpha\bm{H}(\mathcal{I}_{\backslash i} \odot \bm{D}^{(i)} \odot \hat{\bm{y}}^{(i),-}_{exp} \odot  \bm{\partial}_{i,j} ), \\
 \nabla_{\hth[\x_k]}(a_j) &= \underbrace{\mathcal{I}_j(\x_k) \cdot \alpha \cdot \exp(\alpha \cdot \fjj[\x_k] )}_{q_j(\x_k)}\cdot \sum_{i=1}^{N_C}\pfij[\x_k]  \cdot \wi, \\
 & =  {q}_j(\x_k) \cdot \bm{W}^\top \bm{\partial}_j(\x_k)\\
  \nabla_{\hth[\x_k]}(b_j) &= -\underbrace{\mathcal{I}_j^c(\x_k)  \cdot \bm{D}^{(i)}_{k} \cdot \alpha  \cdot \exp(-\alpha \cdot \fjj[\x_k] )}_{{\tilde{q}_j(\x_k)}} \cdot \sum_{i=1}^{N_C} \pfij[\x_k] \cdot \wi,\\
  & = -\tilde{q}_j(\x_k) \cdot \bm{W}^\top \bm{\partial}_j(\x_k)
\end{split}
\end{equation*}
where 
\begin{equation*}
\begin{split}
& \mathcal{I}_i = I\left[\bm{x}_i \in \Ni\right], ~ \mathcal{I}_{\backslash i} = I\left[\bm{x}_i \notin \Ni\right],~~ \mathcal{I}_j(\x) = \bm{I}\left[ x \in \Nj \right],~~  \mathcal{I}^c_j(\x) = 1 - \mathcal{I}_j(\x)\\
&\hat{\bm{y}}^{(i),+}_{exp} = [\exp(\alpha\cdot f^{(i)}(\bm{x}_1)), \cdots, \exp(\alpha \cdot f^{(i)}(\bm{x}_N))]^\top, \\
&\hat{\bm{y}}^{(i),-}_{exp} = [\exp(-\alpha\cdot f^{(i)}(\bm{x}_1)), \cdots, \exp(-\alpha \cdot f^{(i)}(\bm{x}_N))]^\top. \\ 
\end{split}
\end{equation*}
Furthermore, we define  $\bm{q}_j = [q_j(\x_1),\cdots, q_j(\x_N)]^\top$
 and $\bm{\tilde{q}}_j = [\tilde{q}_j(\x_1),\cdots, \tilde{q}_j(\x_N)]^\top$. From Eq.(\ref{eq:udfin}), we have:
\begin{equation*}
\begin{split}
\nabla_{\bm{\theta}}(a_j) &= \sum_{k=1}^N \nabla_{\bm{\theta}}(\hth[\x_k]) \nabla_{\hth[\x_k]}a_j= \bm{U}^{(j)}\bm{q}_j \\
\nabla_{\bm{\theta}}(b_j) &= \sum_{k=1}^N \nabla_{\bm{\theta}}(\hth[\x_k]) \nabla_{\hth[\x_k]}b_j= - \bm{U}^{(j)}\bm{\tilde{q}}_j
\end{split}
\end{equation*}

Then the gradient evaluation can be done by :
\begin{equation*}
\begin{split}
\nabla_{\wi}\hat{R}_{\ell} &= \sum_{j=1}^{N_C} \big(\nabla_{\wi}(a_j)\big) \cdot (b_j) + (a_j) \cdot \big(\nabla_{\wi}(b_j)\big) \\
&= \alpha \bm{H} \left(\sum_{j=1}^{N_C} b_j \mathcal{I}_i \odot \hat{\bm{y}}^{(i),+}_{exp} \odot \bm{\partial}_{i,j} - a_j \mathcal{I}_{\backslash i} \odot \bm{D}^{(i)} \odot \hat{\bm{y}}^{(i),-}_{exp} \odot \bm{\partial}_{i,j} \right),\\
\nabla_{\bm{\theta}}\hat{R}_{\ell} &= \sum_{j=1}^{N_C} \big(\nabla_{\bm{\theta}}(a_j)\big) \cdot (b_j) + (a_j) \cdot\big(\nabla_{\bm{\theta}}(b_j)\big) \\
&= \sum_{j=1}^{N_C} \left[\bm{U}^{(j)}(b_j\bm{q}_j -a_j\bm{\tilde{q}}_j)\right].
\end{split}
\end{equation*}

From the compact forms shown above, the acceleration enjoys a  $O(NN_CT_{grad})$ complexity  instead of the original $O(N_C\sumi\suminj n_in_jT_{grad})$ time.
\subsection{Hinge loss}
\textbf{Loss Evaluation}. First we put down the hinge surrogate loss as:
\begin{equation*}
\hat{R}_{hinge} = \sumi\sumxone  \suminj \sumxtwo \ninj \cdot  \left(\alpha - \left(\fixm - \fixn\right)\right)_+. 
\end{equation*}
The key of our acceleration is to notice that the terms are non-zero only if $\left(\fixm - \fixn\right) \le \alpha$. Moreover, for these non-zero terms $\max(x,0) =x$. This means that the hinge loss degenerates to an identity function for the activated nonzero terms, which enjoys efficient computation. So the key step in our algorithm is to find out the non-zero terms in an efficient manner.

Given a fixed class $i$ and an instance $\x_m \in \Ni$, we denote: 

\[\mathcal{A}^{(i)}(\x_m) =\{\x_n \notin \Ni, \alpha > \fixm -\fixn \}.\]
With $\mathcal{A}^{(i)}(\x_m)$, one can reformulate $\hat{R}_{hinge}$ as:
 \begin{equation*}
 \begin{split}
 \hat{R}_{hinge} &= \sumi\sumxone \left[\left(\suminj \sum_{\x_n \in \Nj \cap \mathcal{A}^{(i)}(\x_m) } \ninj\right) \cdot  \left(\alpha - \fixm \right) + \suminj\sum_{\x_n \in \Nj \cap \mathcal{A}^{(i)}(\x_m) } \ninj \cdot \fixn  \right],\\
 & =  \sumi\sumxone \left[\delta^{(i)}(\x_m) \cdot (\alpha - \fii[\x_m])+ \Delta^{(i)}(\x_m)\right],
 \end{split}
\end{equation*}
where 
\[\delta^{(i)}(\x_m)= \suminj \sum_{\x_n \in \Nj \cap \mathcal{A}^{(i)}(\x_m) } \ninj, ~~ \Delta^{(i)}(\x_m) = \suminj \sum_{\x_n \in \Nj \cap \mathcal{A}^{(i)}(\x_m) }  \ninj \fixn. \]
According to this reformulation, once $\delta^{(i)}(\x)$ and $\Delta^{(i)}(\x)$ are all fixed, we can calculate the loss function within $O(N)$. So the efficiency bottleneck comes from the calculations of $\delta^{(i)}(\x)$ and $\Delta^{(i)}(\x)$.

Next we propose an efficient way to calculate  $\delta^{(i)}(\x_m), \Delta^{(i)}(\x_m), \mathcal{A}^{(i)}(\x_m)$.  To do so, given a specific class $i$, we first sort the relevant (resp. irrelevant) instances descendingly according to their scores $\fii$:
\begin{equation*}
\begin{split}
&\fii[\bm{x}^\downarrow_0] \ge  \fii[\bm{x}^\downarrow_1] \cdots, \ge \fii[\bm{x}^\downarrow_{n_i-1}],~~~~  \bm{x}^\downarrow_i \in \mathcal{N}_i,  \\
&\fii[\bm{\tilde{x}}^\downarrow_0] \ge  \fii[\bm{\tilde{x}}^\downarrow_1] \cdots, \ge \fii[\bm{\tilde{x}}^\downarrow_{N-n_i-1}],~~~~ \bm{\tilde{x}}^\downarrow_i \notin \mathcal{N}_i.
\end{split}
\end{equation*}

It immediately follows that:
\begin{equation*}
\mathcal{A}^{(i)}(\bm{x}^\downarrow_0) \subseteq \mathcal{A}^{(i)}(\bm{x}^\downarrow_1) \subseteq \cdots \subseteq \mathcal{A}^{(i)}(\bm{x}^\downarrow_{\ni-1}). 
\end{equation*}
This allows us to find out all $\mathcal{A}^{(i)}(\bm{x}^\downarrow_k), k=0,1,\cdots,n_i-1,$ within a single pass of the whole dataset with an efficient dynamic programming.

Based on the construction of $\mathcal{A}^{(i)}_k = \mathcal{A}^{(i)}(\bm{x}^\downarrow_k)$, we provide the following recursive rules to obtain $\delta^{(i)}(\x)$ and $\Delta^{(i)}(\x)$:
\begin{equation*}
\begin{split}
&\delta^{(i)}(\x^\downarrow_{k+1}) = \delta^{(i)}(\x^\downarrow_{k}) +  \suminj \sum_{ \Nj \cap  \mathcal{A}^{(i)}_{k+1} \backslash  \mathcal{A}^{(i)}_k } \ninj, \\
&\Delta^{(i)}(\x^\downarrow_{k+1}) = \Delta^{(i)}(\x^\downarrow_{k}) +  \suminj \sum_{\x_n \in \Nj \cap \mathcal{A}^{(i)}_{k+1} \backslash  \mathcal{A}^{(i)}_k  } \ninj \fixn.
\end{split}
\end{equation*}
These recursive rules induce an efficient dynamic programming which only requires a single pass of the whole dataset. Putting all together, we then come to a $O(N_C\bar{N}T_{l})$ time complexity speed-up for hinge loss evaluation. An implementation of this idea is detailed in  Alg.\ref{alg:index} and Alg.\ref{Alg:all}. 

\begin{algorithm}[t]
  \caption{\label{alg:index}\texttt{CalIndex}}
  \begin{algorithmic}[1]
    \Require $\x^\downarrow_0,\cdots, \x^\downarrow_{\ni-1}, \bm{\tilde{x}}^\downarrow_0, \cdots, \bm{\tilde{x}}^\downarrow_{N-\ni-1}, \alpha$.
    \State $p\gets0$ \Comment{pointer for current positive instance }
    \State $q\gets 0$ \Comment{pointer for current negative instance }
    \State $\mathcal{I}_{0:(\ni-1)}= {0}$.  \Comment{The index set that satisfies $\mathcal{A}^{(i)}_j = \{\xneg{i}{k}\}_{k=1}^{\mathcal{I}_k} $ }
    \While{$p < \ni, q < N - \ni $}
    \If{$\fii[\xpos{i}{p}] - \fii[\xneg{i}{q}] < \alpha$} 
     \State $q+=1$ \Comment{If the current pair is nonzero, check the next  negative instance }
    \Else
    \State $\mathcal{I}_p  = \max(q-1,0)$  \Comment{The last element of $\mathcal{A}^{(i)}_p$ is $\xneg{i}{q-1}$ }
    \State $p+=1$                     
    \EndIf
    \EndWhile
  \If{$q = N - \ni$}
  \State$\mathcal{I}_p = \max(q-1,0)$ \Comment{The last element of $\mathcal{A}^{(i)}_p$ is $\xneg{i}{N-\ni-1}$ }
  \EndIf
  \If{$p < \ni$} \Comment{If $\mathcal{A}^{(i)}_p$ includes all the negative instances then for all $\mathcal{A}^{(i)}_{p'} =\mathcal{A}^{(i)}_p$, $p' > p$   }
  \State $\mathcal{I}_{(p+1):(\ni-1)} = \mathcal{I}_p$ 
  \EndIf
  \end{algorithmic}
\end{algorithm}

\begin{algorithm}[t]
  \caption{\label{Alg:all}\texttt{CalHingeLoss}}
  \begin{algorithmic}[1]
    \Require $\bm{X}, \bm{Y},\alpha$.
    \Ensure \texttt{loss}.
    \State $\texttt{loss} \gets 0$
    \For{i}{1}{N_C}
    \\
    \State Obtain $\xpos{i}{0},\cdots, \xpos{i}{\ni-1} $ based on $\fii$
    \State  Obtain $\xpos{i}{0},\cdots, \xpos{i}{N-\ni-1}$ based on $\fii$
    \\
    \State  $\fii_+ = [\fii[\xpos{i}{0}],\cdots,\fii[\xpos{i}{\ni-1}]]$ .
    \State  $\bm{D}^{(i)}  = \left[\dfrac{1}{\ni n_{y(\xneg{i}{0})}},\cdots, \dfrac{1}{\ni n_{y(\xneg{i}{N-1})}}\right]$
    \\
    \State  $t_0  \gets 0,~~ \delta^{(i)}_{\rm{offset}} \gets 0,~ \Delta^{(i)}_{\rm{offset}} \gets 0, ~~loc= 0 $
    \\
    \State $\bm{\Delta}^{(i)}_{0:(\ni-1)} \gets {0},~~ \bm{\delta}^{(i)}_{0:(\ni-1)} \gets 0 $
    \\
    \State  $\mathcal{I} \gets$ \texttt{CalIndex}($\xpos{i}{0},\cdots, \xpos{i}{\ni},\xneg{i}{0},\cdots, \xneg{i}{N-\ni-1},\alpha$)\\
    \While{$k < \ni$}
    
    \If{$\mathcal{I}_k \neq t_0 -1$} \Comment{continue scanning when $\mathcal{A}^{(k)} \neq \mathcal{A}^{(k-1)} $  }
    
    \State $t_1 \gets \mathcal{I}_k +1$  \Comment{make sure that $\mathcal{A}^{(i)}_{k+1} \backslash \mathcal{A}^{(i)}_k = \{\xneg{i}{u}\}_{u=t_0}^{t_1 -1} $}
    
    \State ${\delta}^{(i)}_k \gets \delta^{(i)}_{\rm{offset}} + \sum_{u=t_0}^{t_1-1} \bm{D}^{(i)}_{u}$ \Comment{Recursion for $\delta^{(i)}(\xpos{i}{k})$}
    
    \State  ${\Delta}^{(i)}_k \gets \Delta^{(i)}_{\rm{offset}} + \sum_{u=t_0}^{t_1-1} \bm{D}^{(i)}_{u} \cdot \fii[\xneg{i}{k}]$ \Comment{Recursion for $\Delta^{(i)}(\xpos{i}{k})$}
    
    \State $t_0 \gets t_1$ \Comment{Update the initial element.}
    
    \State $ \delta^{(i)}_{\rm{offset}} \gets {\delta}^{(i)}_k, ~~ \Delta^{(i)}_{\rm{offset}} \gets {\Delta}^{(i)}_k$ \Comment{Update the offsets.}

    \ElsIf{$I_k \neq N-\ni -1$ }\Comment{Check if $\mathcal{A}^{(k)} = \mathcal{A}^{(k-1)} $  and  $\mathcal{A}^{(k)} \neq \mathcal{N}_{\backslash i}$}
    
     \State $\delta^{(i)}_k =\delta^{(i)}_{\rm{offset}}, ~~\Delta^{(i)}_k =\Delta^{(i)}_{\rm{offset}}  $ \Comment{Copy the last updates}
    
    \Else 
    
    \State \texttt{break} \Comment{Stop the iteration when $\mathcal{A}^{(k)} = \mathcal{N}_{\backslash i}$ and $\mathcal{A}^{(k)} = \mathcal{A}^{(k-1)}$. }
    
    \EndIf
    
    \State $k +=1$ 
    
    \EndWhile
    \If{$k < \ni$}
    \State $\bm{\delta}^{(i)}_{k:(\ni-1)} = \delta^{(i)}_{\rm{offset}}, ~~ \bm{\Delta}^{(i)}_{k:(\ni-1)} = \Delta^{(i)}_{\rm{offset}}$ \Comment{Set  $\mathcal{A}^{(i)}_{x} = \mathcal{A}^{(i)}_{k-1}, ~ \forall x \ge k$. }
    \EndIf
    \\
    \State $\texttt{loss} += (\alpha - \fii_+ )^\top\bm{\delta}^{(i)} + \bm{\Delta}^{(i)}\bm{1} $
    \\
    \EndFor
  \end{algorithmic}
\end{algorithm}

\textbf{Gradient Evaluation}. Similar to the loss evaluation, we can provide an efficient acceleration method for the gradient evaluation as well. Based on a similar analysis, we have:

 \begin{equation*}
 \begin{split}
\nabla_{\bm{\Theta}}\hat{R}_{hinge} & = \sumi\sumxone \left[-\left(\suminj \sum_{\x_n \in \Nj \cap \mathcal{A}^{(i)}(\x_m) } \ninj\right) \cdot \nabla_{\bm{\Theta}}\fixm   + \suminj \sum_{\x_n \in \Nj \cap \mathcal{A}^{(i)}(\x_m) } \ninj \nabla_{\bm{\Theta}}\fixn \right]\\
 &~~~~ =  \sumi\sumxone \left(-\delta^{(i)}(\x_m) \cdot \nabla_{\bm{\Theta}}\fixm + \Gamma^{(i)}(\x_m)\right)\\
 \end{split}
\end{equation*}
where 
\[\Gamma^{(i)}(\x_m) = \suminj \sum_{\x_n \in \Nj \cap \mathcal{A}^{(i)}(\x_m) }  \ninj  \nabla_{\bm{\Theta}}\fixn. \]

For any $\x \in \mathcal{S}$, and any class $j$, based on Eq.(\ref{eq:weq})-Eq.(\ref{eq:udfin}), we can calculate $\nabla_{\Theta}\fjj$ from:
\begin{equation*}
\begin{split}
&\nabla_{\wi} \fjj = \pfij[\x] \cdot \hth[\x], \\
&\nabla_{\hth[\x]} \fii = \bm{W}^\top \bm{\partial}_j(\x),\\
&\nabla_{\bm{\theta}} \fii = \nabla_{\bm{\theta}}\hth \bm{W}^\top \bm{\partial}_j(\x).
\end{split}
\end{equation*}

We can also adopt a dynamic programming scheme similar to the loss evaluation to speed it up. The first step is exactly the same: obtain $\bm{x}^\downarrow_{0}, \cdots, \bm{x}^\downarrow_{\ni-1}$ and $\bm{\tilde{x}}^\downarrow_{0}, \cdots, \bm{\tilde{x}}^\downarrow_{N-\ni-1}$ with the same method as the loss evaluation and  get $\mathcal{A}^{(i)}_k= \mathcal{A}^{(i)}(\x^\downarrow_k)$. Then, we have the following recursive expression for the new variable $\Gamma^{(i)}(\x^\downarrow_{k})$:
\begin{equation*}
\begin{split}
\Gamma^{(i)}(\x^\downarrow_{k+1}) = \Gamma^{(i)}(\x^\downarrow_{k}) +  \suminj \sum_{\x_n \in \Nj \cap \mathcal{A}^{(i)}_{k+1} \backslash  \mathcal{A}^{(i)}_k  } \ninj \nabla_{\bm{\Theta}}\fixn.
\end{split}
\end{equation*}
Now we can use a similar algorithm to evaluate the gradients, the details are omitted here due to its resemblance to Alg.\ref{Alg:all}.

\subsection{Squared Loss}
\textbf{Loss Evaluation}.
For each fixed class $i$, we construct an affinity matrix $\boldsymbol{Aff}^{(i)}$ such that 
\begin{equation*}
\boldsymbol{Aff}^{(i)}_{m,n} = 
\begin{cases}
\dfrac{1}{n_in_{y_m}}, & n \in \Ni, m \notin \Ni, \\
\dfrac{1}{n_in_{y_n}}, & m \in \Ni, n \notin \Ni, \\
0, & \text{Otherwise},
\end{cases}
\end{equation*}
which could be written in a matrix form,
    \begin{equation*}
     \boldsymbol{Aff}^{(i)} = \bm{D}^{(i)}({{\mathbf{1}}}-{{\mathbf{Y}}^{(i)}}){{\mathbf{Y}}^{(i)}}^{\top 
     }+{{\mathbf{Y}}^{(i)}}{{({{\mathbf{1}}}-{{\mathbf{Y}}^{(i)}})}^{\top }}\bm{D}^{(i)},
    \end{equation*}

 Then the empirical risk function under the squared surrogate loss could be reformulated as:
   \begin{equation*}
   \hat{R}_{sq}   = \sumi   \bm{\Delta}^{(i)^\top}_{sq}
      \mathcal{L}^{(i)}\bm{\Delta}^{(i)}_{sq},
   \end{equation*} 
   where
   \begin{equation}
   \bm{\Delta}^{(i)}_{sq} = \bm{Y}^{(i)} - \fii[\bm{X}], ~~\fii[\bm{X}] = [\fii[\x_1],\cdots,\fii[\x_N] ]^\top,
   \end{equation}
and $\mathcal{L}^{(i)} = diag(\bm{Aff}^{(i)}\bm{1}) - \bm{Aff}^{(i)}$ is the graph Laplacian matrix associated with $\bm{Aff}^{(i)}$. Based on this reformulation, one can find that the structure of $\mathcal{L}^{(i)}$ provides an efficient factorization scheme to speed-up loss evaluation. To see this,  we have:
   \begin{equation*}
   \hat{R}_{sq}   = \sumi   \frac{1}{2}\bm{\Delta}^{(i)\top}_{sq} (\bm{\kappa}^{(i)}\odot\bm{\Delta}^{(i)}_{sq}) -{\Delta}^{(i)}_1\cdot {\Delta}^{(i)}_2,
   \end{equation*} 
where
\begin{equation*}
\begin{split}
&{\Delta}^{(i)}_2 = \Yit \bm{\Delta}^{(i)}_{sq} \\
&{\Delta}^{(i)}_1 = \bm{\Delta}^{(i)^\top}_{sq}\Di (\Yni)\\
&\bm{\kappa}^{(i)} = \ni \Di (\Yni) + \dfrac{N_C-1}{\ni} \Yi.
\end{split}
\end{equation*}

Obviously both terms in the equation above require only $O(N_CNT_l)$ time. We summarize this factorization scheme in Alg.\ref{alg:sq}.
\begin{algorithm}[t]
  \caption{\label{alg:sq}\texttt{CalSqLoss}}
  \begin{algorithmic}[1]
    \Require $\bm{X}, \bm{Y},\alpha, \varphi(\cdot)$.
    \Ensure \texttt{loss}.
      \State $\texttt{loss} \gets 0$

    \For{i}{1}{N_C}
  
     \State  Obtain $\Di$

     \State $\bm{\Delta}^{(i)} \gets (\boldsymbol{Y}^{(i)} - \fii[\bm{X}]) $

      \State $\bm{\kappa}^{(i)} \gets \ni \Di (\Yni) + \dfrac{N_C-1}{\ni} \Yi$

      \State ${\Delta}^{(i)}_1 \gets \left(\bm{\Delta}^{(i)^\top}_{sq}(\Di \Yni)\right) $

      \State ${\Delta}^{(i)}_2 \gets \Yit \bm{\Delta}^{(i)}_{sq} $

      \State $\texttt{loss} += \frac{1}{2}\bm{\Delta}^{(i)\top}_{sq} (\bm{\kappa}^{(i)}\odot\bm{\Delta}^{(i)}_{sq}) - {\Delta}^{(i)}_1\cdot {\Delta}^{(i)}_2$

    \EndFor
  \end{algorithmic}
  \label{alg}
\end{algorithm}   

\textbf{Gradient Evaluation}. From Alg.\ref{alg:sq}, Eq.(\ref{eq:weq})-Eq.(\ref{eq:udfin}) and the chain rule, we have:
\begin{equation*}
\nabla_{\bm{w}^{(i)}}\hat{R}_{sq}  = \sumi \frac{1}{2} \cdot \nabla_{\bm{\bm{w}}^{(i)}} \bm{\Delta}^{(i)\top}_{sq} (\bm{\kappa}^{(i)}\odot\bm{\Delta}^{(i)}_{sq})  - \nabla_{\bm{w}^{(i)}}{\Delta}^{(i)}_1\cdot {\Delta}^{(i)}_2
\end{equation*}
where 
\begin{equation*}
\begin{split}
\frac{1}{2} \nabla_{\bm{w}^{(i)}} \bm{\Delta}^{(j)^\top }_{sq}(\bm{\kappa}^{(j)}\odot \bm{\Delta}^{(j)}_{sq}) &= \bm{H}(\bm{\Delta}^{(j)}_{sq} \odot \bm{\kappa}^{(j)}\odot \bm{\partial}_{i,j}   ),\\
\nabla_{\bm{w}^{(i)}}{\Delta}^{(i)}_1 {\Delta}^{(i)}_2 &= \bm{H}\left[\left({\Delta}^{(i)}_2 \cdot\Di(\Yni) + {\Delta}^{(i)}_1 \Yi  \right) \odot \bm
{\partial}_{i,j} \right],\\
\frac{1}{2} \nabla_{\bm{\theta}} \bm{\Delta}^{(j)^\top }_{sq}(\bm{\kappa}^{(j)}\odot \bm{\Delta}^{(j)}_{sq}) &= \bm{U}^{(j)} ( \kappa^{(j)} \odot \bm{\Delta}^{(j)}),\\
\nabla_{\bm{\theta}}{\Delta}^{(i)}_1 {\Delta}^{(i)}_2 &= \bm{U}^{(j)}\left[{\Delta}^{(i)}_2 \cdot\Di(\Yni) + {\Delta}^{(i)}_1 \Yi  \right],
\end{split}
\end{equation*} 

Again this shows that the gradient evaluation could be done within $O(N_CNT_{grad})$ time.

\subsection{Acceleration Scheme for general losses}
Unfortunately, the lossless acceleration for general losses is impossible in general. In this subsection, following the spirit of \cite{aucgenloss}, we provide a discussion on how to accelerate general loss functions approximately. However, we can, in turn, construct an approximation framework based on the Bernstein Polynomials. Note that the key difference between our work and \cite{aucgenloss} is that we do not need the minimax reformulation. Moreover, we also present a nice property of such approximations in terms of consistency. \\
	First of all, we define the Berstein Polynomials as the following.
	
	\begin{defi} The Berstein Polynomials of degree $m$ of the function $\varphi:[0,1]\rightarrow \mathbb{R}$ are defined, for any $u \in [0,1]$, by 
	  \begin{equation}
		\mathbb{B}_m(\varphi, u) = \sum_{i=1}^m \varphi\left(\frac{k}{m}\right) \cdot \binom{m}{k}\cdot u^k \cdot (1-u)^{m-k} = \sum_{i=1}^m \binom{m}{k} \cdot \Delta^k\varphi(0)  \cdot u^k.
	  \end{equation}
	  where $ \Delta^k\varphi(0) = \sum_{j=0}^k(-1)^{k-j} \tbinom{k}{j} \varphi\left(\frac{j}{m}\right)$ is the forward difference operator on $\varphi$ at 0.
	\end{defi}
	
	Note that the domain of the scoring functions $f^{(i)}$ are restricted to $[0,1]$, by choosing $\varphi(u) = \ell(2s-1)$, we have: 
	\begin{equation}
	  \begin{split}
		\ell(\fixm -\fixn ) &\approx \mathbb{B}_{K}\left(\varphi, \frac{1+ \fixm - \fixn}{2}\right)\\ 
		& = \sum_{k=0}^K \binom{m}{k} \Delta^k\varphi(0) \left(\varphi, \frac{1+ \fixm - \fixn}{2}\right)^k\\ 
		& = \frac{1}{K+1} \sum_{k=0}^m f^{(i)}_k(\x_m) \cdot \tilde{f}^{(i)}_k(\x_n)   
	  \end{split}
	\end{equation}
	where
	\begin{equation*}
	  \begin{split}
		& f^{(i)}_k(\x_m)  =\left(\frac{1}{2} + f^{(i)}(\x_m)\right)^k\\
		& f^{(i)}_k(\x_n)  = \sum_{j=k}^m \binom{m}{j} \cdot \binom{j}{k} \cdot \frac{(m+1) \cdot \Delta^k \varphi(0)}{2^j} \cdot \left(\frac{1}{2} - \fixn\right)^{j-k}. 
	  \end{split}
	\end{equation*}
	
	\noindent With the Bernstein Polynomials, we now construct generic acceleration methods for surrogate risk computation.
	
	\noindent \textbf{Acceleration for General Loss Calculation}. Through the lens of the Bernstein polynomials, the loss function could be approximated by the following derivation:
	
	\begin{equation*}
	  \begin{split}
		\hat{R}_{\ell} &= \sumi\sumxone  \suminj \sumxtwo \ninj \cdot  \ell\left(\fixm - \fixn\right)\\  
	&\approx \sumi\sumxone  \suminj \sumxtwo \ninj \cdot \frac{1}{K+1} \sum_{k=0}^K f^{(i)}_k(\x_m) \cdot \tilde{f}^{(i)}_k(\x_n)\\ 
	&= \sumi \frac{1}{K+1} \bm{e}_+^{(i)\top} \bm{e}_-^{(i)}\\ 
	& \triangleq \hat{R}^K_{\ell}
	  \end{split}
	\end{equation*}
	where
	\begin{equation*}
	  \begin{split}
		\bm{e}_+^{(i)}  &= \left[\sumxone \frac{f^{(i)}_1(\x_m)}{n_i}, \cdots,  \sumxone \frac{f^{(i)}_{K}(\x_m)}{n_i} \right]^\top,\\
		\bm{e}_-^{(i)}  &= \left[\suminj \sumxtwo \frac{\tilde{f}^{(i)}_1(\x_n)}{n_j}, \cdots,  \suminj \sumxtwo \frac{\tilde{f}^{(i)}_K(\x_n)}{n_j} \right]^\top.
	  \end{split}
	\end{equation*}
	
	\noindent Obviously, it only requires $O(K\cdot N_C \cdot N)$ time to finish the loss computation.

	\noindent \textbf{Acceleration for General Gradient Calculation}. Similarly, the gradient calculation could be approximated by the following derivation: 
	
	\begin{equation*}
	  \begin{split}
		\nabla \hat{R}^K_{\ell} &= \sumi\sumxone  \suminj \sumxtwo \ninj \cdot \nabla \ell\left(\fixm - \fixn\right)\\  
	&\approx \sumi\sumxone  \suminj \sumxtwo \ninj \cdot \frac{1}{K+1} \sum_{k=0}^K f^{(i)}_k(\x_m) \cdot \nabla \tilde{f}^{(i)}_k(\x_n)\\ 
	&= \sumi \frac{1}{K+1}  (\nabla \bm{e}_+^{(i)})^\top \bm{e}_-^{(i)}  + \bm{e}_+^{(i)\top} (\nabla \bm{e}_-^{(i)} )
	  \end{split}
	\end{equation*}
	
	\noindent $\nabla \bm{e}_+^{(i)}$ and $\nabla \bm{e}_-^{(i)}$ could be obtained in a similar spirit to the calculation of the exponential loss. It thus  requires $O(K\cdot N_C \cdot N \cdot T_{grad})$ time to finish  the gradient computation.\\ 
	\noindent \textbf{Consistency of the Approximated Loss function} Now, we proceed to present an elegant property of the approximated risk functions. Specifically, we show that $\hat{R}^K_\ell$
	is, in itself, a consistent loss function if $\ell$ is consistent. To do this, we need the following theorem about the Bernstein polynomials.
	\begin{thm}\label{thm:approx}
	If $f \in C^k[0,1]$, for some $k \ge 0$, then :
	\begin{equation*}
	   m \le \frac{d^k(f(x))}{dx^k} \le M, ~\Rightarrow~ c_k \cdot m \le \frac{d^k(\mathbb{B}_K(f,x))}{dx^k} \le c_k \cdot M
	  \end{equation*}
	\end{thm}
	Then the following corollary gives the final result.
	\begin{col}
	For all surrogate loss function $\ell \in C^2[0,1]$, 
	\[\hat{R}_\ell ~\text{is}~ \mauc ~\text{consistent}~  \Rightarrow  \hat{R}^K_\ell ~\text{is}~ \mauc ~\text{consistent}~ \]
	\end{col}
\begin{pf}
  Again, by choosing $\varphi(u)= \ell(2s-1)$, we have $\varphi(u) \in C^2[0,1]$. The result follows directly from Thm.\ref{thm:approx}, and Thm.\ref{thm:consq}. 
  \qed
\end{pf}

\subsection{Summary and Concluding Remarks}
\begin{table}[htbp]
  \centering \small
  \caption{Acceleration for three losses \label{tab:acc_app}, where  $\bar{N} = \sumi \ni \log \ni + (N-\ni) \log (N-\ni)$   }
  \begin{tabular}{llll}
    Algorithms    & \multicolumn{1}{l}{loss} & \multicolumn{1}{l}{gradient} & \multicolumn{1}{l}{requirement}\\
  \toprule
  exp + acceleration   &   $O(N_C \cdot N \cdot T_\ell)$      & $O(N_C \cdot N \cdot T_{grad})$ & $\min_i \ni \gg 2$  \\
  squared + acceleration & $O(N_C \cdot N \cdot T_\ell)$     & $O(N_C \cdot N \cdot T_{grad})$ & $e^{\frac{1}{2} (N-\ni)} \gg  \ni \gg N - e^{\frac{1}{2}\ni}$ \\
  hinge + acceleration &   $O(N_C\cdot \bar{N} \cdot  T_\ell)$      & $O(N_C \cdot \bar{N} \cdot T_{grad})$ & $\min_i \ni \gg 2$\\
  general + acceleration &   $O(K \cdot N_C\cdot N \cdot  T_\ell)$      & $O( K \cdot N_C \cdot N \cdot T_{grad})$ & $e^{\frac{1}{2} (N-\ni)} \gg  \ni \gg N - e^{\frac{1}{2}\ni}$ \\

  \midrule
  w/o  acceleration &    $O(\sumi \suminj n_in_j\cdot T_\ell)$   & $O(\sumi \suminj n_in_j\cdot T_{grad})$ & $\backslash$ \\
  \bottomrule
  \end{tabular}%
\end{table}%

\begin{equation*}
\sumi \suminj n_in_j - N_C \cdot N =  \sumi \left[ n_i(N-\ni) -  (\ni + (N-\ni))  \right].
\end{equation*}
 To ensure $N_C \cdot N < \sumi \suminj n_in_j$, one must let \[n_i(N-\ni) >  (\ni + (N-\ni)),~~~\forall i. \]  Obviously, this requirement could be met when $\min_i \ni \gg 2$. Similarly, we have:
 \begin{equation*}
\sumi \suminj n_in_j - N_C \cdot \bar{N} =  \sumi \left[ n_i(N-\ni) -  \ni\log(\ni) - (N-\ni)\log(N-\ni)  \right].
\end{equation*}
In this sense  $N_C \cdot \bar{N} < \sumi \suminj n_in_j$ could be satisfied if  \[\frac{1}{2} (N-\ni) >  \log(\ni),~~ \frac{1}{2} (\ni) >  \log(N - \ni), ~~ \forall i.\] This means that
\[ \exp(\frac{1}{2} (N-\ni)) \gg  \ni \gg N - \exp(\frac{1}{2}\ni), ~~\forall i. \] 
should be guaranteed to ensure a significant improvement. It is easy to see that all the requirements are essentially loose restricts on the data distribution which could be satisfied for the majority of real-world datasets.  As another remark, we do not provide a detailed calculation of $\nabla_{\bm{\theta}}\hth$, since it follows the chain rule and could be obtained from off-the-shelf toolboxes such as tensorflow and pytorch where auto differentiation is supported.  Moreover, our acceleration schemes are based either on changing the order of operations or on reformulating the original loss. Hence, when the model is implemented in such auto differentiation platforms, the gradient acceleration does not need to be explicitly implemented. One just needs to implement the accelerated loss evaluation to fix the computational graph, the gradient evaluation then follows the graph naturally.

\section{Experiments}\label{app:exp}

\subsection{Numerical Validations for the Acceleration Algorithms}
\begin{figure*}[h]  
  \centering
   \includegraphics[width=0.99\textwidth]{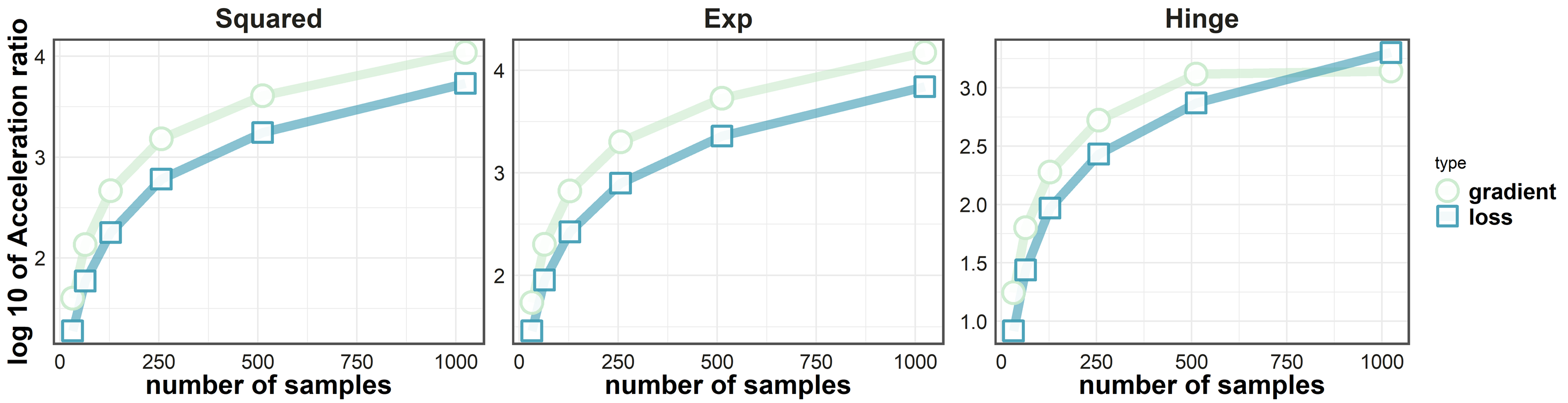} 
  

  \caption{\label{fig:acc}\textbf{Acceleration Ratio vs. Sample Scale}}
  \end{figure*} 
\textbf{Setting}. As a warm-up experiment, we first validate the effectiveness of our proposed speed-up methods detailed in Sec.\ref{sec:d}. Note that our algorithms only concern loss and gradient evaluations, which do not depend on the optimization algorithms in question. Hence, to be fair, we will perform stand-alone running time tests where no optimization algorithms are involved. Following this spirit, our comparisons are based on the results from 30 repetitions of pure loss and gradient evaluations. To do this, we randomly generate a series of dataset $\bm{Z}_i = (\bm{X}_i, \bm{Y}_i), ~ i=1,2,\cdots,5$, where $\bm{X}_i \in \mathbb{R}^{N_k \times d}, \bm{Y}_i \in \mathbb{R}^{N_i \times N_C}$ are the input  feature matrix and the output one-hot label matrix. Specifically, we set $N_i = \{32,64,128,256,512,1024\},~ d = 100, ~ N_C = 5$. $\bm{X}_i$ is generated from the uniform distribution within $[0,1]$. Denote $\rho_i = \nicefrac{\ni}{N}$, then $\bm{Y}$ is generated such that $\rho_1 = 0.2, \rho_2 = 0.1, \rho_3= 0.2, \rho_4 = 0.4, \rho_5 = 0.1$. Moreover, we fix a linear scoring function $f(\cdot)$ with softmax output $f^{(i)}(\bm{x}) = \mathsf{softmax}(\bm{W}^{(i)}\bm{x})$, where the weight matrix $\bm{W} = [\bm{W}^{(1)},\cdots, \bm{W}^{(5)}]$ is generated in the same way as $\bm{X}$. Then we test the running time of the naive evaluation algorithm and our speed-up algorithms respectively over different $\bm{Z}_i$. 

\textbf{Results and Analysis}. To show the efficiency improvements, we plot the mean of the acceleration ratio in Fig.\ref{fig:acc}, which is  defined as:
\begin{equation*}
\begin{split}
&\text{acceleration ratio} \\
&=\frac{\text{Running time of the naive method}}{ \text{Running time of the acceleration method}}.
\end{split}
\end{equation*}
over 30 repetitions against the scale of the datasets. We have the following two findings: (a) The proposed algorithms generate significant speed-up over all sample scales being tested.
They produce $10\sim100$x speed-up when the sample size remains as small as 32 while the result gradually climbs-up as the sample size enlarges and finally reaches  $10000$x speed-up for the squared and the exponential loss when the sample size becomes 1024.  This shows that the proposed algorithms could both work for mini-batch-based and full-batch-based optimization methods. (b) The curves suggest that the acceleration ratios are roughly scaled as $O(N)$, which is consistent with the complexity analysis shown in Tab.\ref{tab:acc_app}. To see this, for squared and exponential loss, we have: 
\begin{equation*}
\begin{split}
\text{acceleration ratio}&\approx \frac{\sumi \suminj n_in_j}{N_C\cdot N }\\ 
&= \frac{\sumi \suminj \rho_i\cdot \rho_j}{N_C} \cdot N
\end{split}
\end{equation*}
Note that we keep $\rho_i$ and $N_C$ as constants in our experiment, this suggests that the acceleration ratio should be $O(N)$. For hinge loss, there are extra factors from $\log \ni$ and $\log(N-\ni)$. However, these factors mostly remain at a constant level considering that the number of samples being tested is not large. This is why we also observe an almost linear scale-up of the acceleration ratio of hinge loss. Moreover, for hinge loss, we see that the increase of the acceleration ratio becomes slower for the gradient evaluations when the number of samples surpasses 500. A proper reason for this phenomenon is that the evaluation algorithms for hinge loss are partially implemented by \texttt{Cython}. For a large scale dataset, the contribution of a  \texttt{Cython} implementation should be more significant for loss than for gradient. This is because that  the matrix operations in gradient evaluation could be accelerated by proper and compact use of \texttt{numpy} even without \texttt{Cython}, while the loss evaluations, containing mostly elementary operations, enjoy a shaper speed-up from  \texttt{Cython}.
\subsection{Dataset Description}\label{App:des}
\begin{table*}[h]
  \begin{minipage}[b]{0.52\linewidth}
  \normalsize
  \centering
  \caption{\label{tab:info}{Basic Information of the Datasets}}
      \begin{tabular}{lcccc}
      Dataset & \multicolumn{1}{c}{$\#samples$} & \multicolumn{1}{c}{$\#classes$} & \multicolumn{1}{c}{$\#features$} & \multicolumn{1}{c}{$r_\chi$} \\
      \toprule
      Balance & $625$   & $3 $    & $4$     & $5.88$ \\
      Dermatology & $357 $  & $6$     & $34$    & $5.55$ \\
      Ecoli & $336 $  & $8$     & $7$     & $71.50$ \\
      New-thyroid & $215$   & $3$     & $5$     & $5.00$ \\
      Pageblocks & $548 $  & $5$     & $10$    & $164.00$ \\
      SegmentImb & $749$   & $7$     & $18 $   & $20.31$\\
      Shuttle & $2,175 $ & $5$     & $9$     & $853.00$ \\
      Svmguide2 & $391$   & $3$     & $20$    & $4.17$ \\
      Yeast & $1,484$  & 10    & 8     &$92.60$ \\
      \midrule
      CIFAR-100-Imb & $23,350$ & $100$ & $2,048$ & $50.00$\\
      User-Imb & $24,400$ & $12$  &$21,527$ & $20.00$ \\
      iNaturalList2017  & $675,170$  &  $5,089$   & $2,048$ &    $276.4$ \\
      \bottomrule
      \end{tabular}%
  \end{minipage}
  \hspace{2cm}
  \begin{minipage}[b]{0.3\linewidth}
  \centering
  \normalsize
   \caption{\label{tab:class} $\ni$ for User-Imb}
      \begin{tabular}{lr}
      class & \multicolumn{1}{l}{sample} \\
          \toprule
      F23-  & $800$ \\
      F24-26 & $400$ \\
      F27-28 & $400$ \\
      F29-32 & $800$ \\
      F33-42 & $1,600$ \\
      F43+  & $400$ \\
      M22-  & $1,600$ \\
      M23-26 & $8,000$ \\
      M27-28 & $800$ \\
      M29-31 & $1,600$ \\
      M32-38 & $8,000$ \\
      M39+  & $8,000$ \\
      \bottomrule
      \end{tabular}%
  \end{minipage}
  
  \end{table*}
  \begin{table}[h]
    \centering
    \caption{ The label distribution for CIFAR-100-Imb Dataset, where each cell in the table presents a specific $\ni$. The row title in the left shows the range of id in the row. The class ids are numbered in the same way as the original dataset.}
    \scalebox{1}{
      \begin{tabular}{lrrrrrrrrrr}
             \multicolumn{11}{c}{number of samples in each class} \\
      \toprule
      $1\sim10$ & \cellcolor[rgb]{ .984,  .851,  .82}$100$ & \cellcolor[rgb]{ .996,  .898,  .859}$50$ & \cellcolor[rgb]{ .918,  .62,  .604}$360$ & \cellcolor[rgb]{ 1,  .922,  .882}$20$ & \cellcolor[rgb]{ 1,  .922,  .882}20 & \cellcolor[rgb]{ .933,  .675,  .655}$300$ & \cellcolor[rgb]{ .906,  .565,  .553}$420$ & \cellcolor[rgb]{ 1,  .922,  .882}$20$ & \cellcolor[rgb]{ .984,  .851,  .82}$100$ & \cellcolor[rgb]{ .957,  .765,  .737}$200$ \\
      $11\sim20$ & \cellcolor[rgb]{ .984,  .851,  .82}100 & \cellcolor[rgb]{ .984,  .851,  .82}$100$ & \cellcolor[rgb]{ .859,  .404,  .404}$600$ & \cellcolor[rgb]{ .996,  .898,  .859}$50$ & \cellcolor[rgb]{ .984,  .851,  .82}$100$ & \cellcolor[rgb]{ .996,  .898,  .859}$50$ & \cellcolor[rgb]{ .996,  .898,  .859}$50$ & \cellcolor[rgb]{ .984,  .851,  .82}$100$ & \cellcolor[rgb]{ .984,  .851,  .82}$100$ & \cellcolor[rgb]{ .984,  .851,  .82}$100$ \\
      $21\sim30$ & \cellcolor[rgb]{ .996,  .898,  .859}$50$ & \cellcolor[rgb]{ .984,  .851,  .82}$100$ & \cellcolor[rgb]{ .859,  .404,  .404}$600$ & \cellcolor[rgb]{ .984,  .851,  .82}$100$ & \cellcolor[rgb]{ .996,  .898,  .859}$50$ & \cellcolor[rgb]{ .859,  .404,  .404}$600$ & \cellcolor[rgb]{ .984,  .851,  .82}$100$ & \cellcolor[rgb]{ 1,  .922,  .882}$20$ & \cellcolor[rgb]{ .957,  .765,  .737}$200$ & \cellcolor[rgb]{ 1,  .922,  .882}$20$ \\
      $31\sim40$ & \cellcolor[rgb]{ .859,  .404,  .404}$600$ & \cellcolor[rgb]{ .984,  .851,  .82}$100$ & \cellcolor[rgb]{ .984,  .851,  .82}$100$ & \cellcolor[rgb]{ .933,  .675,  .655}$300$ & \cellcolor[rgb]{ .996,  .898,  .859}$50$ & \cellcolor[rgb]{ .984,  .851,  .82}$100$ & \cellcolor[rgb]{ 1,  .922,  .882}$20$ & \cellcolor[rgb]{ .859,  .404,  .404}$600$ & \cellcolor[rgb]{ .984,  .851,  .82}$100$ & \cellcolor[rgb]{ .859,  .404,  .404}$600$ \\
      $41\sim50$ & \cellcolor[rgb]{ 1,  .922,  .882}$20$ & \cellcolor[rgb]{ .859,  .404,  .404}$600$ & \cellcolor[rgb]{ .984,  .851,  .82}$100$ & \cellcolor[rgb]{ .933,  .675,  .655}$300$ & \cellcolor[rgb]{ 1,  .922,  .882}$20$ & \cellcolor[rgb]{ .933,  .675,  .655}$300$ & \cellcolor[rgb]{ .906,  .565,  .553}$420$ & \cellcolor[rgb]{ .859,  .404,  .404}$600$ & \cellcolor[rgb]{ .89,  .514,  .506}$480$ & \cellcolor[rgb]{ .984,  .851,  .82}$100$ \\
      $51\sim60$ & \cellcolor[rgb]{ 1,  .922,  .882}$20$ & \cellcolor[rgb]{ 1,  .922,  .882}$20$ & \cellcolor[rgb]{ .996,  .898,  .859}$50$ & \cellcolor[rgb]{ .984,  .851,  .82}$100$ & \cellcolor[rgb]{ 1,  .922,  .882}$20$ & \cellcolor[rgb]{ .906,  .565,  .553}$420$ & \cellcolor[rgb]{ .859,  .404,  .404}$600$ & \cellcolor[rgb]{ .996,  .898,  .859}$50$ & \cellcolor[rgb]{ 1,  .922,  .882}$20$ & \cellcolor[rgb]{ .984,  .851,  .82}$100$ \\
      $61\sim70$ & \cellcolor[rgb]{ .996,  .898,  .859}$50$ & \cellcolor[rgb]{ .984,  .851,  .82}$100$ & \cellcolor[rgb]{ .957,  .765,  .737}$200$ & \cellcolor[rgb]{ .918,  .62,  .604}$360$ & \cellcolor[rgb]{ .984,  .851,  .82}$100$ & \cellcolor[rgb]{ .984,  .851,  .82}$100$ & \cellcolor[rgb]{ .984,  .851,  .82}$100$ & \cellcolor[rgb]{ .996,  .898,  .859}$50$ & \cellcolor[rgb]{ .984,  .851,  .82}$100$ & \cellcolor[rgb]{ .996,  .898,  .859}$50$ \\
      $71\sim80$ & \cellcolor[rgb]{ .996,  .898,  .859}$50$ & \cellcolor[rgb]{ .969,  .808,  .776}$150$ & \cellcolor[rgb]{ .859,  .404,  .404}$600$ & \cellcolor[rgb]{ .996,  .898,  .859}$50$ & \cellcolor[rgb]{ .933,  .675,  .655}$300$ & \cellcolor[rgb]{ 1,  .922,  .882}$20$ & \cellcolor[rgb]{ .969,  .808,  .776}$150$ & \cellcolor[rgb]{ 1,  .922,  .882}$20$ & \cellcolor[rgb]{ 1,  .922,  .882}$20$ & \cellcolor[rgb]{ .984,  .851,  .82}$100$ \\
      $81\sim90$ & \cellcolor[rgb]{ 1,  .922,  .882}$20$ & \cellcolor[rgb]{ .89,  .514,  .506}$480$ & \cellcolor[rgb]{ .996,  .898,  .859}$50$ & \cellcolor[rgb]{ .996,  .898,  .859}$50$ & \cellcolor[rgb]{ .984,  .851,  .82}$100$ & \cellcolor[rgb]{ 1,  .922,  .882}$20$ & \cellcolor[rgb]{ .996,  .898,  .859}$50$ & \cellcolor[rgb]{ 1,  .922,  .882}20 & \cellcolor[rgb]{ .933,  .675,  .655}$300$ & \cellcolor[rgb]{ .996,  .898,  .859}$50$ \\
      $91\sim100$ & \cellcolor[rgb]{ .996,  .898,  .859}$50$ & \cellcolor[rgb]{ .918,  .62,  .604}$360$ & \cellcolor[rgb]{ .984,  .851,  .82}$100$ & \cellcolor[rgb]{ .969,  .808,  .776}$150$ & \cellcolor[rgb]{ 1,  .922,  .882}$20$ & \cellcolor[rgb]{ .984,  .851,  .82}$100$ & \cellcolor[rgb]{ .984,  .851,  .82}$100$ & \cellcolor[rgb]{ 1,  .922,  .882}$20$ & \cellcolor[rgb]{ .984,  .851,  .82}$100$ & \cellcolor[rgb]{ .996,  .898,  .859}$50$ \\
      \bottomrule
      \end{tabular}%
    }
    \label{tab:cifar_dist}%
  \end{table}%

  Here, we provide a more detailed description for traditional datasets from LIBSVM and KEEL.
\begin{enumerate}[ itemindent=0pt, leftmargin =15pt]
\item[(a)]  \textbf{LIBSVM Datasets}. ~~ \url{https://www.csie.ntu.edu.tw/~cjlin/libsvmtools/datasets/}
\begin{itemize}
\item \textbf{Abalone$\star$}. The task of this dataset is to predict the age of abalone from physical measurements. The features include  basic information such as the gender, length, height of a given subject. To construct a multi-class problem from this dataset. We regard each possible age as a class. On top of this, we  merge the first three classes, and we also merge  all the classes greater than 20.
\item \textbf{Shuttle}. The shuttle dataset contains 9 features all of which are numerical. The first one being time. The last column is the class which has been coded as follows: 1 Rad Flow,,2 Fpv Close,3 Fpv Open,4 High,5 Bypass,6 Bpv Close,7 Bpv Open.
\item \textbf{Svmguide-2}. \cite{guide} released this dataset when the authors are providing a guidance for the \texttt{LIBSVM} library.  This dataset comes from a bacteria protein subcellular localization dataset which was proposed in \cite{bacter}.
\end{itemize}

\item[(b)] \textbf{KEEL Datasets}. ~~\url{http://www.keel.es/}

\begin{itemize}
\item \textbf{Balance}. This data set was generated to model psychological experimental results. Each example is classified as having the balance scale tip to the right, tip to the left, or be balanced. The attributes are the left weight, the left distance, the right weight, and the right distance.
\item \textbf{Dermatology}. The differential diagnosis of erythemato-squamous diseases is a real problem in dermatology. They all share the clinical features of erythema and scaling, with few differences. The diseases in this group are psoriasis, seborrheic dermatitis, lichen planus, pityriasis rosea, chronic dermatitis, and pityriasis rubra pilaris. Usually a biopsy is necessary for the diagnosis, but, unfortunately, these diseases share many histopathological features as well. Another difficulty for the differential diagnosis is that a disease may show the features of another disease at the beginning stage and may have the characteristic features at the following stages. Patients were first evaluated clinically with 12 features. Afterwards, skin samples were taken for the evaluation of 22 histopathological features. The values of the histopathological features are determined by an analysis of the samples under a microscope. In the dataset constructed for this domain, the family history feature has the value 1 if any of these diseases has been observed in the family, and 0 otherwise. The age feature simply represents the age of the patient. Every other feature (clinical and histopathological) was given a degree in the range of 0 to 3. Here, 0 indicates that the feature was not presented, 3 indicates the largest amount possible, and 1, 2 indicate the relative intermediate values.

\item \textbf{Ecoli}. The task of this dataset is to predict protein localization sites, where the features are given as a series of scores evaluated by standard detection methods.
\item \textbf{Glass}. A Glass type identification dataset with 9 features.
\item \textbf{New Thyroid}. This is a Thyroid Disease (New Thyroid) dataset with three classes: normal hyper and hypo.

\item \textbf{Page Blocks}. The task of this dataset is to classify all the blocks of the page layout of a document that has been detected by a segmentation process, where each of the observed blocks has 10 features describing its basic information.

\item \textbf{Wine}. A Wine classification dataset where the model should classify a given data point to one of the types of wines based on 13 dimensions of features.

\end{itemize}
\item[(c)] \textbf{Other Datasets}:
\begin{enumerate}[ itemindent=0pt, leftmargin =8pt]
\item[(1)] \textbf{CIFAR-100-Imb}. We sampled an imbalanced version of the CIFAR-100 \footnote{\url{https://www.cs.toronto.edu/~kriz/cifar.html}} dataset, which originally contains 100 image classes each with 600 instances.  The 100 classes are encoded as $1,\cdots 100$. For each class $i$, we randomly sampled $n_i$ instances shown in Tab.\ref{tab:cifar_dist} to gain an imbalanced version of the dataset.

\item[(2)] \textbf{User-Imb}. The original dataset is collected from TalkingData, a famous third-party mobile data platform from China, which predicts mobile users’ demographic characteristics based on their app usage records. The dataset is collected for the Kaggle Competition named Talking Data Mobile User Demographics. \footnote{\url{https://www.kaggle.com/c/talkingdata-mobile-user-demographics/data}} The raw features include logged events, app attributes, and device information. There are 12 target classes 'F23-', 'F24-26','F27-28','F29-32', 'F33-42', 'F43+', 'M22-', 'M23-26', 'M27-28', 'M29-31', 'M32-38', 'M39+', which describe the demographics (gender and age) of users. In our experiments, we sample an imbalanced subset of the original dataset to leverage a class-skewed dataset. The number of instances in each class after resampling is listed as Tab.\ref{tab:class}.
\item[(3)]   
  \textbf{iNaturalist2017.} iNaturalist Challenge 2017 dataset\footnote{https://www.kaggle.com/c/inaturalist-challenge-at-fgvc-2017} is a large-scale image classification benchmark with 675,170 images covering 5,089 different species of plants and animals. We split the dataset into the training set, validation set, and test set at a ratio of 0.7:0.15:0.15. Since directly training deep models in such a large-scale dataset is time-consuming, we instead generate 2048-d features with a ResNet-50 model pre-trained on ImageNet for each image and train models with three fully-connected layers for all methods. We utilize Adam optimizer to train the models, with an initial learning rate of $10^{-5}$. To ensure all categories are covered in a mini-batch, the batch size is set to 8196. Other hyperparameters are the same as those in the CIFAR-100-Imb dataset.

\end{enumerate}

\end{enumerate}

\subsection{Implementation Details}
\textbf{Infrastructure}. All the experiments are carried out on a ubuntu 16.04.6 server  equipped with  Intel(R) Xeon(R) CPU E5-2620 v4 cpu and a TITAN RTX GPU. The codes are implemented via \texttt{python 3.6.7}, the basic dependencies are: \texttt{pytorch} (v-1.1.0), \texttt{sklearn} (v-0.21.3), \texttt{numpy} (v-1.16.2). For traditional datasets, we implement our proposed algorithms with the help of the \texttt{sklearn} and \texttt{numpy}. For hinge loss, we use Cython to accelerate the dynamic programming algorithm. For the deep learning datasets, our proposed algorithms are implemented with \texttt{pytorch}.

\textbf{Evaluation Metric}. Given a trained scoring function $f = (f^{(1)},\cdots, f^{(N_C)})$, all the forthcoming results are evaluated with reward reversion of MAUC, where we denote it as $\mathsf{MAUC}^\uparrow$:
\begin{equation*}
\mathsf{MAUC}^\uparrow  = \frac{1}{N_C \cdot (N_C - 1)} \cdot \sumi\suminj \frac{\left|\left\{(\x_1,\x_2): \x_1 \in \mathcal{N}_i, ~\x_2 \in \mathcal{N}_j, f^{(i)}(\x_1) > f^{(i)}(\x_2) \right\}\right|}{\ni\nj}
\end{equation*}
in this subsection.

\textbf{Traditional Datasets}. For all the experiments, hyper-parameters are tuned based on the training and validation set while the performances are evaluated over a test set, where the training, valid, and test set accounts for $80\%,~ 10\%,~ 10\%$ of the original dataset, respectively. To remain the label distribution of the overall dataset, the dataset splits are generated with a stratified sampling strategy with respect to classes.  The experiments are done with 15 repetitions for each involved algorithm. Since all the datasets fall in this class contains simple features, we use a linear model composited with a softmax transformation as the scoring function. More formally, given an instance $\x$, we have $f(\x)= \mathsf{softmax}(\bm{W}\bm{x})$, where $\bm{W} = [\bm{\omega}^{(1)}, \cdots, \bm{\omega}^{(N_C)}] \in \mathbb{R}^{d \times N_C}$ is the weight of the linear function. For the sampling-based competitors, the corresponding sampling algorithm is first performed on the dataset, then a scoring function is trained based on the multiclass cross-entropy loss. For LR, the scoring function is directly trained based on the multiclass cross-entropy loss without sampling. For our proposed algorithms, we directly wrap the $\mauc$ surrogate losses on the scoring function $f$ and perform the training process without any simpling procedure. The hyper-parameters we adopted to produce the performance of our algorithms are shown in Tab.\ref{tab:par_trad}. 
\begin{table}[h]
\begin{minipage}[a]{0.4\linewidth}
 \centering
 \small
  \caption{Best parameters for ($\lambda, \alpha$) over traditonal datasets}

    \begin{tabular}{lccc}
       Dataset   & Ours1 & Ours2 & Ours3 \\
    \toprule
    balance & $(1e\text{-}4,0.9)$ & $(1e\text{-}4,0.9)$ & $(1e\text{-}4,0.7)$ \\
    dermatology & $(1e\text{-}4,0.9)$ & $(1e\text{-}4,0.9)$ & $(1e\text{-}4,0.3)$ \\
    ecoli & $(6e\text{-}4,0.6)$ & $(4e\text{-}4,0.9)$ & $(1e\text{-}4,0.2)$ \\
    new\text{-}thyroid & $(1e\text{-}4,0.9)$ & $(1e\text{-}4,0.5)$ & $(1e\text{-}4,0.9)$ \\
    pageblocks & $(6e\text{-}4,0.8)$ & $(2e\text{-}4,0.9)$ & $(6e\text{-}4,0.5)$ \\
    segmentImb & $(2e\text{-}4,0.9)$ & $(4e\text{-}4,0.9)$ & $(1e\text{-}4,0.3)$ \\
    shuttle & $(2e\text{-}4,0.8)$ & $(1e\text{-}4,0.7)$ & $(1e\text{-}4,0.5)$ \\
    svmguide2 & $(1e\text{-}4,0.9)$ & $(1e\text{-}4,0.9)$ & $(2e\text{-}4,0.4)$ \\
    yeast & $(6e\text{-}4,0.3)$ & $(9e\text{-}3,0.1)$ & $(1e\text{-}4,0.2)$ \\
    \bottomrule
    \end{tabular}
  \label{tab:par_trad}%
\end{minipage}
\hspace{2cm}
\begin{minipage}[a]{0.4\linewidth}
\begin{minipage}[a]{\linewidth}
\centering
 \small
  \caption{\label{tab:arch_cifar}The Common Architecture for CIFAR-100-Imb Dataset}
    \begin{tabular}{lccc}
    layers & \multicolumn{1}{l}{$n_{units}$} & activation  & \multicolumn{1}{l}{$BN$} \\
    \toprule
    $fc_1$  & $1,024$ & relu   & \cmark \\ 
    $fc_2$  & $256$ & relu  & \cmark \\  
    $output$ & $100$ & softmax  & \xmark \\ 
    \bottomrule
    \end{tabular}%
\end{minipage} 
\textcolor{white}{dsadsa}\\
\begin{minipage}[a]{\linewidth}
\centering
 \small
  \caption{\label{tab:arch_user}The Common Architecture for  User-Imb Dataset}
    \begin{tabular}{lccc}
    layers & \multicolumn{1}{l}{$n_{units}$} & activation  & \multicolumn{1}{l}{$BN$} \\
    \toprule
    $fc_1$  & $128$ & relu   & \cmark \\ 
    $fc_2$  & $64$ & relu  & \cmark \\  
    $output$ & $12$ & softmax  & \xmark \\ 
    \bottomrule
    \end{tabular}%
\end{minipage}
\end{minipage}

\end{table}%

\textbf{Deep Learning Datasets}. For all the experiments, hyper-parameters are tuned based on the training and validation set while the performances are evaluated over a test set, where the training, valid, and test set accounts for $80\%,~ 10\%,~ 10\%$ of the original dataset, respectively. To remain the label distribution of the overall dataset, the dataset splits are generated with a stratified sampling strategy with respect to classes. For these two datasets, all the models are constructed with a deep neural network. The scoring functions share a common form as $f(\x) = \mathsf{softmax}(fc_2(fc_1(\x)))$, where $fc_1, fc_2$ are fully-connected layers. For CIFAR-100-Imb, we adopt the ResNet-50 pre-trained features from \texttt{layer4} after average pooling as the input for each method. For User-Imb, the models are trained from scratch. The architecture we use for  CIFAR-100-Imb and User-Imb are shown in Tab.\ref{tab:arch_cifar} and Tab.\ref{tab:arch_user}, respectively. For the sampling-based competitors, the corresponding sampling algorithm is first performed on the dataset, then a scoring function is trained based on the multiclass cross-entropy loss together with a three-layered deep neural network. For LR, the scoring function is directly trained based on the multiclass cross-entropy loss and the neural net without sampling. For our proposed algorithms, we directly wrap the $\mauc$ surrogate losses on the neural net $f$ and perform the training process without any simpling procedure. The hyper-parameters we adopted to produce the performance of  our algorithms are shown in Tab.\ref{tab:par_cifar} and Tab.\ref{tab:par_user}, respectively for CIFAR-100-Imb and User-Imb.

\begin{table}[htbp]
  \centering
  \small
  \caption{\label{tab:par_cifar}Hyperparameters for CIFAR-100-Imb}
  \begin{tabular}{c|lllllll}
    & batch size & lr    & weight decay & lr decay & epoch & gamma & \#epochs before lr decay \\
\midrule
Ours1 & $1000$  & $0.0012 $&$ 0.000005$ &$0.97$  & $50.00$    & $1.00$     & $5.00$ \\
Ours2 & $1000$  & $0.001$ & $0.000001 $& $0.99$  & $48.00$    & $4.00$     & $5.00$ \\
Ours3 & $1000$  & $0.001$ & $0.00005$ &  $0.99$  & $50.00$    & $4.00$     & $3.00$ \\
\bottomrule
\end{tabular}%
\end{table}%

\begin{table}[htbp]
  \centering
  \small
  \caption{\label{tab:par_user} Hyperparameters for User-Imb}

  \begin{tabular}{l|llllll}
    & batch size & lr    & weight decay & lr decay & gamma & \#epochs before lr decay \\
\toprule
Ours1 & $32$    & $0.005$ &$ .0001$ & $0.97$  & $0.5$   &   $1.00$ \\
Ours2 & $32$    & $0.005$ &$ .0001$ & $0.97$  & $2.0$     & $1.00$ \\
Ours3 & $32$    & $0.005$ &$ .0001$ & $0.97$  & $2.0$     & $1.00$ \\
\bottomrule
\end{tabular}%

\end{table}%

\subsection{Running Time Comparison for Deep Learning Datasets.} 

To validate the effectiveness of our acceleration method, we report the average running time per epoch in Tab.\ref{tab:deepacc} here for all  three deep learning datasets. \emph{Accelerated} refers to our proposed method. For the \emph{Selected} method, the sample indexes of different classes are cached in different tensors. Naive refers to a straightforward implementation. 

\begin{table}[htbp]
  \centering
  \caption{Running Time Comparison for Different Implementation Schemes, - Refers to the Fact that the Running Time is Longer than 12 h. }  \label{tab:deepacc}%
    \begin{tabular}{clrrr}
    loss  & implementation & \multicolumn{1}{c}{CIFAR-100} & \multicolumn{1}{c}{User} & \multicolumn{1}{c}{iNaturalist} \\
    \midrule
    \multirow{3}[1]{*}{Square} & Accelerated & 0.95  & 8.59  & 121.8347 \\
          & Selected & 3111.1 & 89.35 & - \\
          & Naïve & 51747.03 & 360.36 & - \\
          \midrule
    \multirow{3}[1]{*}{EXP} & Accelerated & 0.97  & 9.41  & 117.4387 \\
          & Selected & 2392.55 & 72.31 & - \\
          & Naïve & 56138.42 & 352.77 & - \\
    \midrule
    \multirow{3}[2]{*}{Hinge} & Accelerated & 0.83  & 8.42  & 115.4508 \\
          & Selected & 2703.15 & 75.15 & - \\
          & Naïve & 51302.48 & 352.45 & - \\
    \bottomrule
    \end{tabular}%
  
\end{table}%

\end{document}